\documentclass{article}

\usepackage{microtype}
\usepackage{graphicx}
\usepackage{booktabs}
\usepackage{hyperref}

\usepackage[utf8]{inputenc}
\usepackage[T1]{fontenc}
\usepackage{amsfonts}
\usepackage{subfig}
\usepackage{amsmath}
\usepackage{amssymb}
\usepackage{balance}
\usepackage{bm}
\usepackage{clipboard}
\usepackage{dsfont}
\usepackage{enumitem}
\usepackage{float}
\usepackage{makecell}
\usepackage{mathbbol}
\usepackage{mathtools, nccmath}
\usepackage{microtype}
\usepackage{multirow}
\usepackage{nicefrac}
\usepackage{pifont}
\usepackage{setspace}
\usepackage{url}
\usepackage{xcolor}

\DeclareSymbolFontAlphabet{\mathbb}{AMSb}
\DeclareSymbolFontAlphabet{\mathbbl}{bbold}

\newcommand{\dataset}{\scalebox{0.825}{$\mathcal{D}$}}
\newcommand{\zero}{\scalebox{0.9}{$0$}}
\newcommand{\one}{\scalebox{0.9}{$1$}}
\newcommand{\proof}{\vspace{-0.5em}\textit{Proof}.}
\newcommand{\QED}{\hfill\scalebox{0.85}{$\square$}}
\newcommand{\Appendix}{Appendix \ref{app:proofs}.\QED}
\newcommand\numberthis{\addtocounter{equation}{1}\tag{\theequation}}
\DeclareMathOperator{\nil}{\varnothing}
\DeclareMathOperator*{\argmax}{argmax}

\newcommand{\pix}{\kern 0.1em}
\newcommand{\pmm}{\kern 0.25em$\pm$\kern 0.15em}
\newcommand{\pms}{\kern 0.10em$\pm$\kern 0.05em}

\usepackage[noend]{algorithmic}
\renewcommand\algorithmiccomment[1]{
\hfill$\triangleright$ {#1}\hspace{-0.6em}
}

\newtheorem{theorem}{Theorem}

\newtheorem{proposition}[theorem]{Proposition}

\newtheorem{example}{Example}

\usepackage[accepted]{icml2020}

\definecolor{mydarkblue}{rgb}{0,0.08,0.45}

\newcommand{\mytitle}
{
Inverse Active Sensing\raisebox{0.5pt}{\scalebox{0.9}{\pix:}}\\
\scalebox{0.95}{Modeling and Understanding Timely Decision-Making}
}

\icmltitlerunning{Inverse Active Sensing}

\begin{document}
\twocolumn[
\icmltitle{\mytitle}
\icmlsetsymbol{equal}{*}

\begin{icmlauthorlist}
\icmlauthor{Daniel Jarrett}{cam}
\icmlauthor{Mihaela van der Schaar}{cam,ucla}
\end{icmlauthorlist}

\vspace{-0.75em}

\icmlaffiliation{cam}{Department of Mathematics, University of Cambridge, UK.}
\icmlaffiliation{ucla}{Department of Electrical Engineering, UCLA, USA}
\icmlcorrespondingauthor{Daniel Jarrett}{daniel.jarrett@maths.cam.ac.uk}
\icmlkeywords{Machine Learning, ICML}

\vskip 0.3in
]

\printAffiliationsAndNotice{}

\allowdisplaybreaks

\begin{abstract}

Evidence-based decision-making entails collecting (costly) observations about an underlying phenomenon of interest, and subsequently committing to an (informed) decision on the basis of accumulated evidence. In this setting, \textit{active sensing} is the goal-oriented problem of efficiently selecting which acquisitions to make, and when and what decision to settle on. As its complement, \textit{inverse active sensing} seeks to uncover an agent's preferences and strategy given their observable decision-making behavior. In this paper, we develop an expressive, unified framework for the general setting of evidence-based decision-making under endogenous, context-dependent time pressure\textemdash which requires negotiating (subjective) tradeoffs between accuracy, speediness, and cost of information. Using this language, we demonstrate how it enables \textit{modeling} intuitive notions of surprise, suspense, and optimality in decision strategies (the forward problem). Finally, we illustrate how this formulation enables \textit{understanding} decision-making behavior by quantifying preferences implicit in observed decision strategies (the inverse problem).

\end{abstract}\section{Introduction}\label{sec:introduction}

Modeling decision-making processes is a central concern in computational and behavioral science, with important applications to medicine \citep{li2015sequential}, economics \citep{clithero2018response}, and cognition \citep{drugowitsch2014relation}. \mbox{In evid-} ence-based decision-making, the agent first \textit{collects} a series of observations about an underlying phenomenon of interest, then subsequently \textit{commits} to an informed decision based on the accumulated evidence. As popular examples, consider the problems of hypothesis testing, medical diagnostics, and employee hiring: In each case, the decision-maker first conducts \textit{acquisitions} for information (i.e. hypothesis tests, diagnostic procedures, and candidate interviews), the results on which the final \textit{decision} is then based (i.e. the selected hypothesis, the declared disease, and the hiring decision).

In this context, \textit{active sensing} is the goal-directed task of selecting which acquisitions to make, when to stop gathering information, and what decision to ultimately settle on. Active sensing strategies have been studied for such applications as multi-hypothesis testing \citep{naghshvar2013active}, sensory inference \citep{ahmad2013active}, and visual search \citep{butko2010infomax}. These are typically formulated simply as sequential identification problems with an infinite horizon\textemdash that is, of minimizing inaccuracies against a unit sampling cost. However, for the general task of evidence-based decision-making, two critical shortcomings bear emphasis\textemdash a lack of \textit{expressivity}, and a need for \textit{specification}.

\begin{table}[b]\small
\setlength\tabcolsep{2.3pt}
\renewcommand{\arraystretch}{1.05}
\vspace{-1.2em}
\label{tab:introduction_applications}
\begin{center}
\begin{tabular}{cccc} \toprule
\multicolumn{1}{c}{\textbf{Problem Setting}} &
\multicolumn{1}{c}{Decisions} &
\multicolumn{1}{c}{Acquisitions} &
\multicolumn{1}{c}{Outcomes} \\
\midrule
Hypothesis Testing   & Hypotheses   & Hyp. Tests  & Observations \\
Medical Diagnosis    & Diseases     & Diag. Tests & Results      \\
Cognitive Science    & Responses    & Perceptions & Evidence     \\
Sensory Inference    & Targets      & Fixations   & Sensations   \\
Marketing \& Sales   & Demographic  & Outreaches  & Engagements  \\
Recruiting \& Hiring & Hire or Fire & Interviews  & Assessments  \\
\bottomrule
\end{tabular}
\end{center}
\vspace{-1.2em}
\caption{\textit{Applications and terminology in timely decision-making}.}
\vspace{-0.2em}
\end{table}

\newcolumntype{A}{>{\centering\arraybackslash}m{2.7cm}}
\newcolumntype{B}{>{\centering\arraybackslash}m{3.5cm}}
\newcolumntype{C}{>{\centering\arraybackslash}m{2.9cm}}
\newcolumntype{D}{>{\centering\arraybackslash}m{2.5cm}}
\newcolumntype{E}{>{\centering\arraybackslash}m{4.8cm}}

\begin{table*}[t]\small
\setlength\tabcolsep{2.0pt}
\renewcommand{\arraystretch}{1.0}
\vspace{-0.8em}
\caption{\textit{Comparison of models for timely decision-making}. Our general framework accounts for the endogeneity (due to $\lambda$) and context-dependence (due to $\theta$) of time pressure, as well as differential costs of acquisition, deadline penalties, and preferences. $^{1}$\pix\citet{ahmad2013active}, $^{2}$\pix\citet{chernoff1959sequential}, $^{3}$\pix\citet{naghshvar2013active}, $^{4}$\pix\citet{alaa2016balancing}, $^{5}$\pix\citet{dayanik2013reward}, $^{6}$\pix\citet{frazier2008sequential}. While the second row of models incorporates (external) deadlines, they do not consider \textit{active} sensing\textemdash the first only considers sampling from a single stream, and the latter two only consider a \textit{passive} supply of information, whence the problem readily reduces to optimal stopping.}
\label{tab:related_forward}
\begin{center}
\begin{tabular}{ABCDE}
\toprule
\textbf{Framework} & {Accuracy of Decision} & {Breach of Deadline} & {Cost of Acquisition} & {Time Pressure} \\
\midrule
Ahmad,$^{1}$ Chernoff,$^{2}$ Naghshvhar,$^{3}$ etc.
& \raisebox{0.5em}{${\textstyle\sum}_{\theta^{\prime}}\eta_{\text{\pix a},\theta^{\prime}}\mathds{1}_{\{\theta=\theta^{\prime},\theta\neq\hat{\theta}\}}$}
& \raisebox{0.2em}{-}
& \raisebox{0.5em}{$\eta_{\text{\pix c}}\tau$}
& \raisebox{0.5em}{$\mathbb{P}\{\delta=t\}=0$} \\
Alaa,$^{4}$ Dayanik,$^{5}$ Frazier,$^{6}$ etc.
& \raisebox{0.75em}{${\textstyle\sum}_{\theta^{\prime}}\eta_{\text{\pix a},\theta^{\prime}}\mathds{1}_{\{\theta=\theta^{\prime},\theta\neq\hat{\theta},\tau<\delta\}}$}
& \raisebox{0.75em}{$\eta_{\text{\pix b}}\mathds{1}_{\{\tau=\delta\}}$}
& \raisebox{0.75em}{$\eta_{\text{\pix c}}\tau$}
& \raisebox{0.75em}{e.g.$^{5}$ $\mathbb{P}\{\delta=t\}=p(\one-p)^{t}$} \\
\textbf{(Ours)}
& \raisebox{0.2em}{${\textstyle\sum}_{\theta^{\prime}}\eta_{\text{\pix a},\theta^{\prime}}\mathds{1}_{\{\theta=\theta^{\prime},\theta\neq\hat{\theta},\tau<\delta\}}$}
& \raisebox{0.2em}{${\textstyle\sum}_{\theta^{\prime}}\eta_{\text{\pix b},\theta^{\prime}}\mathds{1}_{\{\theta=\theta^{\prime},\tau=\delta\}}$}
& \raisebox{0.2em}{$\textstyle\sum_{t=0}^{\tau-1}\eta_{\text{\pix c},\lambda_{t}}c_{\lambda_{t}}$}
& \raisebox{0.2em}{$\mathbb{P}\{\delta=t\}=p_{\theta,\lambda_{t}}\prod_{t^{\prime}=0}^{t-1}(\one-p_{\theta,\lambda_{t^{\prime}}})$} \\
\bottomrule
\end{tabular}
\end{center}
\vspace{-1.8em}
\end{table*}

First, any sufficiently realistic decision model must account for the presence, endogeneity, and context-dependence of \textit{time pressure}. While deadlines are studied in \citet{frazier2008sequential} and \citet{dayanik2013reward}, they are external variables, and their settings are passive (i.e. sampling from a single exogenous supply of information). This is unrealistic: While lengthy aptitude tests may be more discriminative for recruiting purposes, their grueling nature may also cause more candidates drop out of the pipeline entirely. Similarly, the probability of an adverse medical event that aborts the diagnostic process depends on both the nature of the test chosen (i.e. \textit{endogenous}) and the underlying disease itself (i.e. \textit{context-dependent}). What we desire is a more expressive framework capable of modeling such tradeoffs.

Second, even the simplest decision models suffer from a need for specification. At a minimum, they require explicit knowledge of the relative penalties of decision inaccuracies and costs of acquisition. The need for complete specification of (subjective) \textit{preferences} severely dampens the practical utility of any analysis. For instance, we expect doctors to care much more about correctly diagnosing a lethal disease than another condition that presents with similar symptoms. Do they actually? By how much? Similarly, do recruiters care more about identifying the best candidates, or simply avoiding the worst at all costs? What we desire is a way to perform \textit{inverse active sensing}\textemdash that is, to uncover preferences that effectively underlie observed decision behavior.

\textbf{Contributions}.
We tackle both challenges simultaneously. In this paper, we first develop an expressive, unified framework for decision-making under endogenous and context-dependent time pressure. In this formulation, a decision-maker is required to negotiate the \textit{subjective} tradeoff between accuracy, speediness, and the cost of information. Second, using this language, we demonstrate how it enables modeling intuitive notions of surprise, suspense, and optimality in decision strategies (the \textit{forward} problem). Finally, we illustrate how this formulation enables understanding decision-making behavior by quantifying preferences implicit in observed decision strategies (the \textit{inverse} problem).

\textbf{Implications}.
Decision-making behavior is heterogeneous, and different agents are driven by different priorities. The implications are clear: An expressive forward model allows \textit{prescribing} (optimal) decision-making in the presence of subjective preferences, while an inverse procedure allows \mbox{\textit{describing}} (observed) decision-making in terms of preferences implicit among agents and institutions. In medicine, some populations may be subject to less rigorous diagnostic scrutiny than others \cite{mckinlay2007sources}, and test prescriptions often skewed by financial incentive \citep{song2010regional}. The potential for detecting biases and quantifying hidden priorities in decision systems offers a first step towards a more methodical understanding of clinical practice.

\vspace{0.1em}
\section{Timely Decision-Making}
\vspace{0.1em}

\begin{figure}[t]
\centering
\includegraphics[width=\linewidth]{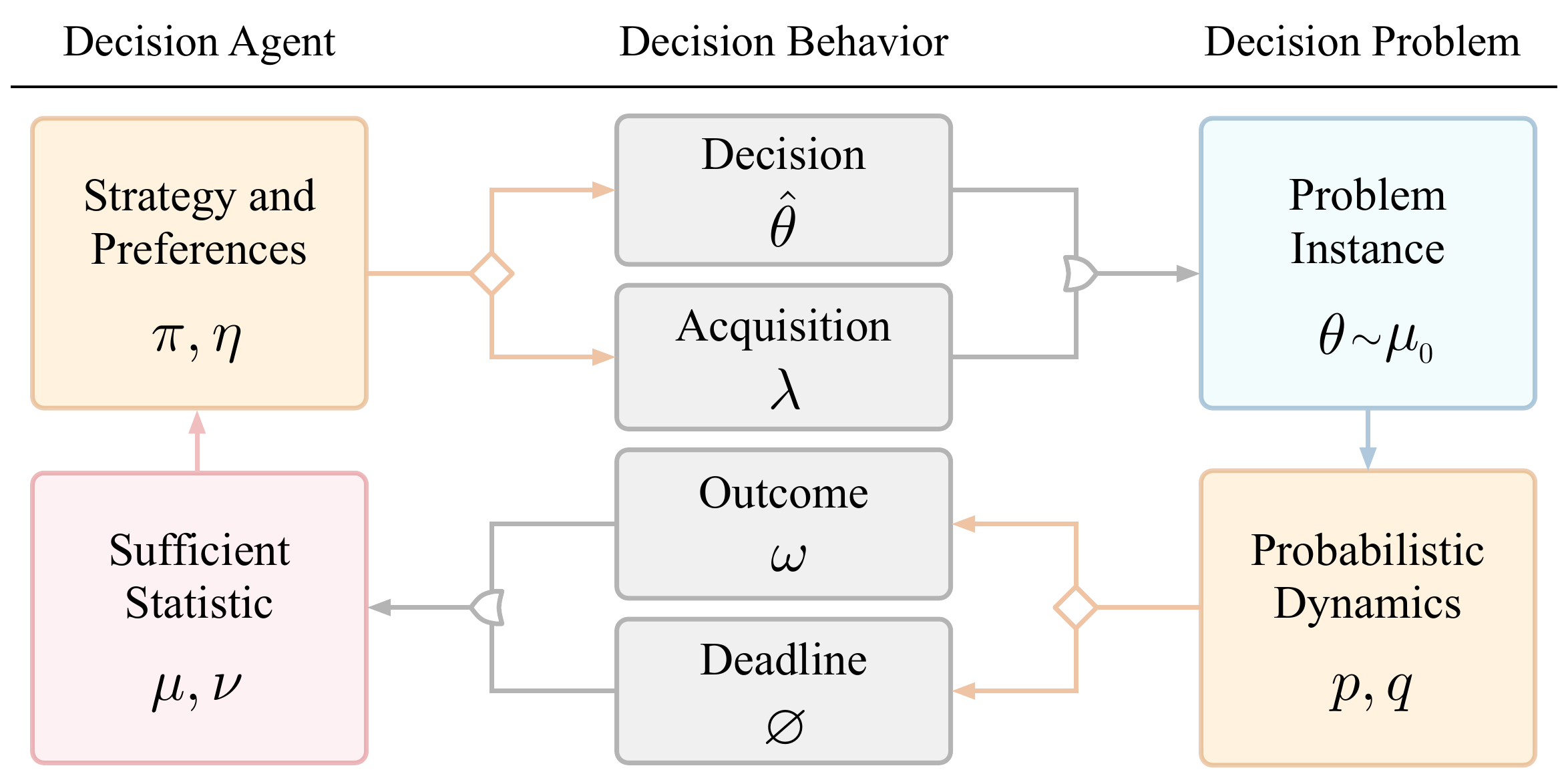}
\vspace{-1.9em}
\caption{\small \textit{Active sensing and inverse active sensing}. Decision problems are characterized by their probabilistic dynamics (right), and each instance is drawn from a known distribution (blue). Decision agents are characterized by their strategy and preferences (left), and maintain an internal representation according to a Bayesian recognition model (red). When an agent is presented with a problem, \textit{active sensing} produces observable behavior (center). Conversely, given an agent's observed behavior with respect to a problem, \textit{inverse active sensing} seeks to recover their preferences and strategy.}
\label{fig:pull}
\vspace{-1.5em}
\end{figure}

First, we formulate the problem of timely decision-making (Section \ref{sec:decision_problem}), and derive the Bayesian recognition model for an agent (Section \ref{sec:recognition_model}). Next, we characterize optimal active sensing strategies (Section \ref{sec:forward_problem}), on the basis of which we formalize and describe a solution for inverse active sensing (Section \ref{sec:inverse_problem}). Figure \ref{fig:pull} provides a high-level block-diagram. {\color{mydarkblue}\textsc{\textbf{note}}}: As a guide for anchoring our subsequent developments, see Figure \ref{fig:pull2} (on page \hyperlink{page.7}{7}) for a map of our key results.

\vspace{-0.5em}
\subsection{Decision Problem}\label{sec:decision_problem}

Let $\Theta$ give the space of decisions (e.g. possible diagnoses), $\Lambda$ the space of acquisitions (e.g. medical tests), and $\Omega$ the space of outcomes (e.g. diagnostic results) of acquisitions. We consider the setting where these spaces are finite, but note that our analysis easily extends to continuous outcomes, or distinct spaces of outcomes per test. Briefly, the goal of an agent is to commit to a decision $\hat{\theta}\in\Theta$ (e.g. issue an official diagnosis) at some decision time $\sigma\in\mathbb{N}$ before a probabilistic deadline $\delta\in\mathbb{N}$ (e.g. complication of the underlying disease). The active sensing challenge is in adaptively, sequentially choosing \textit{which} acquisitions \mbox{to perform, \textit{when} to} stop gathering information, and \textit{what} decision to settle on.

The outcome $\omega\in\Omega$ of each acquisition $\lambda\in\Lambda$ is a random variable distributed according to the (stationary) generating distribution $q_{\theta,\lambda}(\omega)\doteq\mathbb{P}\{\omega|\theta;\lambda\}$, where $\theta\in\Theta$ is the unknown latent multinoulli variable representing the correct decision (e.g. the true underlying disease). We assume that $\{q_{\theta,\lambda}\}_{\theta\in\Theta,\lambda\in\Lambda}$ are known (e.g. the known power of each medical test). Each $\lambda_{t}$ conducted at time $t\in\mathbb{N}$ yields a corresponding outcome $\omega_{t+1}$ at the next step, and the outcomes are conditionally independent over time. For brevity, let $\lambda_{0:t}$ denote $\{\lambda_{0},...,\lambda_{t}\}$, and analogously $\omega_{1:t}$ for $\{\omega_{1},...,\omega_{t}\}$.

If the probabilistic deadline $\delta$ interrupts the trial, further interaction is void. The deadline is a random variable distributed as $\mathbb{P}\{\delta=t\}=p_{\theta,\lambda_{t}}\prod_{t^{\prime}=0}^{t-1}(\one-p_{\theta,\lambda_{t^{\prime}}})$ where constants $p_{\theta,\lambda}\in(0,1)$ are specific to the acquisition $\lambda$ (i.e. endogenous) and latent variable $\theta$ (i.e. context-dependent). This is in contrast to typical sequential identification models with either no deadline or external deadlines (see Table \ref{tab:related_forward}). We assume that $\{p_{\theta,\lambda}\}_{\theta\in\Theta,\lambda\in\Lambda}$ are known (e.g. the known risks of complication for various diseases and procedures). The presence of time pressure is important\textemdash in general decision-makers are not given an infinite window of opportunity to ponder their options; moreover, more informative acquisitions are often riskier (e.g. more invasive).

\textbf{Episodes and Risks}.
Each \textit{problem instance} is drawn as $\theta\sim\mu_{0}$ for some prior $\mu_{0}$. In the most general case, statistics $\mu_{0}$ may be stratified by subpopulation (e.g. as a function of patient demographics), or even as a learned mapping from covariates. Here we simply take it that $\mu_{0}$ is known (e.g. from medical experience or literature), and defer further aspects of modeling for later work. A \textit{decision episode} is characterized by the tuple $(\lambda_{0:\tau-1},\tau,\hat{\theta})$, where $\tau=\min\{\delta,\sigma\}$ denotes the \textit{stopping time} for the episode; note that $\hat{\theta}=\nil$ should the deadline occur before a decision is registered.

Each episode is generated by a \textit{decision strategy} $\pi$, which produces\textemdash possibly stochastically\textemdash for each time $t$ either a (continuing) acquisition $\lambda_{t}\in\Lambda$ or (terminating) decision $\hat{\theta}\in\Theta$. Let $c_{\lambda}$ denote the immediate fixed cost of performing $\lambda$ (e.g. the monetary expense of ordering a test), and let coefficient vectors \smash{$\eta_{\text{\pix a}}\in\mathbb{R}\raisebox{-2pt}{$_{^{+}}^{|\Theta|}$}$}, \smash{$\eta_{\text{\pix b}}\in\mathbb{R}\raisebox{-2pt}{$_{^{+}}^{|\Theta|}$}$}, and \smash{$\eta_{\text{\pix c}}\in\mathbb{R}\raisebox{-2pt}{$_{^{+}}^{|\Lambda|}$}$} respectively denote preference weights assigned to the importance of deciding \textit{accurately} (i.e. on the correct decision), \textit{speedily} (i.e. before the deadline), and \textit{efficiently} (i.e. with minimal cost).
Then the loss function is given as follows:
\vspace{-1pt}
\begin{align*}\label{loss_function}
\ell&(\lambda_{0:\tau-1},\tau,\hat{\theta};\eta) \numberthis
\\[-1pt]
&\doteq
{\textstyle\sum}_{\theta^{\prime}\in\Theta}
\eta_{\text{\pix a},\theta^{\prime}}
\mathds{1}_{\{\theta=\theta^{\prime},\theta\neq\hat{\theta},\tau<\delta\}}
&&\pix\raisebox{-2pt}{\scalebox{0.9}{$\triangleleft\text{\small~Accuracy of Decision}$}}
\\[-1pt]
&\pix+
{\textstyle\sum}_{\theta^{\prime}\in\Theta}
\eta_{\text{\pix b},\theta^{\prime}}
\mathds{1}_{\{\theta=\theta^{\prime},\tau=\delta\}}
&&~~~\raisebox{-1pt}{\scalebox{0.9}{$\triangleleft\pix\text{\small~Breach of Deadline}$}}
\\[-1pt]
&\pix+
\textstyle\sum_{t=0}^{\tau-1}
\eta_{\text{\pix c},\lambda_{t}}
c_{\lambda_{t}}
&&~~~\scalebox{0.9}{$\triangleleft\text{\small~Cost of Acquisition}$}
\end{align*}

\vspace{-7pt}
where we explicitly indicate dependence on \smash{$\eta\doteq(\eta_{\text{\pix a}},\eta_{\text{\pix b}},\eta_{\text{\pix c}})$}. Then risk associated with executing strategy $\pi$ is given by:
\begin{equation}
\setlength{\abovedisplayskip}{6pt}
\setlength{\belowdisplayskip}{6pt}
L(\pi;\eta)
\doteq
\mathbb{E}_{p,q}[
\ell(\lambda_{0:\tau-1},\tau,\hat{\theta};\eta)
|\mu_{0},\pi]
\end{equation}
where the expectation is taken with respect to problem dynamics $p$ and $q$. Importantly, this gives the most flexible framework: In addition to incorporating differential costs of acquisition \smash{$c\in\mathbb{R}\raisebox{-2pt}{$_{^{+}}^{|\Lambda|}$}$}, the subjective penalties now depend on both $\theta$ and $\lambda$ (see Table \ref{tab:related_forward}). For instance, failing to correctly diagnose a relatively innocuous condition may incur less damage than a lethal disease (viz. $\eta_{\text{\pix a}}$); likewise, scaring away a good candidate with tough interviews may entail a larger sacrifice than losing a bad one to attrition (viz. $\eta_{\text{\pix b}}$).

\vspace{-0.5em}
\subsection{Beliefs and Information}\label{sec:recognition_model}

In order to model and understand decision strategies, we first describe an agent's Bayesian recognition model for representing information. Note that this is not an assumption: We are \textit{not} effectively assuming that real-world decision-makers indeed perform exact Bayesian inference\textemdash they most likely do not; instead, we are simply deriving a compact (internal) representation for use in describing their (external) behavior.

To this end, we highlight two implications of the endogeneity and context-dependence of time pressure. First, since acquisitions are no longer independent from time-to-event (i.e. survival), we cannot use the standard Bayes update as in prior work (Proposition \ref{thm:sufficient_statistic}). Second, each step now conveys two pieces of information: one from the acquired outcomes, and another from the process survival itself (Proposition \ref{thm:active_passive}).

\vspace{-3pt}
\Copy{thm_sufficient_statistic}{
\begin{proposition}[Sufficient Statistic]\upshape
\label{thm:sufficient_statistic}
Let $\nu_{t}\doteq\mathds{1}_{\{\delta>t\}}$ denote the \textit{survival} process, with initial value $\nu_{0}=1$. Then the \textit{posterior} process \smash{$\mu_{t}\in\Delta(\Theta)$} is given by the following:
\begin{equation}
\setlength{\abovedisplayskip}{6pt}
\setlength{\belowdisplayskip}{6pt}
\begin{split}
\mu_{t}=(\one-\pix&\nu_{t-1})\mu_{t-1}
+(
(\one-\nu_{t})\bar{M}(\lambda_{t-1},\mu_{t-1}) \\[-1pt]
&+
\nu_{t}M(\lambda_{t-1},\mu_{t-1},\omega_{t})
)\nu_{t-1} \\
\end{split}
\end{equation}
where the \textit{continual} update $M:\Lambda\times\Delta(\Theta)\times\Omega\rightarrow\Delta(\Theta)$ re- turns a distribution assigning to element $\theta$ the probability:
\begin{equation}
\setlength{\abovedisplayskip}{6pt}
\setlength{\belowdisplayskip}{6pt}
\begin{split}
\frac{(\one-p_{\theta,\lambda_{t-1}})q_{\theta,\lambda_{t-1}}(\omega_{t})\mu_{t-1}(\theta)}{\sum_{\theta^{\prime}\in\Theta}(\one-p_{\theta^{\prime},\lambda_{t-1}})q_{\theta^{\prime},\lambda_{t-1}}(\omega_{t})\mu_{t-1}(\theta^{\prime})} \\
\end{split}
\end{equation}
and where the \textit{terminal} update $\bar{M}:\Lambda\times\Delta(\Theta)\rightarrow\Delta(\Theta)$ re- turns a distribution assigning to element $\theta$ the probability:
\begin{equation}
\setlength{\abovedisplayskip}{4pt}
\setlength{\belowdisplayskip}{4pt}
\begin{split}
p_{\theta,\lambda_{t-1}}\mu_{t-1}(\theta)/{\textstyle\sum}_{\theta^{\prime}\in\Theta}p_{\theta^{\prime},\lambda_{t-1}}\mu_{t-1}(\theta^{\prime})
\end{split}
\end{equation}
Moreover, the sequence $(\mu_{t},\nu_{t})_{t=0}^{\infty}$ is a \textit{controlled Markov process}, where the control inputs are the acquisitions $\lambda_{t}$.
\end{proposition}
\proof
}
\Appendix

This allows us to formally define decision strategies $\pi$ as maps from $\mu_{t},\nu_{t}$ into $\Delta(\Lambda\cup\Theta)$. However, the dynamics of acquisition and survival are \textit{entangled} here. The following result separately identifies the two sources of information:

\vspace{-3pt}
\Copy{thm_active_passive}{
\begin{proposition}[Active and Passive Information]\upshape
\label{thm:active_passive}
The information gleaned from (costly) acquisitions and (costless) observations of survival can be uniquely decomposed as:
\begin{equation}
\setlength{\abovedisplayskip}{2pt}
\setlength{\belowdisplayskip}{2pt}
\begin{split}
\mu_{t}=\tilde{\mu}_{t}+\alpha_{t}+\beta_{t}
\end{split}
\end{equation}
where $\tilde{\mu}_{t}$ is a \textit{martingale} that captures information obtained from the (actively) acquired results, the (continual) compensator \smash{$\alpha_{t}=A(\mu_{t-1},\lambda_{t-1},\nu_{t-1},\nu_{t})$} (passively) incorporates the bias from the ongoing process \textit{survival} (where $\alpha_{0}$\pix$=$\pix\pix$0$):\\
\begin{equation}
\setlength{\abovedisplayskip}{-8pt}
\setlength{\belowdisplayskip}{5pt}
\begin{split}
\alpha_{t}(\theta)
=
\alpha_{t-1}&(\theta)
-
\mu_{t-1}(\theta)\nu_{t-1}\nu_{t} \\[-3pt]
&~\cdot(p_{\theta,\lambda_{t-1}}-\bar{p}_{\mu_{t},\lambda_{t-1}})/(\one-\bar{p}_{\mu_{t},\lambda_{t-1}}) \\
\end{split}
\end{equation}
and \smash{$\beta_{t}=B(\mu_{t-1},\lambda_{t-1},\nu_{t-1},\nu_{t})$} is the (terminal) compensator that analogously incorporates the bias from process \textit{stoppage} (where $\beta_{0}=0$)\textemdash if the deadline were breached:
\begin{equation}
\setlength{\abovedisplayskip}{5pt}
\setlength{\belowdisplayskip}{5pt}
\begin{split}
\beta_{t}(\theta)
=
\beta_{t-1}(\theta)
&+
\mu_{t-1}(\theta)\nu_{t-1}(\one-\nu_{t}) \\[-3pt]
&~~~~~~\cdot(p_{\theta,\lambda_{t-1}}-\bar{p}_{\mu_{t},\lambda_{t-1}})/\bar{p}_{\mu_{t},\lambda_{t-1}} \\
\end{split}
\end{equation}
where for brevity we denote the weighted average posterior probability of failure \smash{$\bar{p}_{\mu_{t},\lambda_{t-1}}\doteq{\textstyle\sum}_{\theta^{\prime}\in\Theta}p_{\theta^{\prime},\lambda_{t-1}}\mu_{t-1}(\theta^{\prime})$}.
\end{proposition}
\proof
}
\Appendix

This is intuitive: Before the deadline, the posterior process for any $\theta\in\Theta$ behaves like a \textit{supermartingale} whenever the corresponding deadline risk \smash{$p_{\theta,\lambda_{t-1}}$} is greater than the average \smash{$\bar{p}_{\mu_{t},\lambda_{t-1}}$}, and behaves like a \textit{submartingale} where it is less risky. Equality holds only if the deadline is exogenous, whence we recover the classic sequential identification setting where the posterior process is (always) a martingale.

\section{Strategies and Preferences}

Having formalized the decision problem and recognition model, we are ready for the forward and inverse problems. First, we consider optimal active sensing strategies (translating preferences \textit{into} behavior). Our results then enable inverse active sensing (inferring preferences \textit{from} behavior).

\vspace{-0.5em}
\subsection{Optimal Active Sensing}\label{sec:forward_problem}

Two distinguishing characteristics of our timely decision framework is that it requires \textit{active} strategies, and that decisions are made under \textit{time pressure}. This is in contrast to the (passive) settings in \citet{dayanik2013reward} and \citet{frazier2008sequential} with only a single choice of acquisition, where the decision problem readily reduces to optimal stopping. This is also in contrast to the (infinite) decision horizons in \citet{ahmad2013active} and \citet{naghshvar2011information}, where optimal strategies are free from considerations of survival.

First, we characterize the optimal value function, and show that it is unique and computable (Proposition \ref{thm:optimal_value}). We then describe the optimal choice between continuing and terminating (Proposition \ref{thm:continue_terminate}). Finally, we interpret the risk-benefit tradeoff underlying the optimal acquisition (Proposition \ref{thm:surprise_suspense}).

To begin, observe that at each time $t$, we wish to minimize the \textit{to-go} component of risk, motivating the value function:
\begin{equation}
\setlength{\abovedisplayskip}{4pt}
\setlength{\belowdisplayskip}{4pt}
\begin{split}
V^{\pi}(\mu_{t}&,\nu_{t};\eta)
\doteq\pix
\mathbb{E}_{p,q}[
\ell(\lambda_{0:\tau-1},\tau,\hat{\theta};\eta) \\[-2pt]
&|\lambda_{0:t-1},\mu_{t},\nu_{t},\pi]
-
\textstyle\sum_{t^{\prime}=0}^{t-1}
\eta_{\text{\pix c},\lambda_{t^{\prime}}}
c_{\lambda_{t^{\prime}}} \\
\end{split}
\end{equation}
for strategy $\pi$ and preferences $\eta$. Now, it is tempting to immediately identify the optimal value function $V^{*}(\mu_{t},\nu_{t};\eta)$ with the fixed point of a dynamic programming operator. However, similar to even the passive case of \citet{dayanik2013reward}, active sensing is \textit{not} a discounted (nor fixed-horizon) problem; further, the stopping time itself is an endogenous (choice) random variable. Consequently, such an operator is not necessarily contractive (hence the optimal value is not necessarily unique or computable). Fortunately, we can leverage the (almost surely) finite decision deadline, and the following result assures us that these properties still hold:

\vspace{-3pt}
\Copy{thm_optimal_value}{
\begin{proposition}[Optimal Value]\upshape
\label{thm:optimal_value}
The optimal value function $V^{*}(\mu_{t},\nu_{t};\eta)$ is a fixed point of the operator $\mathbb{B}$ defined over the space of functions \smash{$V\in\mathbb{R}\raisebox{-2pt}{$_{^{+}}^{\Delta(\Theta)\times\{0,1\}}$}$} as follows:
\begin{equation}
\setlength{\abovedisplayskip}{5pt}
\setlength{\belowdisplayskip}{6pt}
\begin{split}
&~~~~~~~~~~~~~~~~~~~~~~~~~~~~(\mathbb{B}V)(\mu_{t},\nu_{t};\eta)= \\[-2pt]
&\min\{
{\textstyle\inf}_{\hat{\theta}^{\prime}\in\Theta}
\bar{Q}_{\hat{\theta}^{\prime}}(\mu_{t},\nu_{t};\eta)
,
{\textstyle\inf}_{\lambda^{\prime}_{t}\in\Lambda}
Q_{\lambda^{\prime}_{t}}(\mu_{t},\nu_{t};\eta)
\}
\end{split}
\end{equation}
where the (continual) $Q$-factors for \textit{acquisitions} quantify the risk-to-go upon performing acquisition $\lambda_{t}$, given by:
\begin{equation}
\setlength{\abovedisplayskip}{5pt}
\setlength{\belowdisplayskip}{6pt}
\begin{split}
Q_{\lambda_{t}}(\mu_{t}&,\nu_{t};\eta)
=
(\one-\nu_{t})V(\mu_{t},\zero;\eta)
+
\eta_{\text{\pix c},\lambda_{t}}
c_{\lambda_{t}} \\[-2pt]
&+
\nu_{t}
\mathbb{E}_{p,q}[
V(\mu_{t+1},\nu_{t+1};\eta)
|\lambda_{t},\mu_{t},\nu_{t}=\one] \\
\end{split}
\end{equation}
and the (terminal) $Q$-factors for \textit{decisions} quantify the risk upon settling on the final choice of decision $\hat{\theta}$, given by:
\begin{equation}
\setlength{\abovedisplayskip}{5pt}
\setlength{\belowdisplayskip}{6pt}
\begin{split}
\bar{Q}_{\hat{\theta}}(\mu_{t},\nu_{t};\eta)
=
(\one&-\nu_{t}){\textstyle\sum}_{\theta^{\prime}\in\Theta}
\eta_{\text{\pix b},\theta^{\prime}}
\mu_{t}(\theta^{\prime}) \\[-2pt]
&~~~~~\pix\pix+
\nu_{t}
{\textstyle\sum}_{\theta^{\prime}\in\Theta,\theta\neq\hat{\theta}}
\eta_{\text{\pix a},\theta^{\prime}}
\mu_{t}(\theta^{\prime})
\end{split}
\end{equation}
Moreover, the operator $\mathbb{B}$ is \textit{contractive}, and the optimal value function is therefore the \textit{unique} fixed point admitted.
\end{proposition}
\proof
}
\Appendix

As a result, we have that $V^{*}(\mu_{t},\nu_{t};\eta)$ is (uniquely) identifiable, and is (iteratively) computable via successive approximations. Now, the natural question becomes \textit{when} to keep collecting information, versus stopping and committing to a decision. The following gives a geometric characterization:

\vspace{-3pt}
\Copy{thm_continue_terminate}{
\begin{proposition}[Continuation and Termination]\upshape
\label{thm:continue_terminate}
~Denote by $m_{\theta}\in\Delta(\Theta)$ each vertex in the simplex, and let the optimal \textit{aggregate} $Q$-factor for continuation be given by:
\begin{equation}
\setlength{\abovedisplayskip}{3pt}
\setlength{\belowdisplayskip}{3pt}
\begin{split}
Q^{*}(\mu_{t},\nu_{t};\eta)\doteq{\textstyle\inf}_{\lambda^{\prime}_{t}\in\Lambda}Q^{*}_{\lambda^{\prime}_{t}}(\mu_{t},\nu_{t};\eta)
\end{split}
\end{equation}
and likewise \smash{$\bar{Q}(\mu_{t},\nu_{t};\eta)\doteq{\textstyle\inf}_{\hat{\theta}^{\prime}\in\Theta}\pix\bar{Q}_{\hat{\theta}^{\prime}}\pix(\mu_{t},\nu_{t};\eta)$}. Then \smash{$Q^{*}$} is a \textit{concave} function with respect to \smash{$\mu_{t}$}, and moreover takes on values strictly greater than \smash{$\bar{Q}$} at every vertex $m_{\theta}$:
\begin{equation}
\setlength{\abovedisplayskip}{3pt}
\setlength{\belowdisplayskip}{3pt}
\begin{split}
\forall m_{\theta}:
Q^{*}(m_{\theta},\nu_{t};\eta)
&>
\bar{Q}(m_{\theta},\nu_{t};\eta) \\
\end{split}
\end{equation}
Hence the \textit{termination set} $\mathcal{T}$ is the (disjoint) union of $|\Theta|$ \textit{convex} regions delimited by the intersection of $Q^{*}$ and $\bar{Q}$:
\begin{equation}
\setlength{\abovedisplayskip}{3pt}
\setlength{\belowdisplayskip}{3pt}
\begin{split}
\mathcal{T}(\eta)=\{\mu_{t}:Q^{*}(\mu_{t},\nu_{t};\eta)\geq\bar{Q}(\mu_{t},\nu_{t};\eta)\} \\
\end{split}
\end{equation}
and contains each of the simplex vertices. Finally, the (possibly null) \textit{continuation set} is its complement $\Delta(\Theta)\setminus\mathcal{T}$.
\end{proposition}
\proof
}
\Appendix

In the \textit{passive} setting (i.e. with a single acquisition choice), we are done. For \textit{active} sensing, the key question concerns \textit{which} acquisition to perform. Intuitively, we expect this choice to be made with respect to some notion of maximal information gain. Simultaneously, we also expect this to be balanced against the deadline risk associated with different acquisitions. Again, this tradeoff is critical: An aggressive, extensive email survey may be maximally informative for marketing outreach, but may also be most likely to cause recipients to unsubscribe from the campaign entirely; also recall our earlier examples on candidate tests and interviews.

\textbf{Negotiating Tradeoffs}.
The tradeoff between \textit{surprise} (i.e. information gain) and \textit{suspense} (i.e. riskiness of actions) has been explored in prior work on Bayesian reasoning in economics \citep{ely2015suspense} and in binary decisions with samples from a single continuous stream \citep{alaa2016balancing}. We first formalize these notions appropriately in the context of active sensing for timely decisions.

Two distinctions warrant attention: First, the informativeness of an acquisition must be \textit{timely} (i.e. arriving \textit{before} the deadline). Now, the classical definition for the informativeness of an acquisition simply measures the difference between the prior and expected posterior values for some appropriate measure of risk or uncertainty \citep{degroot1962uncertainty}. While this notion of surprise readily applies to the infinite-horizon setting \citep{naghshvar2013active}, here we only care about informativeness while the process is still alive: Realizing the correct decision \textit{after} the opportunity has closed carries no value for the original decision problem.

Second, here the riskiness of different choices of acquisitions is \textit{subjective} (i.e. weighted by an agent's preferences). Now, the notion of suspense is previously simply taken with respect to the the posterior survival probability \citep{alaa2016balancing}. In the presence of preferences, we now care about the \textit{preference-weighted} survival function.

Formally, define Markov operator $\mathbb{M}_{\lambda_{t}}$ for any appropriate measure of risk or uncertainty $U:\Delta(\Theta)\times\{0,1\}\rightarrow\mathbb{R}_{^{+}}$:
\begin{equation}
\setlength{\abovedisplayskip}{3.3pt}
\setlength{\belowdisplayskip}{4pt}
\begin{split}
&~~~~~~~~~~~~(\mathbb{M}_{\lambda_{t}}U)(\mu_{t},\nu_{t})
=
(\one-\nu_{t})
U(\mu_{t},\nu_{t})
+ \\[-2pt]
&\nu_{t}
\mathbb{E}_{p,q}[U(M(\lambda_{t},\mu_{t},\omega_{t+1}),\nu_{t+1})|\lambda_{t},\mu_{t},\nu_{t+1}=\one]
\end{split}
\end{equation}
capturing the expected posterior value of $U$ should the deadline not intercede, and simply returns the prior if the process were already dead. Then we naturally have the following:

(\textit{Timely}) \textbf{Surprise}. With respect to $U:\Delta(\Theta)\times\{0,1\}\rightarrow\mathbb{R}_{^{+}}$, the informativeness of any choice of acquisition $\lambda_{t}$ in reference to the statistic $(\mu_{t},\nu_{t})$ is given by the following:
\begin{equation}
\setlength{\abovedisplayskip}{3.3pt}
\setlength{\belowdisplayskip}{4pt}
\begin{split}
I_{t}(\lambda_{t})
&\doteq
U(\mu_{t},\nu_{t}) \\[-3pt]
&-(\one-\bar{p}_{\mu_{t},\lambda_{t-1}})(\mathbb{M}_{\lambda_{t}}U)(\mu_{t},\nu_{t})
\end{split}
\end{equation}
(\textit{Subjective}) \textbf{Suspense}. With respect the importance parameters $\eta$, the preference-weighted posterior probability of survival until after acquisition $\lambda_{t}$ is given by the following:
\begin{equation}
\setlength{\abovedisplayskip}{3.3pt}
\setlength{\belowdisplayskip}{4pt}
S_{t}(\lambda_{t})
\doteq
1
-
\frac{
{\textstyle\sum}_{\theta^{\prime}\in\Theta}\eta_{\text{\pix b},\theta^{\prime}}p_{\theta^{\prime},\lambda_{t}}\mu_{t}(\theta^{\prime})
}{
{\textstyle\sum}_{\theta^{\prime}\in\Theta}\eta_{\text{\pix b},\theta^{\prime}}\mu_{t}(\theta^{\prime})
}
\end{equation}
These generalized notions of surprise and suspense simply inherit existing definitions while accounting for the presence of time pressure and subjective preferences. In particular, setting $p_{\theta,\lambda}=0$ recovers the original (infinite-horizon) definition of information gain \citep{naghshvar2013active}, whence the optimal strategy simply greedily maximizes information. Likewise, setting $\eta_{\text{\pix b},\theta}=1$ recovers the original (preference-free) definition of suspense \citep{alaa2016balancing}.

\newcolumntype{F}{>{\centering\arraybackslash}m{3.6cm}}
\newcolumntype{G}{>{\centering\arraybackslash}m{4.1cm}}
\newcolumntype{H}{>{\centering\arraybackslash}m{3.5cm}}
\newcolumntype{I}{>{\centering\arraybackslash}m{1.8cm}}
\newcolumntype{J}{>{\centering\arraybackslash}m{1.5cm}}
\newcolumntype{K}{>{\centering\arraybackslash}m{1.8cm}}

\begin{table*}[t]\small
\setlength\tabcolsep{2.0pt}
\renewcommand{\arraystretch}{1.0}
\vspace{-0.8em}
\caption{\textit{Examples of inverse optimization problems with respect to various underlying settings}. A generic formulation is shown for each example category.
$^{1}$\pix\citet{barmann2017emulating},
$^{2}$\pix\citet{dempe2006inverse},
$^{3}$\pix\citet{dong2018generalized},
$^{4}$\pix\citet{keshavarz2011imputing},
$^{5}$\pix\citet{heuberger2004inverse},
$^{6}$\pix\citet{krumke1998approximation},
$^{7}$\pix\citet{yang1999two},
$^{8}$\pix\citet{ahmadian2018algorithms},
$^{9}$\pix\citet{abbeel2004apprenticeship},
$^{10}$\pix\citet{ziebart2008maximum}.
Notation: dimensions $m,n,k$ indicate arbitrary, problem-dependent dimensions; $G,D$ respectively denote graphs and digraphs with vertices $V$ and edges $E$; $\mathcal{S}_{G}$ is the set of spanning trees of $G$ and $\mathcal{P}_{D}$ is the set of paths in $D$; $\phi$ are known basis features, and $w$ their respective weights.}
\label{tab:related_inverse}
\begin{center}
\begin{tabular}{FGHIJK}
\toprule
\textbf{Framework} & {Problem Class} & {Objective Function} & {Signal} & {Parameter} & {Response} \\
\midrule
B{\"a}rmann,$^{1}$ Dempe,$^{2}$ etc.
& Inverse Linear Optimization
& $b_{1}^{\top}x,Ax=b_{2},x\geq0$
& $A\in\mathbb{R}^{m\times n}$
& $b\in\mathbb{R}^{m+n}$
& $x\in\mathbb{R}^{n}$
\\
Dong,$^{3}$ Keshavarz,$^{4}$ etc.
& Inverse Convex Optimization
& $f(a,b,x),g(a,b,x)\leq0$
& $a\in\mathbb{R}^{m}$
& $b\in\mathbb{R}^{k}$
& $x\in\mathbb{R}^{n}$
\\
Heuberger,$^{5}$ Krumke,$^{6}$ etc.
& Inverse Minimum Spanning Tree
& $c^{\top}A_{\psi}$, $\psi\in\mathcal{S}_{G}$
& $G=\langle V,E\rangle$
& $c\in\mathbb{R}^{E}$
& $\psi\subseteq E$
\\
\raisebox{-0.15em}{Yang,$^{7}$ Ahmadian,$^{8}$ etc.}
& \raisebox{-0.15em}{Inverse Integral Shortest Paths}
& $c^{\top}A_{\xi}$, $\xi\in\mathcal{P}_{D}$
& \raisebox{-0.15em}{$D$; $s,t\in V$}
& \raisebox{-0.15em}{$c\in\mathbb{R}^{E}$}
& \raisebox{-0.15em}{$\xi_{(s,t)}\subseteq E$}
\\
\raisebox{-0.10em}{Abbeel,$^{9}$ Ziebart,$^{10}$ etc.}
& \raisebox{-0.10em}{Inverse Reinforcement Learning}
& $\sum_{t=0}^{\infty}\mathbb{E}[\gamma^{t}w^{\top}\phi(S_{t})]$
& \raisebox{-0.30em}{$S_{0:T}$}
& \raisebox{-0.30em}{$w\in\mathbb{R}^{|\Phi|}$}
& \raisebox{-0.30em}{$A_{0:T}$}
\\
\textbf{(Ours)}
& Inverse Active Sensing
& $\mathbb{E}[\ell($\scalebox{1.05}{$\lambda_{0:\tau-1}$}$,$\pix\scalebox{1.2}{$\tau$}$,\hat{\theta};\eta)]$
& \raisebox{ 0.15em}{$\langle$\scalebox{1.05}{$\omega_{1:\tau-1},\mu_{0}$}$\rangle$}
& \raisebox{-0.05em}{$\eta,\rho$\pix$\in$\pix$\mathcal{H}$$\times$$\mathbb{R}$}
& \raisebox{ 0.15em}{$\langle$\scalebox{1.05}{$\lambda_{0:\tau-1}$}$,\pix$\scalebox{1.2}{$\tau$}$,\hat{\theta}\rangle$}
\\
\bottomrule
\end{tabular}
\end{center}
\vspace{-1.9em}
\end{table*}

We can now formally characterize the risk-benefit tradeoff, highlighting the expressivity of our framework. It turns out that the optimal acquisition $\lambda_{t}^{*}$ (i.e per the optimal strategy) naturally strikes a balance between surprise and suspense:

\vspace{-3pt}
\Copy{thm_surprise_suspense}{
\begin{proposition}[Surprise and Suspense]\upshape
\label{thm:surprise_suspense}
When $\mu_{t}$\pix$\not\in$\pix$\mathcal{T}(\eta)$, the optimal acquisition directly trades off surprise and su- spense (in addition to the immediate cost of acquisition):
\begin{equation}
\setlength{\abovedisplayskip}{2.5pt}
\setlength{\belowdisplayskip}{3pt}
\begin{split}
\lambda_{t}^{*}
&=
\textstyle\arg\sup_{\lambda_{t}\in\Lambda}
h(I_{t}(\lambda_{t}),S_{t}(\lambda_{t}))
-
\eta_{\text{\pix c},\lambda_{t}}
c_{\lambda_{t}}
\end{split}
\end{equation}
where $h$ is increasing in $I_{t}(\lambda_{t})$ and $S_{t}(\lambda_{t})$, and the uncertainty function for the information gain is taken as $U=V^{*}$.
\end{proposition}
\proof
}
\Appendix

Depending on an individual's preferences, this tradeoff automatically expresses a range of behaviors from ``surprise-optimal'' to ``suspense-optimal''. Note that this is absent in passive settings such as \citet{frazier2008sequential} and \citet{dayanik2013reward}\textemdash i.e. no choice of surprise; it is also absent in settings with no deadline risk, such as \citet{ahmad2013active} and \citet{naghshvar2013active}\textemdash i.e. no element of suspense.

Finally, for completeness we also state the optimal decision $\hat{\theta}^{*}$ when $\mu_{t}$\pix$\in$\pix$\mathcal{T}(\eta)$. Immediately from Equation {\color{mydarkblue}12} for $\bar{Q}$:
\begin{equation}
\setlength{\abovedisplayskip}{2.5pt}
\setlength{\belowdisplayskip}{2pt}
\begin{split}
\hat{\theta}^{*}
&=
\textstyle\arg\sup_{\hat{\theta}\in\Theta}
\eta_{\text{\pix a},\hat{\theta}}\mu_{t}(\hat{\theta})
\end{split}
\end{equation}
Together, this (decision) rule and (acquisition) rule of Proposition \ref{thm:surprise_suspense} fully identify the optimal active sensing strategy.

\vspace{-0.5em}
\subsection{Inverse Active Sensing}\label{sec:inverse_problem}

In the opposite direction, the inverse active sensing {\small(``IAS'')} problem translates (observed) behavior back to (unobserved) preferences. We should be precise with semantics: Neither are we \textit{assuming} that real-world agents indeed act a certain way, nor are we \textit{prescribing} that they act optimally. Instead, our objective is thoroughly \textit{descriptive}: Based on an agent's behavior, what do they appear (ceteris paribus) to effectively prioritize? For instance, what diseases are more important to diagnose correctly, and which tests are being over-prescribed? While actual behavior may not be induced (explicitly) by conscious optimization with respect to preferences (e.g. bounded rationality), we wish to understand them in terms of what is\textemdash in effect\textemdash prioritized (implicitly).

\textbf{Inverse Optimization}.
We approach {\small IAS} from an inverse optimization {\small(``IO'')} perspective with multiple observations. Broadly, {\small IO} deals with finding an objective function to best explain a set of observations \citep{barmann2017emulating, dong2018generalized}, and is applicable to a wide variety of underlying problems. Specifically, an objective is determined by a (fixed) \textit{parameter} and a (variable) \textit{signal}. Different signals induce different instances of the optimization problem, hence different \textit{responses} (i.e. solutions) from an optimizing agent. Given a collection of (observed) signal-response pairs, we seek to infer the (unobserved) parameter. Table \ref{tab:related_inverse} sets out classic examples of this paradigm\textemdash along with {\small IAS}.

\vspace{-0.3em}
Three considerations drive our approach:

\vspace{-1.0em}
\begin{enumerate}[leftmargin=*]
\itemsep-1pt
\item \textbf{Uncertainty}. Almost always, more than one configuration will accord with observed behavior. In other settings where the focus is typically on the would-be \textit{performance} of acting under the recovered objective, this is hardly an issue. In contrast, our focus is on \textit{understanding} drivers of behavior, so extracting a single configuration is of limited utility. Ideally, we wish to work with a distribution.
\item \textbf{Bounded Rationality}. Instead of conditioning purely on Bayes-optimal strategies, we may additionally consider other well-studied objectives in behavioral literature\textemdash such as greedy look-ahead \citep{najemnik2005optimal} or information-maximizing \citep{butko2010infomax} strategies. We may even wish to compare how well different classes of strategies describe observed behavior.
\item \textbf{Imperfect Response}. Even if we concede a known class of strategies, observed responses are often imperfect due to compliance, measurement noise, implementation error, and model uncertainty \citep{aswani2018inverse, esfahani2018data}. We would like to account (probabilistically) for the fact that the observed response to a signal may deviate from the perfect (i.e. objective-maximizing) choice.
\end{enumerate}
\vspace{-0.75em}

\textbf{Posterior Inference}.
We consider a \textit{Bayesian} approach to {\small IAS}; this accommodates [1]. Compacting notation, first let $\tilde{\lambda}_{t}\in\Lambda\cup\Omega$ denote either $\lambda_{t}$ (prior to stopping) or $\hat{\theta}\in\Omega$. Likewise, let $\tilde{\omega}_{t}\in\Omega\cup\{\nil\}$ indicate $\omega_{t}$ or $\omega_{\tau}=\nil$. Then
\begin{equation}
\setlength{\abovedisplayskip}{3pt}
\setlength{\belowdisplayskip}{2pt}
\mathcal{D}\doteq\{(\tilde{\lambda}_{n,t},\tilde{\omega}_{n,t+1})_{t=0}^{\tau_{n}-1}\}_{n=1}^{N}
\end{equation}
denotes a collection of acquisitions and outcomes, with decision episodes indexed $n\in\{1,...,N\}$.
Let $\pi$ be drawn from some prior $\mathbb{P}\{\pi\}$ over the space of strategies $\mathcal{P}$. Then
\begin{equation}
\setlength{\abovedisplayskip}{3pt}
\setlength{\belowdisplayskip}{2pt}
\begin{split}
\mathbb{P}_{p,q}\{\pi|\dataset\}
\scalebox{1}{\pix=\pix}
\tfrac{
\mathbb{P}_{p,q}\{\dataset|\pi\}
\mathbb{P}\{\pi\}
}{
{\textstyle\int}_{\mathcal{P}}
\mathbb{P}_{p,q}\{\dataset|\pi\}
d\mathbb{P}\{\pi\}
}
\end{split}
\end{equation}
is the posterior over strategies\textemdash given the observed decision behavior. Next, we specify what each strategy $\pi$ consists in.

\textbf{Behavioral Strategies}.
Most of the time, what we aim to do is inverse \textit{optimal} active sensing (cf. \mbox{inverse \textit{optimization}}) \textemdash that is, to locate preferences most consistent with an agent making Bayes-optimal acquisitions. But what if we want to accommodate a different criterion? Humans may act myopically (e.g. greedy lookahead), pursue approximate goals (e.g. infomax), or simply follow rulebooks (e.g. attribute-wise scoring schemes common in hiring and medical settings). For this, we need the notion of \textit{generalized} $Q$-factors.

Let a strategy be characterized by its set of (not \mbox{necessarily} Bayes-optimal) factors \smash{$\{Q^{\kappa}_{\lambda}$\pix$:$\pix$\Delta(\Theta)$$\times$$\{0,1\}$$\rightarrow$$\mathbb{R}_{^{+}}\hspace{-2pt}\}_{\lambda\in\Lambda}$}, wh- ere $\kappa$ identifies the sensing criterion; this accommodates [2]. Denote with \smash{$\tilde{Q}\raisebox{2pt}{$^{\kappa}_{\tilde{\lambda}}$}$} either $Q^{\kappa}_{\lambda}$ or $\bar{Q}_{\hat{\theta}}$; so \smash{$\tilde{Q}\raisebox{2pt}{$^{\kappa}_{\tilde{\lambda}_{t}}$}(\mu_{t},\nu_{t};\eta)$} encodes the \textit{desirability} of \smash{$\tilde{\lambda}_{t}$} under $\mu_{t},\nu_{t}$ given $\eta$. For instance, the greedy look-ahead criterion {\small (``GL'')} simply corresponds to:
\begin{equation}
\setlength{\abovedisplayskip}{5pt}
\setlength{\belowdisplayskip}{6pt}
\begin{split}
Q_{\lambda_{t}}^{\text{GL}}(\mu_{t},\nu_{t};\eta)
&\doteq
\eta_{\text{\pix c},\lambda_{t}}
c_{\lambda_{t}}
+
g(\eta_{\pix\text{d}})
+~\\[-2pt]
&~~~~~\mathbb{E}_{p,q}[
\bar{Q}(\mu_{t+1},\nu_{t+1};\eta)
|\lambda_{t},\mu_{t},\nu_{t}] \\
\end{split}
\end{equation}
where $g$ is some function of decision-threshold parameters \smash{$\eta_{\pix\text{d}}$\pix$\in$\pix$\mathbb{R}\raisebox{-2pt}{$_{^{+}}^{|\Theta|}$}$}, and here \smash{$\eta\doteq(\eta_{\text{\pix b}},\eta_{\text{\pix c}},\eta_{\text{\pix d}})$}. As an arbitrary functional of $\eta$, we can likewise encode any criteria of choice, such as infomax \smash{$Q_{\lambda}^{\text{IM}}$} by way of (weighted) entropy, and (of course) the Bayes-optimal case \smash{$Q_{\lambda}^{*}$}. Given some \smash{$\tilde{Q}^{\kappa}\doteq\{\tilde{Q}\raisebox{1.5pt}{$_{\tilde{\lambda}}^{\kappa}$}\}\raisebox{1.5pt}{$_{\tilde{\lambda}\in\Lambda\cup\Theta}$}$}, we consider the \textit{Boltzmann} behavioral strategy with inverse temperature $\rho$; this accommodates [3]:
\vspace{-1.0em}
\begin{equation}
\setlength{\abovedisplayskip}{6pt}
\setlength{\belowdisplayskip}{6pt}
\pi^{\kappa}_{\rho}(\tilde{\lambda}_{t}|\mu_{t},\nu_{t};\eta)
\scalebox{1}{$\pix\pix\doteq\pix\pix$}
\frac{\scalebox{0.94}{$\exp$}(\scalebox{0.85}{$-$}\rho \tilde{Q}\raisebox{1.5pt}{$^{\kappa}_{\tilde{\lambda}_{t}}$}(\mu_{t},\nu_{t};\eta))}{\scalebox{0.85}{$\sum$}_{\tilde{\lambda}^{\prime}_{t}\in\Lambda\cup\Theta}\scalebox{0.94}{$\exp$}(\scalebox{0.85}{$-$}\rho \tilde{Q}\raisebox{1.5pt}{$^{\kappa}_{\tilde{\lambda}^{\prime}_{t}}$}(\mu_{t},\nu_{t};\eta))}
\end{equation}
Formally, then, a strategy $\pi$ is specified by $(\kappa,\eta,\rho)$. If we restrict our attention to known classes \smash{$\kappa\in\{\text{\small GL},\text{\small IM},*,...\}$}, then the prior $\mathbb{P}\{\pi\}$ is equivalently captured by $\mathbb{P}\{\kappa,\eta,\rho\}$ $=$ $\mathbb{P}\{\kappa\}\mathbb{P}\{\eta|\kappa\}\mathbb{P}\{\rho\}$. (The conditioning on $\kappa$ accommodates a single global space of preference weights for different $\kappa$).

Now, $\mathcal{D}$ depends on the dynamics of both the decision problem and recognition model. However, the latter involves no uncertainty (we impose a Bayesian recognition model), and the former simply drops out when evaluating the posterior:

\vspace{-3pt}
\Copy{thm_bayesian_inference}{
\begin{proposition}[Strategy Posterior]\upshape\label{thm:strategy_posterior}
The posterior $\mathbb{P}\{\pi|\dataset\}$ over $\mathcal{P}$ (Equation {\color{mydarkblue}22}) satisfies the following proportionality:\\
\begin{equation}
\setlength{\abovedisplayskip}{-8pt}
\setlength{\belowdisplayskip}{6pt}
\begin{split}
\mathbb{P}\{\pi_{\rho}^{\kappa}\text{\small(}...;\eta\text{\small)}|\dataset\}
&\propto
\mathbb{P}\{\kappa\}\mathbb{P}\{\eta|\kappa\}\mathbb{P}\{\rho\} \\[-2pt]
&\pix\cdot
{\textstyle\prod}_{n=1}^{N}
{\textstyle\prod}_{t=0}^{\tau_{n}-1}
\pi_{\rho}^{\kappa}(\tilde{\lambda}_{n,t}|\mu_{n,t},\nu_{n,t};\eta) \\
\end{split}
\end{equation}
where $\mu_{n,t}$ is recursively computed via update $M$, $\nu_{n,t}$$=$\pix$\one$ prior to stopping, and $\pi_{\rho}^{\kappa}\text{\small(}...;\eta\text{\small)}$ is defined as in Equation {\color{mydarkblue}24}.
\end{proposition}
\proof
}
\Appendix

Note that setting $\mathbb{P}\{\kappa\}$ as the Dirac delta centered on the Bayes-optimal criterion recovers the case of inverse \textit{optimal} active sensing; this is the formulation presented in Table \ref{tab:related_inverse}.

\textbf{Maximum A Posteriori}.
Seeking the {\small MAP} estimate effectively reduces {\small IAS} to a (posterior) optimization problem. Let \smash{$\mathcal{H}$\pix$=$\pix$\mathbb{R}^d$} denote the $d$-dimensional (global) space of $\eta$, and $\mathcal{K}$ the space of $\kappa$. Then the {\small MAP} estimate is given as follows, where $\text{LSE}\pix$\raisebox{1pt}{[}$\pix\cdot\pix$\raisebox{1pt}{]} denotes logsumexp and $\mathcal{P}$ $\doteq$ $\mathcal{K}$\pix\pix$\times$\pix\pix$\mathcal{H}$\pix$\pix\times$\pix\pix$\mathbb{R}$:
\begin{equation}
\setlength{\abovedisplayskip}{6pt}
\setlength{\belowdisplayskip}{6pt}
\begin{split}
{\displaystyle\argmax_{(\kappa,\eta,\rho)\in\mathcal{P}}}
\big\{
\log\mathbb{P}\{k\}
-
\log\mathbb{P}\{\eta&\pix|\kappa\}
-
\log\mathbb{P}\{\rho\} \\[-9pt]
~-
{\textstyle\sum}_{n=1}^{N}\pix
{\textstyle\sum}_{t=0}^{\tau_{n}-1}\big(\pix
\rho \tilde{Q}&\raisebox{1.5pt}{$^{\kappa}_{\tilde{\lambda}_{n,t}}$}(\mu_{n,t},\nu_{n,t};\eta) \\[-2pt]
~+
\text{LSE}_{\tilde{\lambda}^{\prime}_{n,t}\in\Lambda\cup\Theta}
[\pix-\pix\rho \tilde{Q}&\raisebox{1.5pt}{$^{\kappa}_{\tilde{\lambda}_{n,t}}$}(\mu_{n,t},\nu_{n,t};\eta)\pix]
\pix\big)\big\}
\end{split}
\end{equation}
Given some (finite) set of known $\kappa$'s, we can simply compute the {\small MAP} for each (over $\mathcal{H}\times\mathbb{R}$), then compare over $\mathcal{K}$ using $\mathbb{P}\{\kappa\}$. This can be done via standard gradient methods or numerical optimization. Using Bayes-optimal strategies, it is easy to show differentiability with respect to $\eta$ and $\rho$:

\vspace{-3pt}
\Copy{thm_diff_posterior}{
\begin{proposition}[Differentiable Posterior]\upshape\label{thm:differentiable_posterior}
Assuming diffe- rentiable priors $\mathbb{P}\{\eta\pix|*\},\mathbb{P}\{\rho\}$, the posterior $\mathbb{P}\{\eta,\rho|*,\dataset\}$ for optimal strategies is differentiable (almost everywhere).
\end{proposition}
\proof
}
\Appendix

\textbf{Sampling from Posterior}.
Instead of a point estimate, we can generate samples from the posterior for each $\kappa$. {\small MCMC} sampling is common in inverse problem settings \citep{ye2019optimization, bardsley2012mcmc, ramachandran2007bayesian}. We perform geometric random walks over coordinates of a lattice in $\mathcal{H}\times\mathbb{R}$ \citep{frieze1994sampling, applegate1990sampling} to yield samples from posterior $\mathbb{P}\{\eta,\rho|\kappa,\dataset\}$ \mbox{(Algorithm {\color{mydarkblue}1})}. Consider a discrete subset $\mathcal{L}$ of $\mathbb{R}^{d+1}$ comprising coordinates that are integer multiples of a chosen resolution. The algorithm simply tries to move to one of its neighbors $\mathcal{N}$ at each step, with acceptance ratios determined by posteriors.

\vspace{-3pt}
\rule{\linewidth}{0.75pt}\vspace{-2pt}
\textbf{Algorithm 1}\pix\pix~Posterior Sampler for IAS
\vspace{-7.5pt}\\
\rule{\linewidth}{0.5pt}\vspace{-15pt}
\def\NoNumber#1{{\def\alglinenumber##1{}\State #1}\addtocounter{ALG@line}{-1}}
\begin{algorithmic}[1]
\setstretch{1.02}
\STATE {\bfseries Input}: Decision behavior $\mathcal{D}$ and priors $\mathbb{P}\{\eta\pix|\kappa\}$, $\mathbb{P}\{\rho\}$
\STATE Randomly select $(\eta,\rho)_{0}\in\mathcal{L}$
\STATE $\tilde{Q}_{0}\leftarrow\texttt{ActiveSensing}(\kappa,\eta_{\pix0})$
\FOR{$i=1,...$}
\STATE Randomly select $(\eta,\rho)^{\prime}\in\mathcal{N}((\eta,\rho)_{i-1})$ \COMMENT{\small neighbor}
\STATE $\tilde{Q}^{\prime}\leftarrow\texttt{ActiveSensing}(\kappa,\eta^{\prime})$
\STATE $R\leftarrow\mathbb{P}\{(\eta,\rho)^{\prime}|\kappa,\dataset\}/\mathbb{P}\{(\eta,\rho)_{i-1}|\kappa,\dataset\}$
\STATE \textbf{w.\pix p.} $\min\{1,R\}$ \textbf{do} $(\eta,\rho)_{i}\leftarrow(\eta,\rho)^{\prime}$\COMMENT{\small accept\pix\pix\pix}
\STATE \textbf{otherwise} $(\eta,\rho)_{i}\leftarrow(\eta,\rho)_{i-1}$\COMMENT{\small \pix\pix reject\pix~~~~}
\ENDFOR
\STATE {\bfseries Output}: Estimate of posterior $\hat{\mathbb{P}}\{\eta,\rho|\kappa,\dataset\}$
\end{algorithmic}
\vspace{-1.45em}
\rule{\linewidth}{0.5pt}

\vfil\penalty-200\vfilneg

\begin{figure}[t]
\centering
\vspace{-0.2em}
\includegraphics[width=\linewidth]{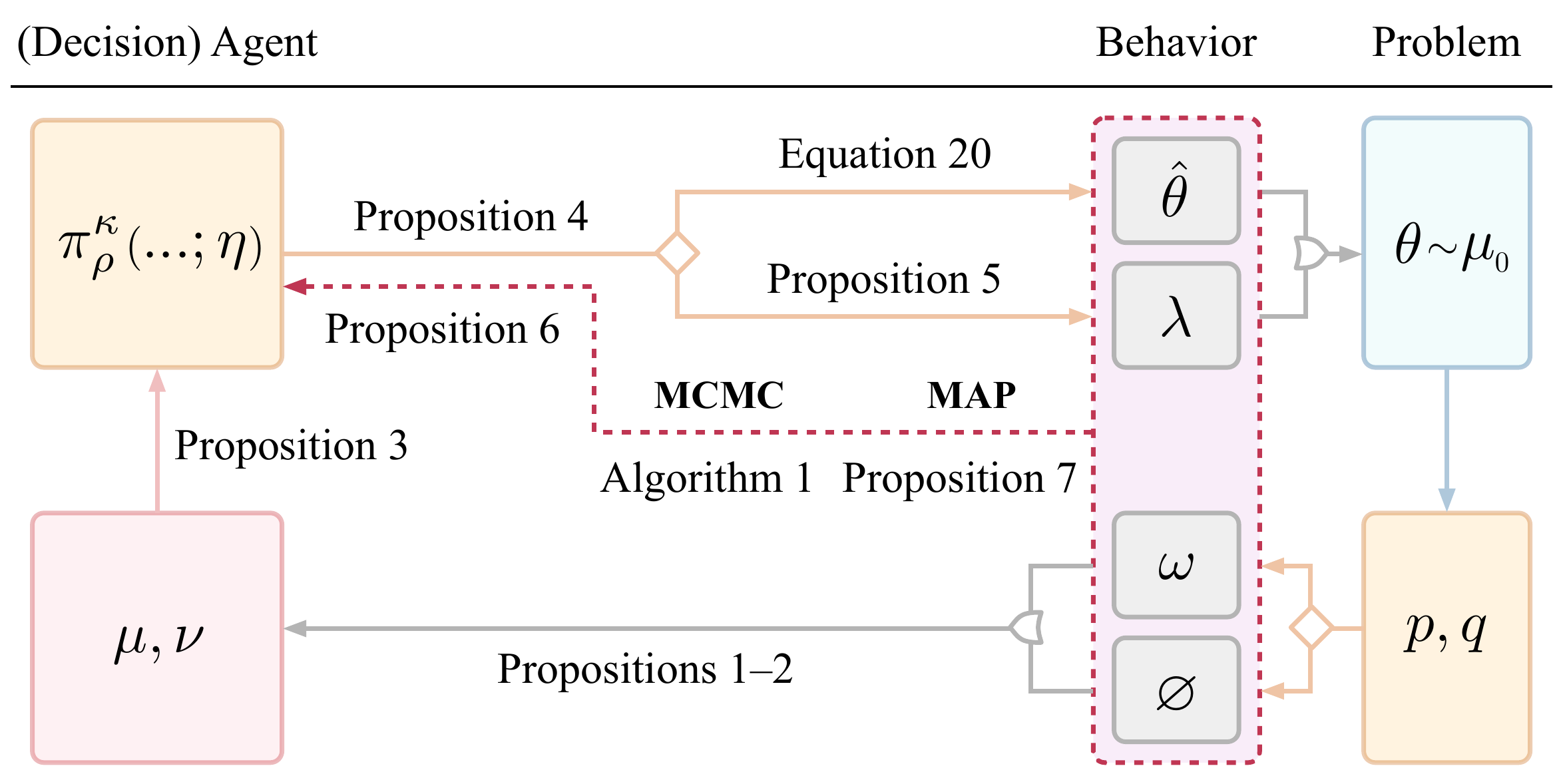}
\vspace{-1.8em}
\caption{\small \textit{Active sensing and inverse active sensing (redux)}. Forward direction shown in clockwise solid, and inverse in purple dashed. Using a Bayesian recognition model \mbox{(Propositions \ref{thm:sufficient_statistic}--\ref{thm:active_passive})}, the optimal value is computable (Proposition \ref{thm:optimal_value}) via dynamic prog- ramming to obtain the optimal strategy (Propositions \ref{thm:continue_terminate}--\ref{thm:surprise_suspense}, Equation {\color{mydarkblue}20}). In IAS, the strategy can be inferred (Proposition \ref{thm:strategy_posterior}) via \scalebox{0.9}{MAP} estimation (Proposition \ref{thm:differentiable_posterior}) or via \scalebox{0.9}{MCMC} sampling (Algorithm {\color{mydarkblue}1}).}
\label{fig:pull2}
\vspace{-1.0em}
\end{figure}

In a nutshell, we summarize the entire framework by harking back to Figure \ref{fig:pull}. We now have all the tools for {\small IAS}: Figure \ref{fig:pull2} shows a map of our key developments in both the forwa- \mbox{rd (clockwise solid)} and inverse (purple dashed) directions.

\begin{figure*}[t]
\vspace{-4.25em}
\centering
\subfloat[$Q$-factors for Decisions]
{\includegraphics[width=0.24\linewidth, trim=15em 5.5em 23em 40em]{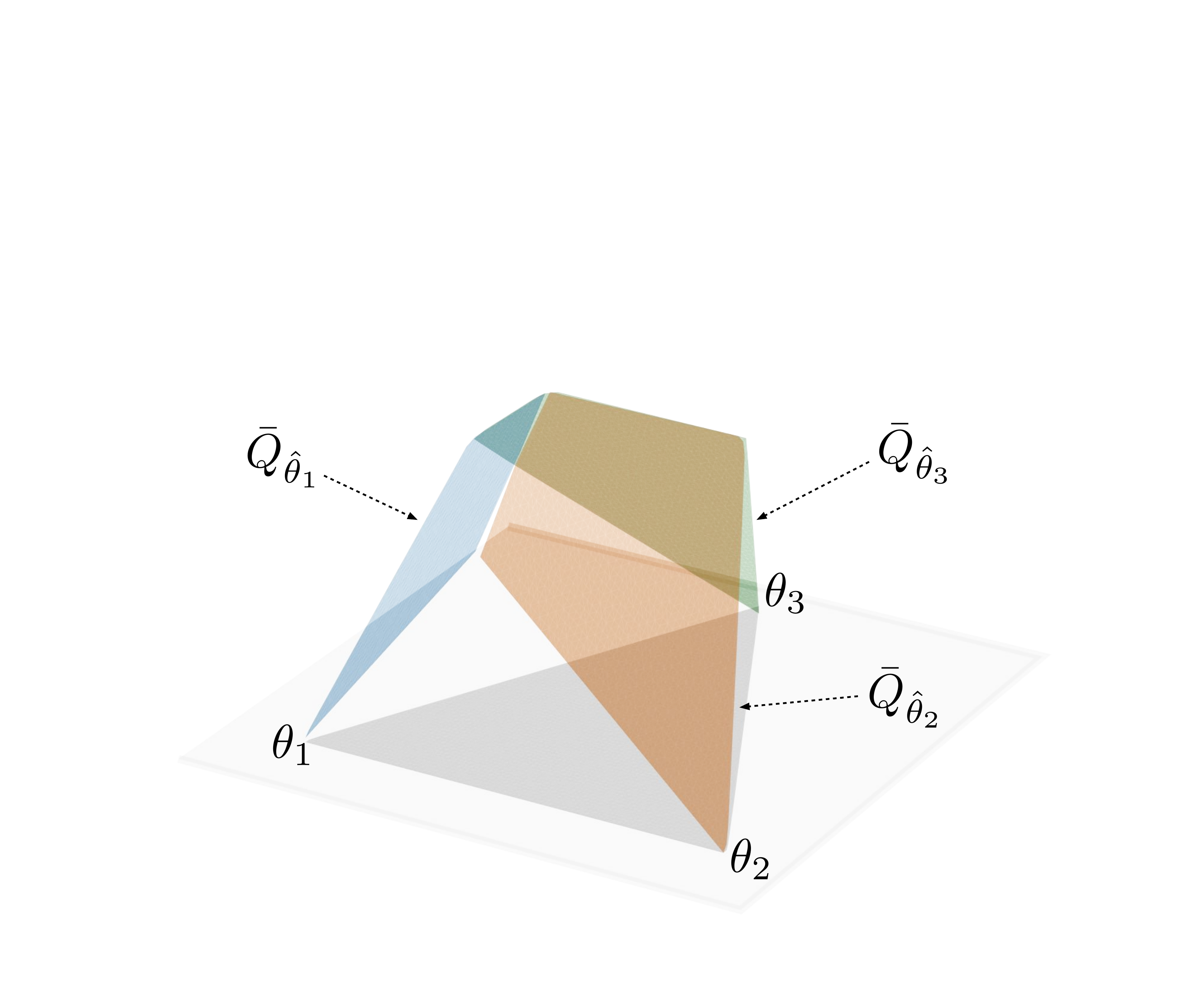}}
\hfill
\subfloat[Intersection of $Q^{*}$ and $\bar{Q}$]
{\includegraphics[width=0.24\linewidth, trim=15em 5.5em 23em 40em]{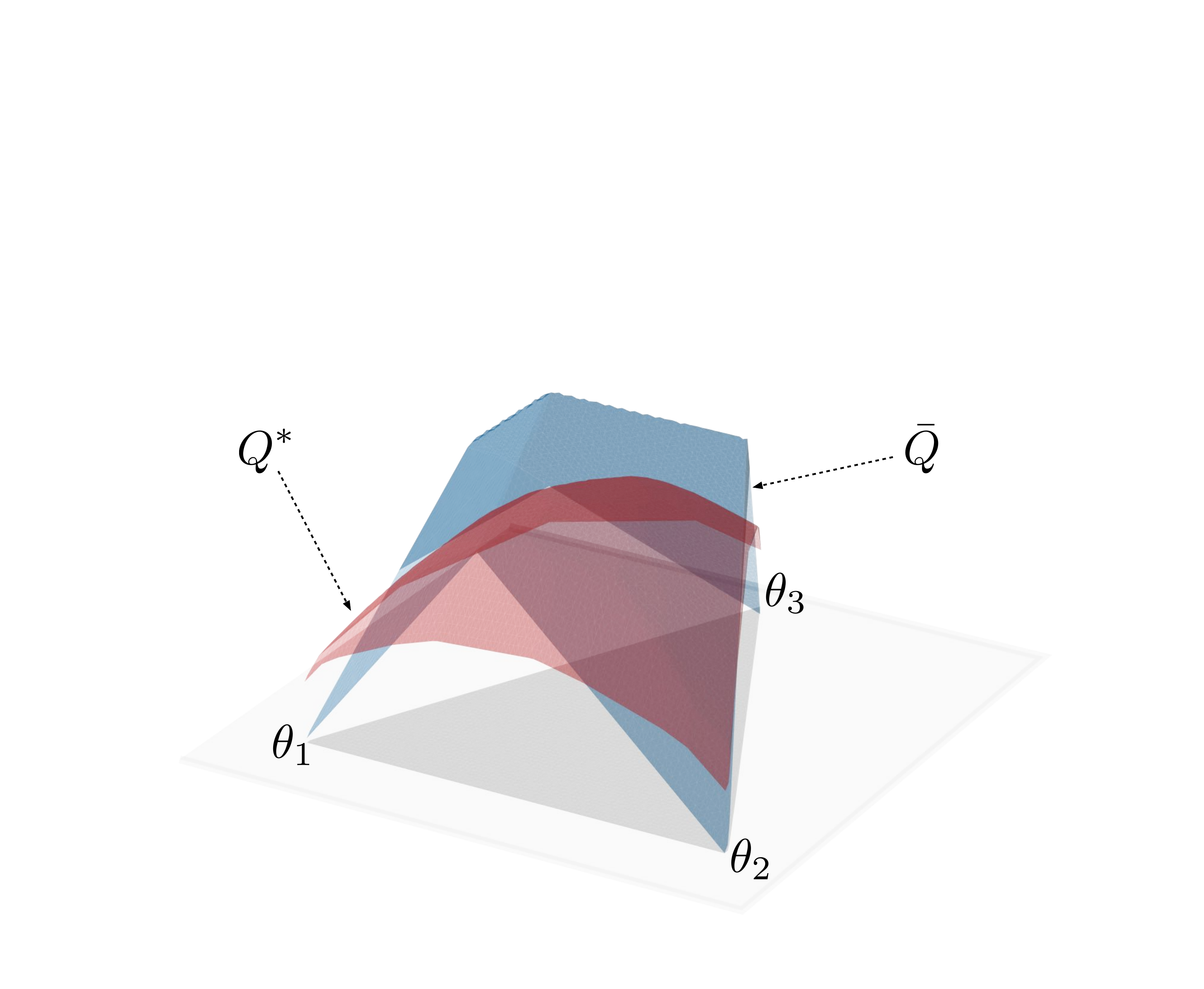}}
\hfill
\subfloat[Continue vs. Terminate]
{\includegraphics[width=0.23\linewidth, trim=7em 8.5em 7em -20em]{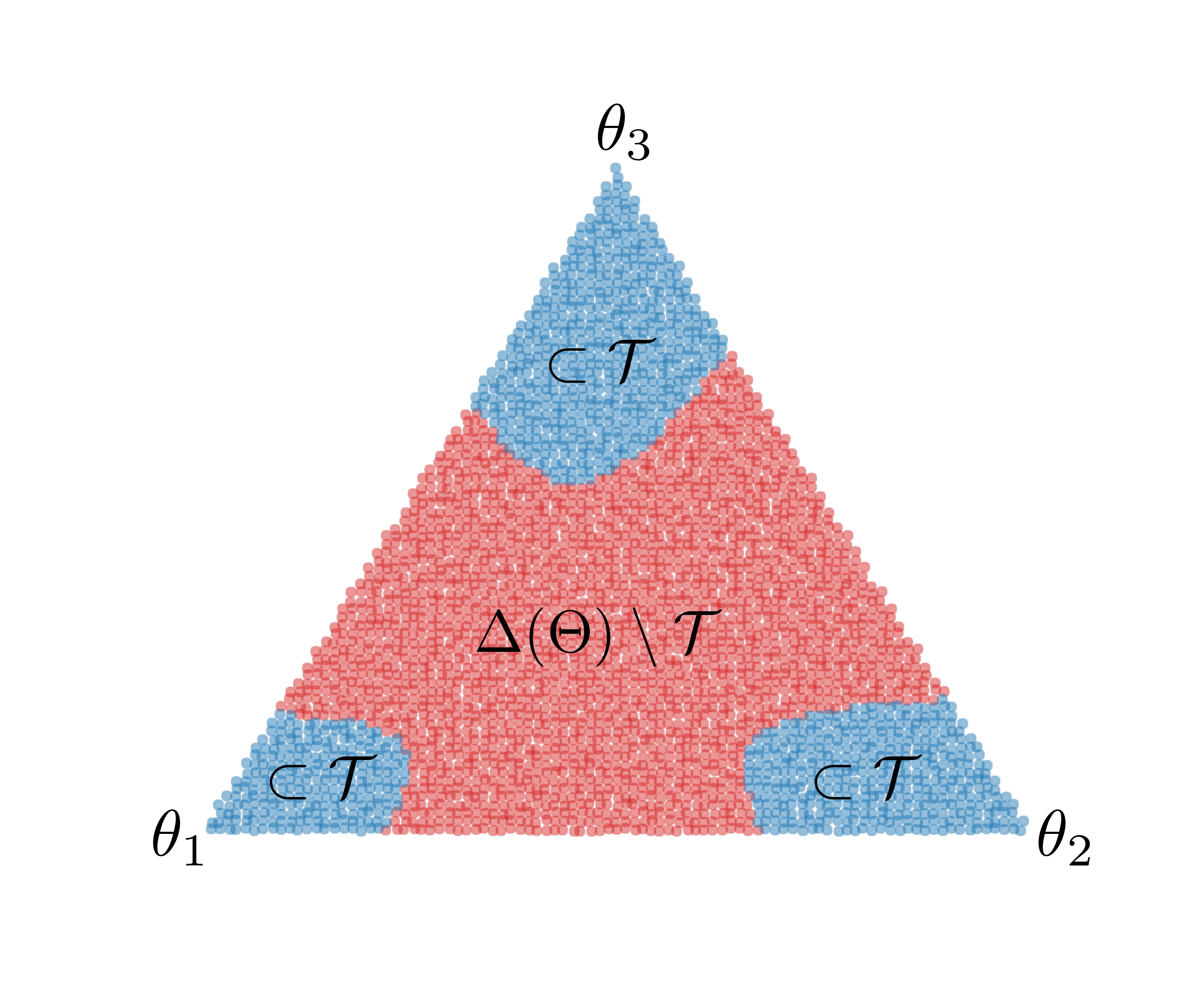}}
\hfill
\subfloat[Acquisitions vs. Decisions]
{\includegraphics[width=0.23\linewidth, trim=7em 8.5em 7em -20em]{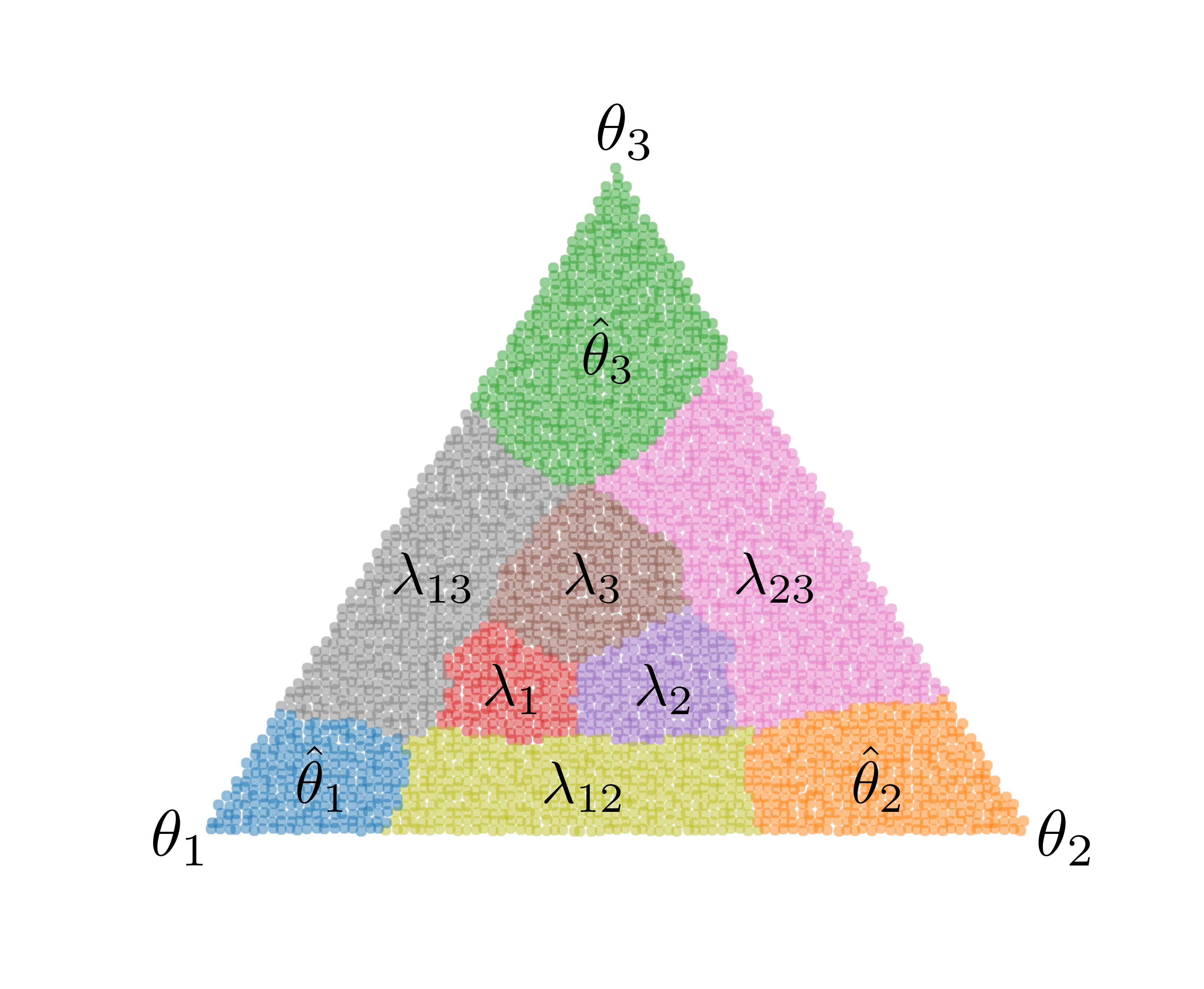}}
\vspace{-0.5em}
\caption{\small \textit{Optimal active sensing}. Posterior simplex (before deadline) for Example \ref{eg:1}. (a) Aggregate $Q$-factor for decisions, cf. Equation {\color{mydarkblue}20}. (b) Intersection of aggregate $Q$-factors for acquisitions and decisions, cf. Proposition \ref{thm:continue_terminate}. (c) Continuation and termination sets; the latter comprises convex regions around vertices, where $Q^{*}\geq\bar{Q}$; the former is its complement. (d) Complete strategy map; continuation and termination sets are partitioned into individual acquisition and decision regions, cf. Proposition \ref{thm:surprise_suspense}, Equation {\color{mydarkblue}20} (\textit{continued on page \hyperlink{page.9}{9}}).}
\label{fig:plex}
\vspace{-1.0em}
\end{figure*}

\begin{figure*}[t]
\vspace{0.6em}
\centering
\subfloat[Recovering Distribution of Preferences]
{\includegraphics[width=0.32\linewidth, trim=4em 0.5em 4em  4em]{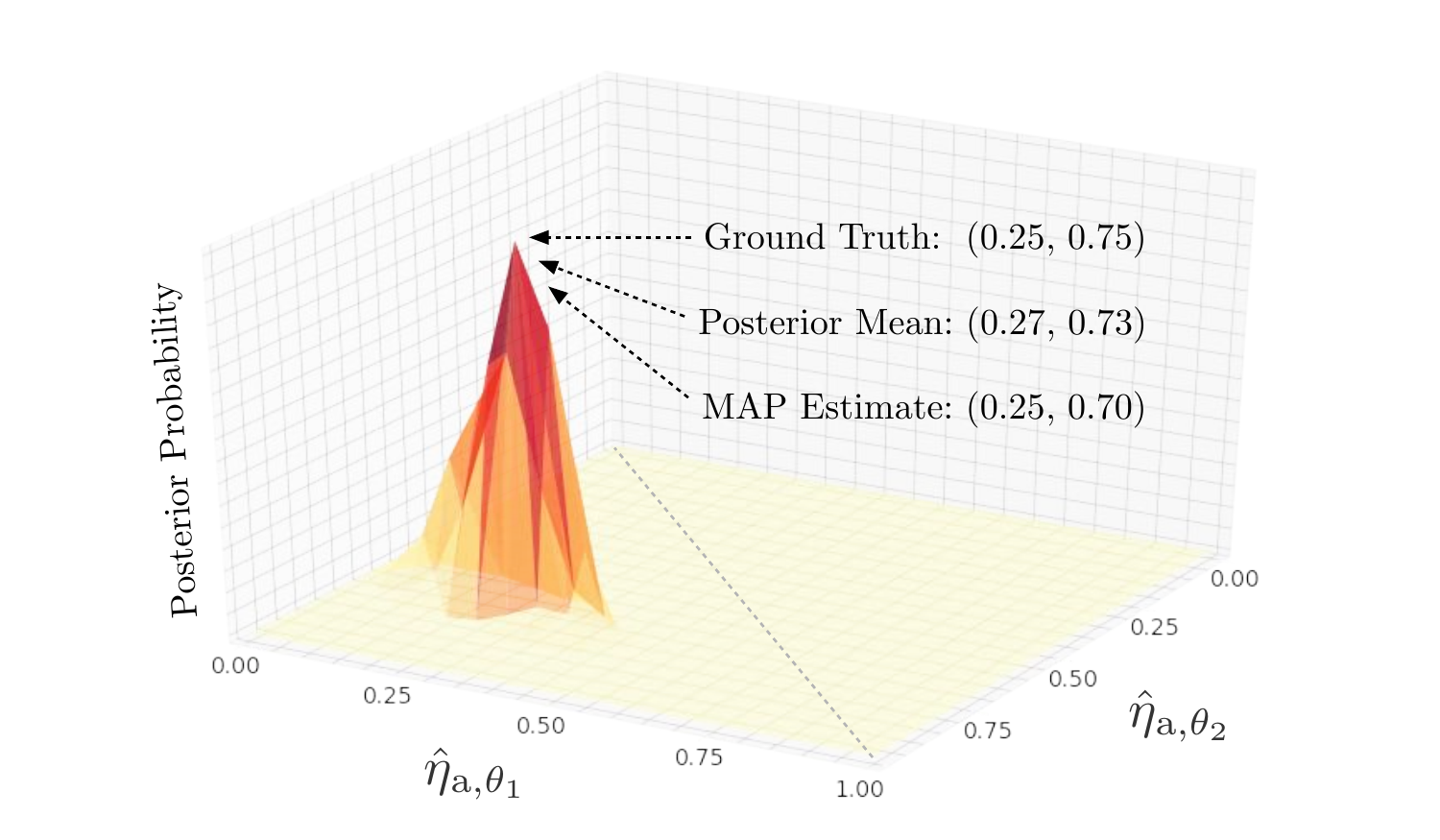}}
\hfill
\subfloat[Comparing Distributions of Preferences]
{\includegraphics[width=0.32\linewidth, trim=4em 0.5em 4em  4em]{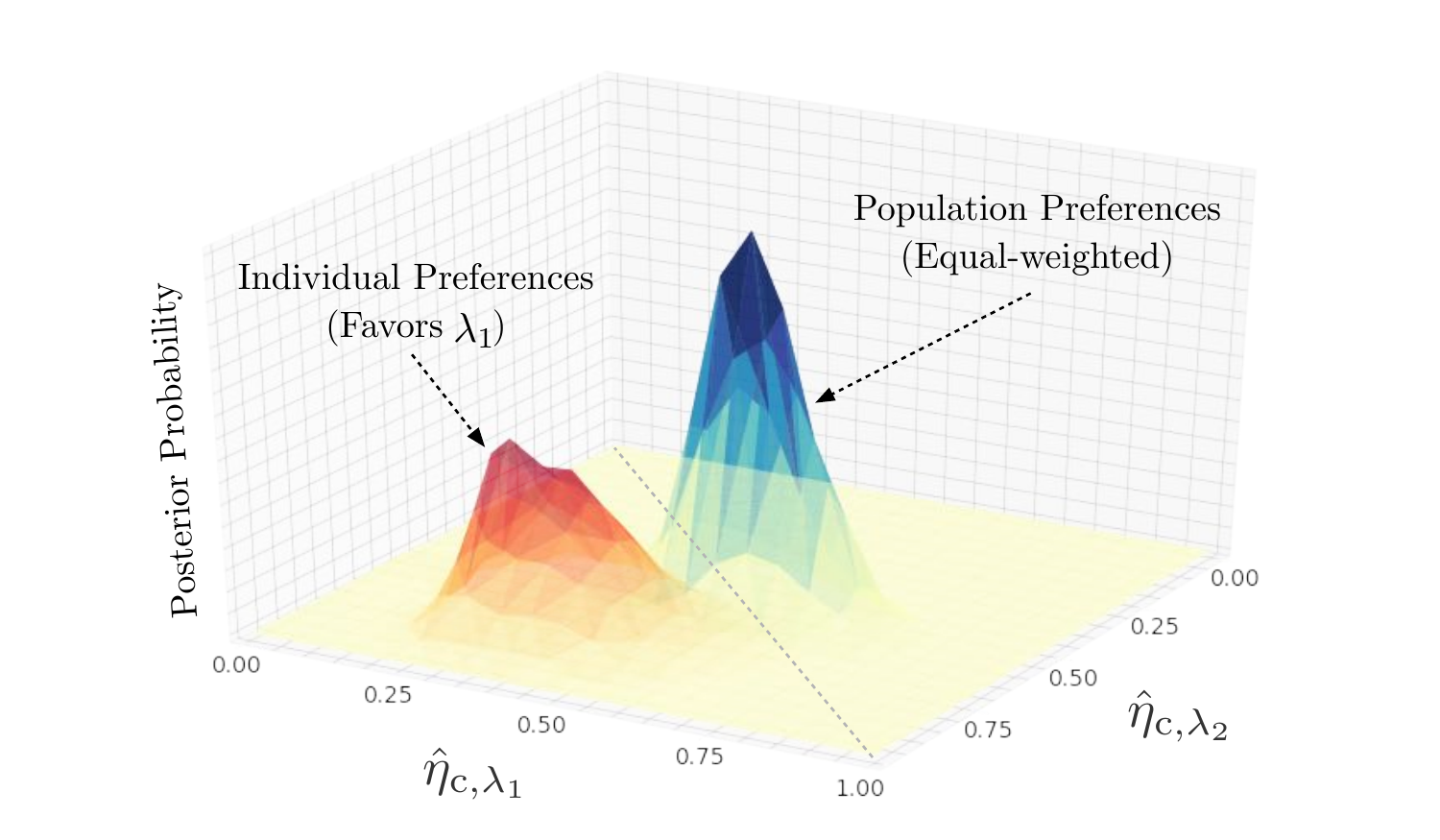}}
\hfill
\subfloat[Understanding Effective Preferences]
{\includegraphics[width=0.32\linewidth, trim=4em 0.5em 4em  4em]{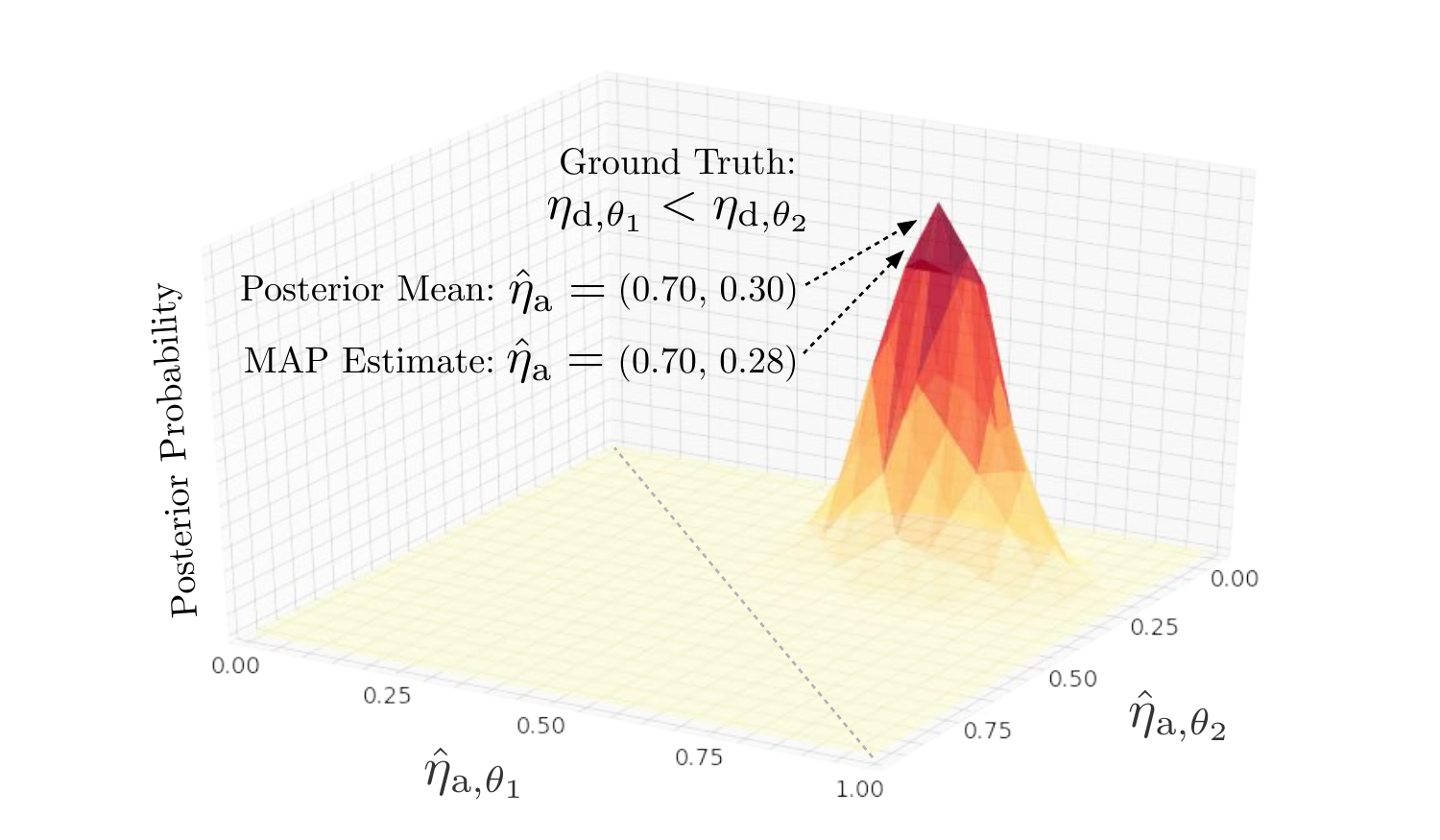}}
\vspace{-0.5em}
\caption{\small \textit{Inverse active sensing}. (Relevant dimensions of) posteriors $\mathbb{P}\{\eta,\rho|*,\dataset\}$ for Examples \ref{eg:3}--\ref{eg:5} (cf. Proposition \ref{thm:strategy_posterior}). Each distribution is generated as \scalebox{0.9}{$1000$} \textsc{mcmc} samples, cf. Algorithm {\color{mydarkblue}1}; \textsc{map} estimates are per Equation {\color{mydarkblue}26}. \scalebox{0.9}{$N=300$} episodes are simulated for optimal softmax agents in (a) and the (biased) ``individual'' agent in (b), and \scalebox{0.9}{$N=1000$} for the (unbiased) ``population'' agent in (b); a greedy lookahead softmax agent (\scalebox{0.9}{$N=300$}) is used in (c). Uniform priors $\mathbb{P}\{\eta|*\}$ and $\mathbb{P}\{\rho\}$ are employed in all instances (\textit{continued on page \hyperlink{page.9}{9}}).}
\label{fig:inverse}
\vspace{-0.75em}
\end{figure*}

\section{Illustrative Examples}

We show archetypical examples that exercise our framework through numerical simulation. Examples \ref{eg:1}--\ref{eg:2} give intuition for optimal active sensing, and \ref{eg:3}--\ref{eg:5} exemplify potential applications of {\small IAS}. Due to space limitation, \mbox{commentary is} necessarily brief; Appendix \ref{app:sim_detail} gives more context and detail.

\vspace{-5pt}
\begin{example}[Ternary Hypothesis]\upshape\label{eg:1}
We first give \textit{geometric} intuition for the forward problem\textemdash exercising Propositions \ref{thm:continue_terminate}--\ref{thm:surprise_suspense} and Equation {\color{mydarkblue}20}. Consider the ternary hypothesis space $\Theta=\{\theta_{1},\theta_{2},\theta_{3}\}$, where the decision-maker is equipped with unary tests $\lambda_{1},\lambda_{2},\lambda_{3}$ (each probabilistically confirming or denying an individual hypothesis) and binary tests $\lambda_{12},\lambda_{23},\lambda_{13}$ (each probabilistically distinguishing between pairs of hypotheses). Figures \ref{fig:plex}(a)--(d) depict the output of (optimal) active sensing (by dynamic programming, cf. Proposition \ref{thm:optimal_value}) in the posterior simplex, showing the relationships among $Q$-factors and their intersections, acquisitions and decisions, as well as continuation and termination sets.
\end{example}

\vspace{-9pt}
\begin{example}[Decision Tree]\upshape\label{eg:2}
{We illustrate \textit{belief trajectories} for a common class of decision problems. Consider a medical diagnosis setting where the disease space $\Theta=\{\theta_{1},\theta_{2},\theta_{3},\theta_{4}\}$ is arranged in a hierarchy, and where the diagnostic agent has access to a (top-level) test $\lambda_{0}$ that (probabilistically) distinguishes between the groups $\{\theta_{1},\theta_{2}\}$ vs. $\{\theta_{3},\theta_{4}\}$ (and are otherwise uninformative), and (level-2) tests $\lambda_{12}$ and $\lambda_{34}$ that respectively (probabilistically) distinguish between $\theta_{1}$ vs. $\theta_{2}$, and $\theta_{3}$ vs. $\theta_{4}$ (and are otherwise uninformative). Figures \ref{fig:plex2}(e)--(g) visualize episodes for differ- ent cost-sensitivity preferences $\eta_{\text{c}}$, as well as verifying our intuition that the optimal strategy navigates \textit{down} the tree.\parfillskip=0pt\par}
\end{example}

\vspace{-9pt}
\begin{example}[Differential Importance]\upshape\label{eg:3}
We show an arche- typical application of {\small IAS} in recovering preferences from behavior\textemdash exercising Propositions \ref{thm:strategy_posterior}--\ref{thm:differentiable_posterior} and Algorithm {\color{mydarkblue}1}. Consider a \textit{preoperative testing} problem, where the aim is to confirm (\smash{$\theta_{1}$}) or deny (\smash{$\theta_{2}$}) the absence of comorbidities that may complicate surgery. Given the downside risk, we certainly hope to verify that Type I errors are taken more seriously than Type II (i.e. accuracy weights \smash{$\eta_{\text{\pix a},\theta_{2}}>\eta_{\text{\pix a},\theta_{1}}$}). Suppose that (unbeknownst to us) this is true in practice: we simulate a random collection $\mathcal{D}$ of decision episodes for a Bayes-optimal softmax decision agent driven by $\eta_{\text{\pix a}}=(0.25, 0.75)$. Figure \ref{fig:inverse}(a) depicts the output of our \hyperlink{page.7}{\small MAP} and \hyperlink{page.7}{\small MCMC} solutions, showing (relevant dimensions of) recovered estimates for optimal softmax strategies, along with the true weights.
\end{example}

\vspace{-9pt}
\begin{example}[Differential Treatment]\upshape\label{eg:4}
We highlight the applicability of {\small IAS} in comparing preferences \textit{across} different agents or populations. Consider the problem of detecting the phenomenon of \textit{prescription bias} with respect to two different diagnostic tests ($\lambda_{1},\lambda_{2}$) for the same disease. Absent bias, by definition it must be the case that all \smash{$\eta_{\text{\pix c},\lambda}$} take on identical values. Suppose that (unbeknownst to us) one hospital secretly favors $\lambda_{1}$ (ceteris paribus) more than $\lambda_{2}$, unlike the rest of the hospital network; we simulate episodes for each accordingly. Figure \ref{fig:inverse}(b) shows (relevant dimensions of) the output of Algorithm {\color{mydarkblue}1} and Equation {\color{mydarkblue}26}\textemdash for both the institution in question and the population (of other institutions). The former's bias in favor of \mbox{$\lambda_{1}$ (i.e. with cost-sen-} sitivity weights \smash{$\eta_{\text{\pix c},\lambda_{1}}$\pix$<$\pix$\eta_{\text{\pix c},\lambda_{2}}$}) is evident, in contrast with the (apparently) unbiased behavior of the latter (\smash{$\eta_{\text{\pix c},\lambda_{1}}$\pix$\approx$\pix$\eta_{\text{\pix c},\lambda_{2}}$}).
\end{example}

\vspace{-9pt}
\begin{example}[Effective Preferences]\upshape\label{eg:5}
{Finally, we give an example that emphasizes the \textit{interpretative} nature of {\small IAS}. Consider an agent who (unbeknownst to us) behaves myopically (cf. greedy lookahead), with a higher decision-threshold parameter for one hypothesis (i.e. $\eta_{\pix\text{d},\theta_{1}}$\pix$<$\pix$\eta_{\pix\text{d},\theta_{2}}$). Now, if we were just interested in the generic question of what best describes their behavior, we would simply run {\small IAS} across the entire space $\mathcal{P}$\textemdash including all classes $\kappa$ of interest. In this case, for instance, we would\textemdash unsurprisingly\textemdash recover some configuration $(\text{\small GL},\eta,\rho)$ as {\small MAP} estimate. But suppose we are actually interested in \textit{interpreting} their behavior \textit{as if} they were optimal with respect to some preferences (which we wish to identify). We are now asking the question: No matter what your internal decision-making processes are, what are you \textit{effectively} prioritizing? For instance, if the medical consensus is that $\lambda_{1}$ is more important to catch than $\lambda_{2}$, then recovering the \textit{effective} values of $\eta_{\pix\text{a}}$ (via $\kappa=*$) would give an immediate assessment of this. Figure \ref{fig:inverse}(c) shows (relevant dimensions of) our {\small IAS} output on the (greedily) simulated decision episodes, where we find (ceteris paribus) that thresholds $\eta_{\pix\text{d},\theta_{1}}<\eta_{\pix\text{d},\theta_{2}}$ \textit{effectively} translate into accuracy weights $\eta_{\pix\text{a},\theta_{1}}>\eta_{\pix\text{a},\theta_{2}}$, which (in this case) accords\textemdash at least ordinally\textemdash with the medical consensus.\parfillskip=0pt\par}
\end{example}
\vspace{-5pt}

\setcounter{figure}{2}
\begin{figure*}[t]
\vspace{-0.75em}
\centering
\setcounter{subfigure}{4}
\subfloat[Decision Tree]
{\includegraphics[width=0.145\linewidth]{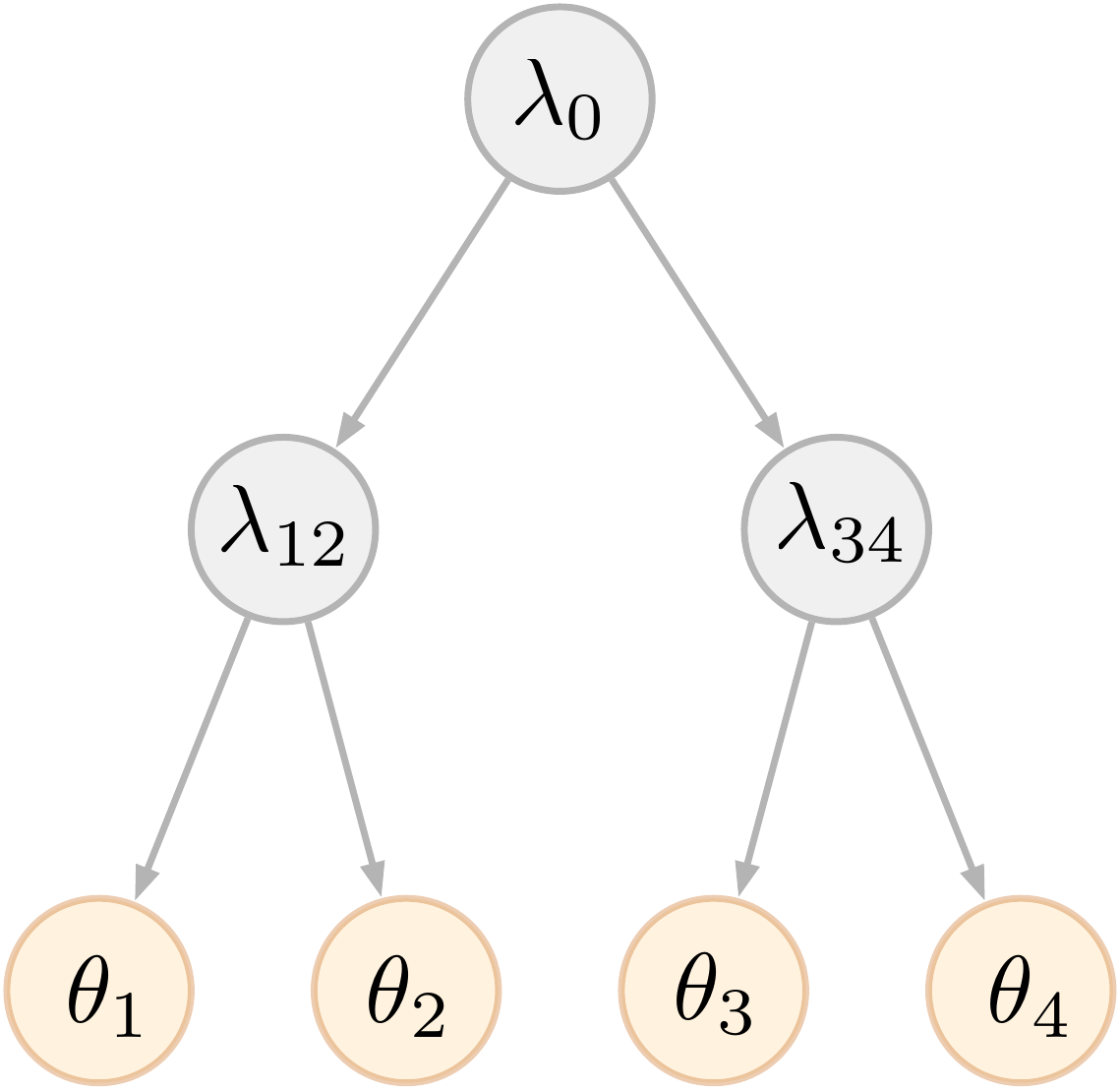}}
\hfill
\subfloat[Typical Belief Trajectory]
{\includegraphics[height=0.15\linewidth]{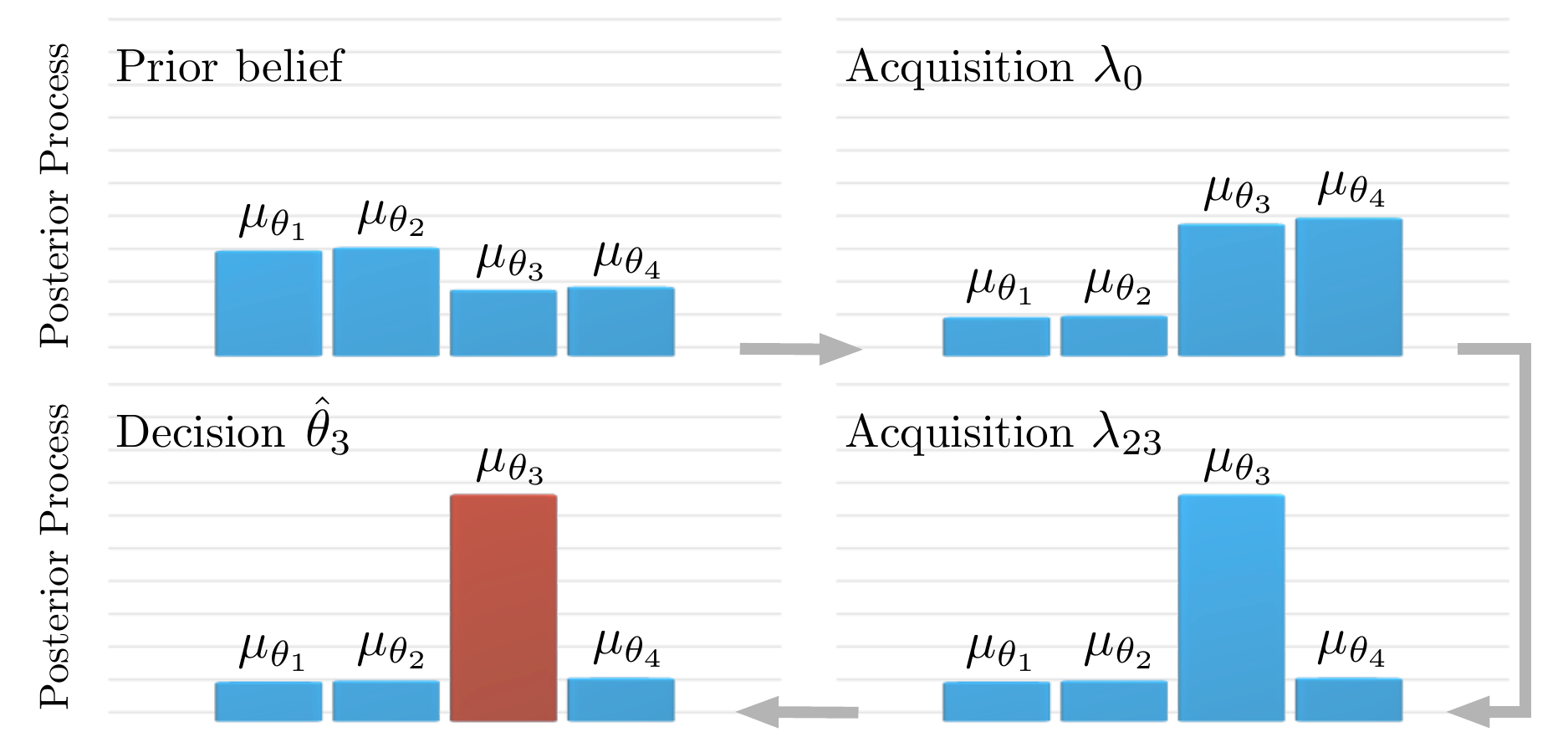}}
\hfill
\subfloat[Typical Belief Trajectory with Lowered $\eta_{\text{\pix c}}$]
{\includegraphics[height=0.15\linewidth]{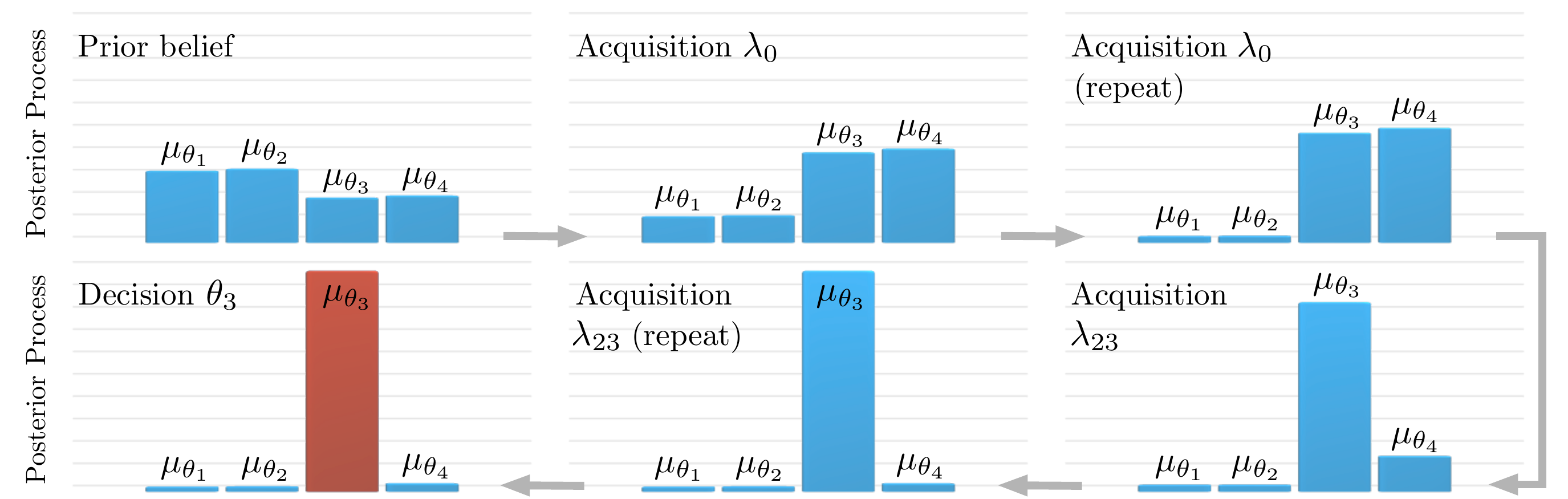}}
\vspace{-0.5em}
\caption{\small \textit{Optimal active sensing (continued from page \hyperlink{page.8}{8})}. Example \ref{eg:2} considers a medical diagnosis setting with diseases $\theta_{1},\theta_{2},\theta_{3},\theta_{4}$ arranged in a hierarchy (e)\textemdash each $\lambda$ (probabilistically) distinguishes between its child elements. Intuitively, we expect that the optimal strategy navigate \textit{down} the decision-tree, sequentially going from high-level tests to low-level tests before declaring specific diagnoses. Figure (f) shows a typical \textit{belief trajectory} for the optimal strategy computed; observe from its decision-behavior that it indeed successively narrows down the space of hypotheses through the tree. Figure (g) additionally shows the effect of uniformly decreasing the cost-sensitivity parameter $\eta_{\text{\pix c}}$\textemdash as expected, the optimal strategy can now afford to ``double-check'' each test result before committing down each branch.}
\label{fig:plex2}
\vspace{-1.0em}
\end{figure*}

\begin{figure*}[t]
\vspace{0.6em}
\centering
\setcounter{subfigure}{3}
\subfloat[Preference Weights and Posteriors]
{\includegraphics[width=0.32\linewidth, trim=4em 0.75em 4em  4em]{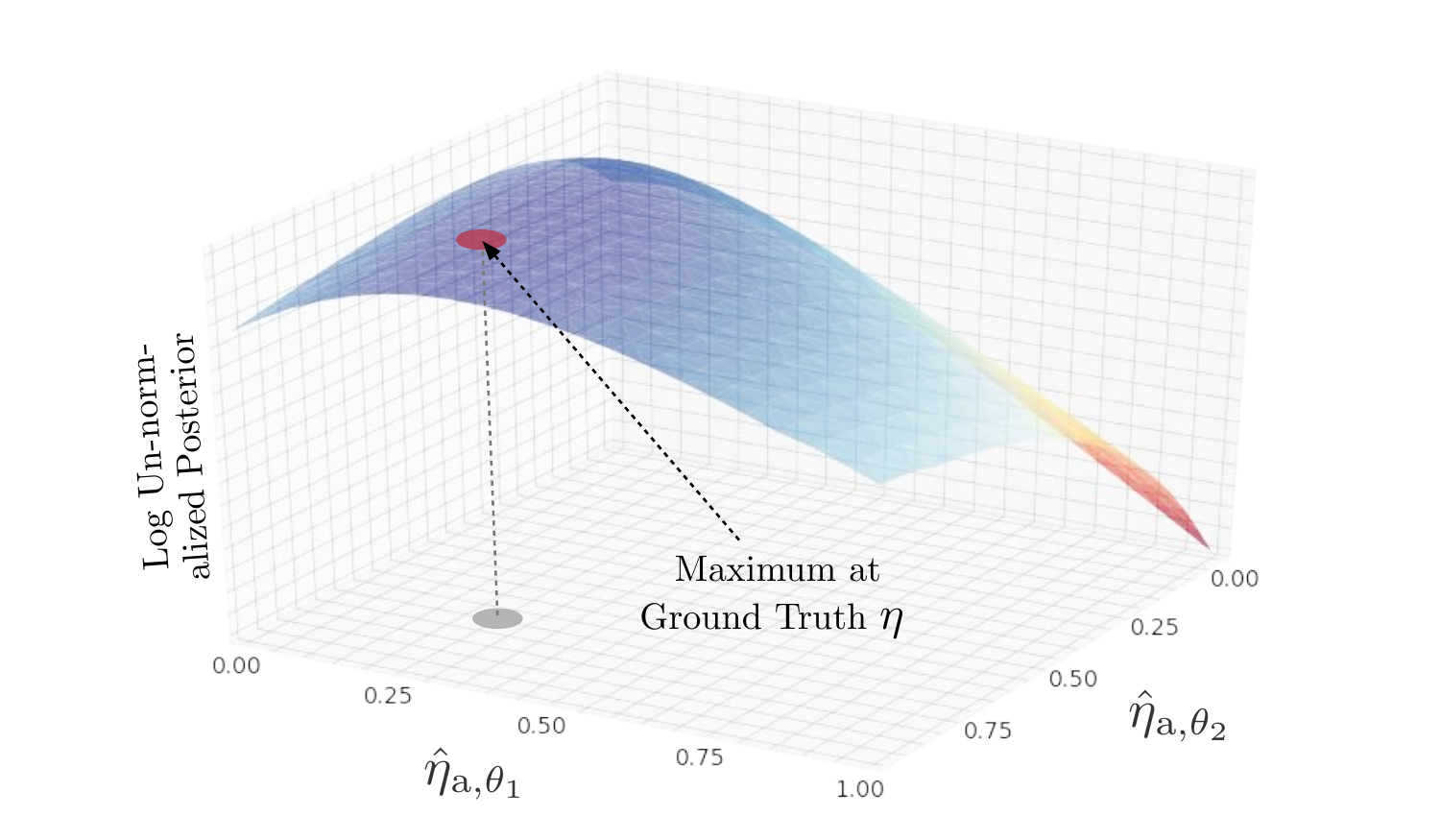}}
\hfill
\subfloat[Inverse Temperature and Posteriors]
{\includegraphics[width=0.32\linewidth, trim=4em 0.75em 4em  4em]{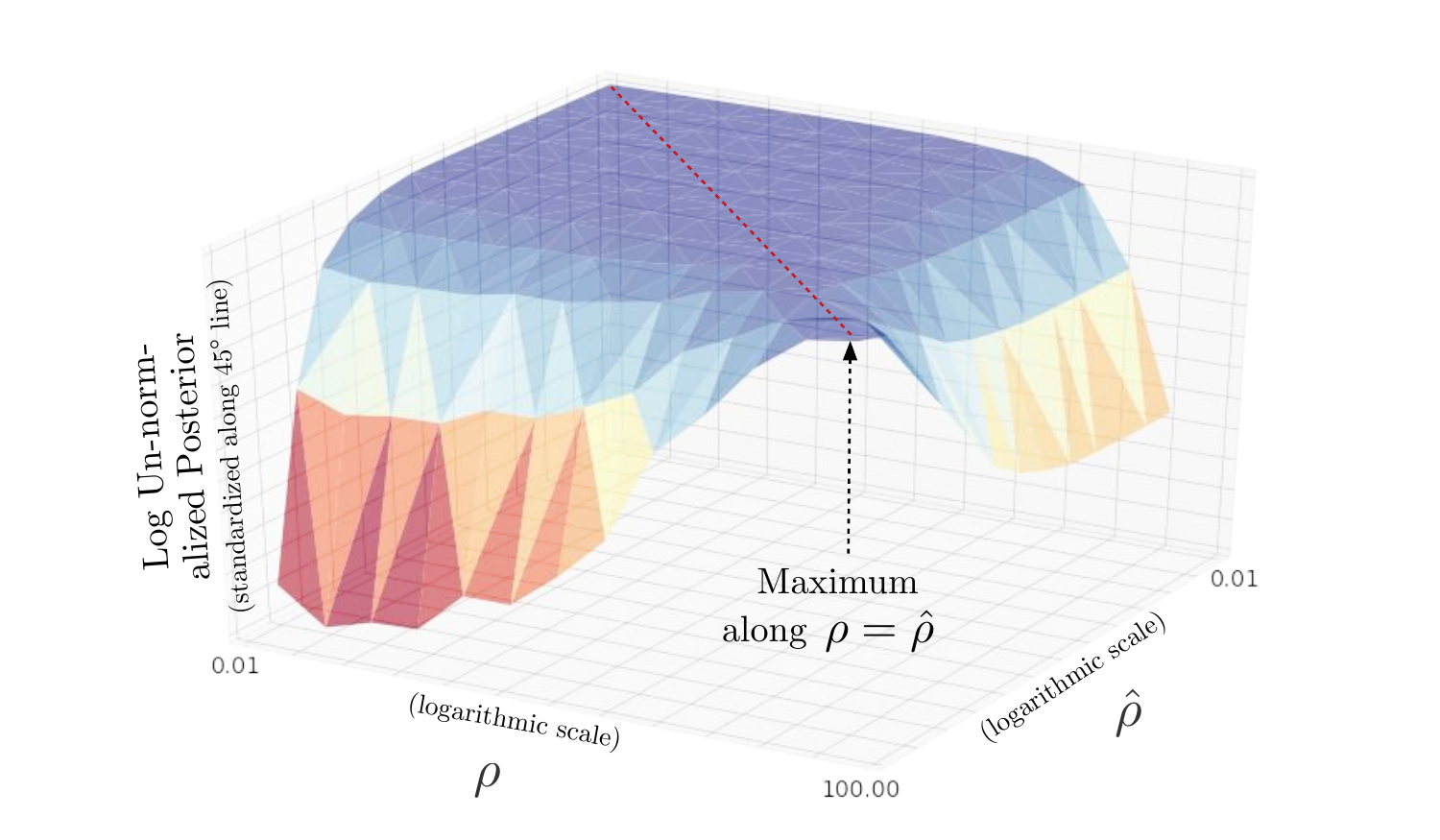}}
\hfill
\subfloat[Preference Weights and Average Risk]
{\includegraphics[width=0.32\linewidth, trim=4em 0.75em 4em  4em]{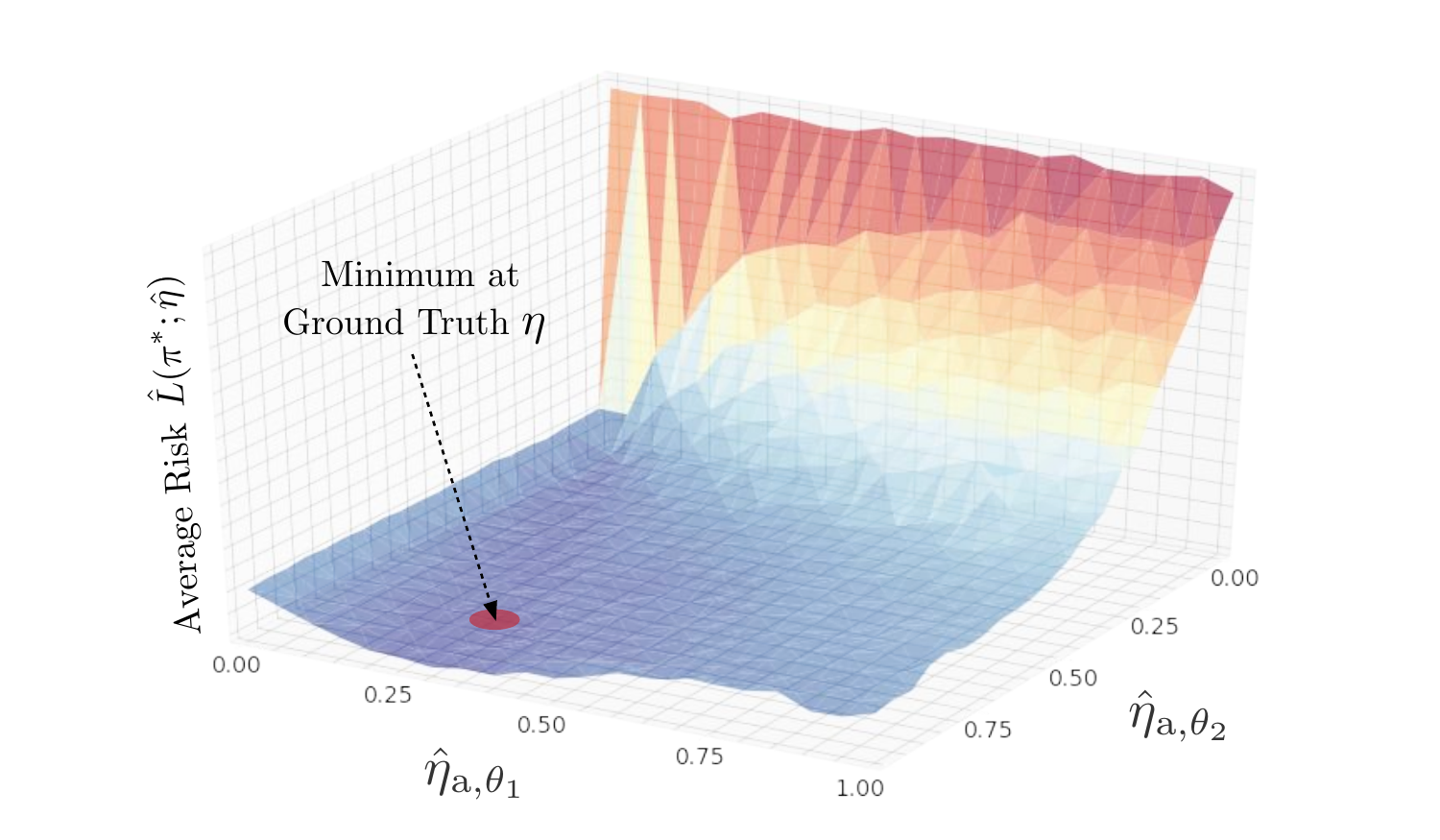}}
\vspace{-0.5em}
\caption{\small \textit{Inverse active sensing (continued from page \hyperlink{page.8}{8})}. For additional visual intuition, (d) computes the (log un-normalized) posterior in relation to (relevant dimensions of) the space of preferences $\eta$ for Example \ref{eg:3}; we observe (as expected) that the posterior is maximized at values that coincide with the ground truth. Similarly, (e) shows the (log un-normalized) posterior in relation to the inverse temperature $\rho$ in the context of Example \ref{eg:3}; here we explicitly simulate a range of ground-truth values and observe (as expected) that the posterior is maximal along the $45^{\circ}$ line (to make this clearer, we standardize all probabilities along this line). Finally, in (f) we observe (as expected) that the (Bayes-optimal) strategy induced by the true parameters is in fact the strategy that achieves the lowest average (ground-truth) risk.}
\label{fig:inverse2}
\vspace{-0.75em}
\end{figure*}

\textbf{Discussion}.
{In this work, we developed a unified theoretical framework for evidence-based decision-making under time pressure, illustrating how it enables modeling intuitive tradeoffs in decision strategies, and understanding behavior by quantifying preferences implicit in observed strategies.\parfillskip=0pt\par}

\vspace{-0.1em}
In modeling the forward problem, our formulation of active sensing inherits several assumptions from prior literature \cite{ahmad2013active, alaa2016balancing, chernoff1959sequential, dayanik2013reward, frazier2008sequential, naghshvar2013active}\textemdash that the spaces of decisions, acquisitions, and outcomes are given; that the distributions of deadlines, outcomes, and problem instances are known or must be appropriately estimated beforehand; and that the outcomes of acquisitions are conditionally independent over time\textemdash which is not always true depending on the type of acquisition in question, and is a shortcoming of this approach that requires extra care in practical applications. On the other hand, our framework departs from prior work by way of expressivity and specification\textemdash in accounting for the presence, endogeneity, and context-dependence of time pressure; in accommodating differential costs of acquisition and penalties for inaccuracies and deadline breaches; and in modeling preference weights directly via behavioral data without prior specification. Where need be, note that it is possible for future work to generalize the present approach to continuous outcome spaces, or disjoint outcome spaces $\Omega_{\lambda}$ per acquisition; to incorporate learnable mappings from instance-specific features $\mathbf{x}$ to priors $\mu_{0}(\mathbf{x})$; and to allow unknown environment parameters to be jointly estimated (although complexity may be of concern in high dimensions).

\vspace{-0.1em}
{In approaching the inverse problem, we inherit a data-driven formulation of inverse optimization\textemdash i.e. where solutions to multiple problem instances are observed \cite{aswani2018inverse, barmann2017emulating, dong2018generalized, esfahani2018data}. Further, our Bayesian method is similar to prior approaches in inverse problems settings \citep{bardsley2012mcmc, ramachandran2007bayesian, ye2019optimization}, and the posterior sampler bears resemblance to a Bayesian solution to inverse optimal control \citep{abbeel2004apprenticeship}, which analogously adapts geometric random walks to the underlying parameter space. The complexity of Algorithm {\color{mydarkblue}1} is therefore identical; it has been shown that such a Markov chain is rapidly-mixing (i.e. sampling terminates in a polynomially-bounded number of steps) under some assumptions \cite{ramachandran2007bayesian, applegate1990sampling}. For a more detailed survey of related literature, see Appendix \ref{app:related_work}.\parfillskip=0pt\par}

\setlength{\bibsep}{5.5pt}

\clearpage
\section*{Acknowledgments}

This work was supported by Alzheimer's Research UK (ARUK), the US Office of Naval Research (ONR), and the National Science Foundation (NSF): grant numbers 1407712, 1462245, 1524417, 1533983, 1722516. We thank all reviewers for their generous comments and suggestions.

\bibliography{bib}\balance

\bibliographystyle{icml2020}

\clearpage\nobalance\appendix

\section{Notes on Simulations}\label{app:sim_detail}

\textbf{Context for Example 1}.
{Propositions \ref{thm:continue_terminate}--\ref{thm:surprise_suspense} and Equation {\color{mydarkblue}20} give a theoretical characterization of optimal active sensing. The aim of this example is to visualize the geometry of the forward problem in the simplex, illustrating these various results through a non-trivial example. In addition to the main points to note in the captions to Figures \ref{fig:plex}(a)--(d), in this example we set $\eta_{\text{\pix a},\theta_{1}}$$<$$\eta_{\text{\pix a},\theta_{2}}$$<$$\eta_{\text{\pix a},\theta_{3}}$ and likewise $\eta_{\text{\pix b},\theta_{1}}$$<$$\eta_{\text{\pix b},\theta_{2}}$$<$$\eta_{\text{\pix b},\theta_{3}}$, where $\eta_{\text{\pix a},\theta}$$<$$\eta_{\text{\pix b},\theta}$ for all $\theta$ (as is often the case\textemdash for medical diagnosis, for instance\textemdash failing to make a decision before the deadline is at least as bad as making the incorrect decision); observe that this preference ordering among the hypotheses is reflected in the termination regions in Figure \ref{fig:plex}(c): The optimal strategy most readily commits to $\theta_{3}$ since it is the most important to catch, whereas it can afford to be surer of $\theta_{1}$ before committing to it. Finally, note that Proposition \ref{thm:surprise_suspense} operates implicitly behind Figure \ref{fig:plex}(d): In this example, we set $p_{\text{unary}},q_{\text{unary}}$ and $p_{\text{binary}},q_{\text{binary}}$ such that the former are more powerful but more risky, and the latter are less powerful but less risky, which induces a surprise-suspense tradeoff; note that increasing the power (or decreasing the risk) of unary tests would naturally expand the (inner) acquisition regions or $\lambda_{1},\lambda_{2},\lambda_{3}$ relative to $\lambda_{12},\lambda_{23},\lambda_{13}$, and vice versa in the opposite direction. (Moreover, the tradeoff in Equation {\color{mydarkblue}20} is similarly (but trivially) implicit in Figure \ref{fig:plex}(a): The peak of the $Q$-factor for decisions gravitates away from vertices with higher $\eta_{\text{\pix a},\theta}$).\parfillskip=0pt\par}

\textbf{Context for Example 2}.
{While Example 1 illustrates properties of the optimal $Q$-factors, Example 2 and Figure \ref{fig:plex2}(e)--(g) visualizes the optimal strategy \textit{in action} (i.e. showing typical belief trajectories) through an intuitive example from medical diagnosis. Consider the diagnostic problem with diseases $\theta_{1},\theta_{2},\theta_{3},\theta_{4}$ arranged in a hierarchy as in Figure \ref{fig:plex2}(e) such that each test $\lambda$ probabilistically distinguishes between its child elements, which can be specific diseases, groups of diseases, or even disease stages as in progressive cognitive impairment \cite{jarrett2019dynamic}; for real-world analogies see for instance \citet{national2016preoperative, national2017all}. We naturally expect that the optimal strategy navigate \textit{down} the decision-tree, starting first from high-level tests, then onto low-level tests, before finally declaring specific diagnoses of diseases. Panel (f) shows a typical \textit{belief trajectory} for the optimal strategy; observe from its decision behavior that it indeed successively narrows down the space of hypotheses through the tree. Panel (g) additionally shows the effect of uniformly decreasing the cost-sensitivity parameter $\eta_{\text{\pix c}}$: as expected, the optimal strategy now affords to ``double-check'' test results before committing to a branch.\parfillskip=0pt\par}

\setlength{\baselineskip}{11.81pt}
\textbf{Context for Example 3}.
Unlike the previous two (which serve to illustrate our results for the forward problem), this gives an archetypical example exercising the \textit{full} framework for \textsc{ias} that we have been building towards. In this case, we specifically use the problem of preoperative testing as a concrete setting, but more broadly we are simply demonstrating the central capability of \textsc{ias}\textemdash that is, in understanding preferences from behavior: Given the decision-behavior of an agent acting according to unknown preferences, can we recover their preferences? To do so, here we perform inverse optimal active sensing on a simulated agent that in fact behaves as $\kappa=*$ (i.e. the model matches the behavior); in Example 5, we highlight the interpretive nature of \textsc{ias} through a more general example (where there is a mismatch). First, we simulate a collection $\mathcal{D}$ of 300 decision episodes for a Bayes-optimal softmax agent with access to a single preoperative test for surgery-complicating comorbidities. The agent is driven by $\eta_{\text{\pix a}}=(0.25, 0.75)$; that is, Type I errors are taken more seriously than Type II errors\textemdash but this is (of course) unknown from the \textsc{ias} point of view, and the pretext is that we wish to estimate $\eta_{\text{\pix a}}$ from $\mathcal{D}$. Complete \textsc{ias} (cf. Proposition \ref{thm:strategy_posterior}) would yield an estimate for the full tuple $(\kappa,\eta,\rho)$; in Figure \ref{fig:inverse}(a) we show dimensions of the result for $\kappa=*$ relevant to this example. The \textsc{map} estimate is computed as Equation {\color{mydarkblue}26}, and \mbox{the posterior as Algorithm {\color{mydarkblue}1}}. For additional visual intuition, Figures \ref{fig:inverse2}(d)--(f) depict the (log un-normalized) posterior probabilities in relation to values of $\eta$ and $\rho$ in this example, and also verify numerically\textemdash through 10,000 random episodes\textemdash that the (Bayes-optimal) strategy induced by the true parameter values is in fact the strategy with the lowest average (ground-truth) risk.

\vspace{-0.2em}
\textbf{Context for Example 4}.
Clearly \textsc{ias} allows analyzing preference weights \textit{within} a decision-agent (i.e. differential importances)\textemdash that is our objective from the beginning. However, we are often also interested in comparing preference weights \textit{across} agents and/or populations. In the case of healthcare, for instance, current diagnostic guidelines are largely based only on consensus \citep{martin2017routine}, with remarkable physician-, provider-, and population-level variability in clinical practice even among routine procedures \citep{song2010regional}, which may incur significant harms and costs \citep{bock2016preoperative}. This example illustrates the potential use of \textsc{ias} in assessing such differences in behavior. As a concrete setting, consider the phenomenon of \textit{prescription bias} w.r.t. two different diagnostic tests ($\lambda_{1},\lambda_{2}$) for the same disease. Using our timely decision-making framework, prescription bias is naturally defined, simulated, and detected as inequalities between \smash{$\eta_{\text{\pix c},\lambda}$} for different $\lambda$. Ceteris paribus, we simulate the presence of bias in an ``individual'' institution of of interest (via 300 trajectories driven by cost-sensitivity weights \smash{$\eta_{\text{\pix c},\lambda_{1}}$$<$$\eta_{\text{\pix c},\lambda_{2}}$}); similarly, we simulate the absence of bias in the broader ``population'' (via 1000 trajectories driven by \smash{$\eta_{\text{\pix c},\lambda_{1}}$$\approx$$\eta_{\text{\pix c},\lambda_{2}}$}). (Two runs of) \textsc{ias} would yield estimates $(\kappa,\eta,\rho)$ each for the individual and population parameters; in Figure \ref{fig:inverse}(b) we show relevant dimensions of the results for $\kappa=*$, where we observe the apparent deviation \mbox{of the individual's preferences from that of the population.}

\setlength{\baselineskip}{12pt}
\textbf{Context for Example 5}.
While Examples 3--4 show the result of \textsc{ias} with $\kappa=*$ on an agent that behaves as $\kappa=*$, here we emphasize the \textit{interpretive} nature of \textsc{ias} for understanding decision-making behavior through a more general example\textemdash where there is a mismatch. Of course, the (obvious) caveat here\textemdash as in any parameter estimation problem\textemdash is that the mismatch cannot be too large. Clearly a complete mismatch would yield nonsensical results in \textsc{ias}: consider a strategy that simply selects acquisitions and decisions uniformly at random. In practice, however, while there may be a range of (active sensing) decision-making behaviors in the world, we generally expect that they be (somewhat imperfect) approximations to the optimal strategy. For instance, the acquisition behavior induced by the greedy generalized $Q$-factor (Equation {\color{mydarkblue}23}) can be seen as a one-step approximation to $Q^{*}_{\lambda}$ where (apart from the soft decision-threshold) $V^{*}$ is simply replaced by $\bar{Q}$. Figure \ref{fig:inverse}(c) shows what happens when we interpret behavior (unbeknownst to us) generated as $\kappa=\textsc{gl}$, in terms of the \textit{effective} preferences under $\kappa=*$\textemdash namely, that (ceteris paribus) greedy look-ahead behavior driven by $\eta_{\pix\text{d},\theta_{1}}$\pix$<$\pix$\eta_{\pix\text{d},\theta_{2}}$ is roughly equivalent to $\eta_{\pix\text{a},\theta_{1}}>\eta_{\pix\text{a},\theta_{2}}$. This (perhaps obvious) point is worth belaboring\textemdash that is, while decision agents may not necessarily be optimal in practice, this has little bearing on the fact that inverse optimal active sensing can still be able to provide a common yardstick by which different decision behaviors can be quantified and compared.

\begin{table*}[t]\small
\setlength\tabcolsep{3.0pt}
\renewcommand{\arraystretch}{1.02}
\vspace{-1.1em}
\caption{\textit{Comparison with related work in sequential analysis}. Viewed from the perspective of sequential analysis, our decision problem can be framed as one of active multiple-hypothesis testing via adaptive and sequential sensing in the presence stochastic, endogenous, and context-dependent time pressure. An exemplary work is shown for each category. Importantly, we focus on the significance of \textit{subjective preferences}, and develop a most general framework accommodating both \textit{forward} (i.e. modeling) \& \textit{inverse} (i.e. understanding) problems.}
\label{tab:related_representation}
\begin{center}
\begin{tabular}{l|ccccccc|c} \toprule
\textbf{Literature} & Acquisition & Decision & Strategy & Evidence & Costs & Horizon & Deadline & Problem \\
\midrule
\citet{wald1948optimum}           & Passive & Binary   & -        & Sequential & Fixed        & No          & -          & Forward \\
\citet{blahut1974hypothesis}      & Passive & Binary   & -        & Batch      & Fixed        & No          & -          & Forward \\
\citet{bertsekas1995dynamic}      & Passive & Binary   & -        & Sequential & Fixed        & Fixed       & External   & Forward \\
\citet{frazier2008sequential}     & Passive & Binary   & -        & Sequential & Fixed        & Stochastic  & External   & Forward \\
\citet{lorden1977nearly}          & Passive & Multiple & -        & Sequential & Fixed        & No          & -          & Forward \\
\citet{tuncel2005error}           & Passive & Multiple & -        & Batch      & Fixed        & No          & -          & Forward \\
\citet{dayanik2013reward}         & Passive & Multiple & -        & Sequential & Fixed        & Stochastic  & External   & Forward \\
\citet{polyanskiy2011binary}      & Active  & Binary   & Fixed    & Sequential & Fixed        & No          & -          & Forward \\
\citet{hayashi2009discrimination} & Active  & Binary   & Adaptive & Batch      & Fixed        & No          & -          & Forward \\
\citet{naghshvar2011information}  & Active  & Binary   & Adaptive & Sequential & Fixed        & No          & -          & Forward \\
\citet{nitinawarat2013controlled} & Active  & Multiple & Fixed    & Batch      & Fixed        & No          & -          & Forward \\
\citet{atia2012controlled}        & Active  & Multiple & Adaptive & Batch      & Fixed        & No          & -          & Forward \\
\citet{naghshvar2013active}       & Active  & Multiple & Adaptive & Sequential & Fixed        & No          & -          & Forward \\
\midrule
\textbf{(Ours)}                   & Active  & Multiple & Adaptive & Sequential & Differential & Stochastic  & Endogenous & ~Forward\pix+\pix Inverse \\
\bottomrule
\end{tabular}
\end{center}
\vspace{-1.8em}
\end{table*}

\textbf{Computation}.
For all examples, agents are simulated with inverse temperature $\rho=10$. The precise setting is unimportant, and we observe that similar results obtain for an order of magnitude larger or smaller; however, note that very large values result in more deterministic behavior, which may not be realistic ($\rho=\infty$ gives fully-deterministic strategies), and very small values result in more random behavior, which may result in difficulties in parameter estimation ($\rho=0$ gives strategies that are completely random). For \textsc{mcmc}, we choose the lattice given by the union of $\mathcal{G}_{\eta}\cap[0,1]^{d}\in\mathcal{H}$ and $\mathcal{G}_{\rho}\in\mathbb{R}$, where $\mathcal{G}_{\eta}\doteq\{x:x_{j}~\text{is an integer multiple of}~r\}$ with $r=0.05$ being our choice of resolution for the elements of $\eta$ (and $j$ being the index into elements of $x$), and where $\mathcal{G}_{\rho}\doteq\{0.01,0.03,...,30,100\}$ is the set of roughly (logarithmically) uniformly-spaced values for $\rho$. Note that restricting the values of $\eta$ to $[0,1]$ by itself involves no loss of expressivity, since different values of $\rho$ are equivalent to a scaling of the $Q$-factors, which (by linearity of expectations) is equivalent to a scaling of all elements of $\eta$. What does have an effect on expressivity is the choice of resolution $r$; now, our goal is to understand the \textit{relative} magnitudes of preference weights underlying decision behaviors, and setting $r=0.05$ with the $[0,1]$ bounds means that we can already represent relative importance weights taking on values up to a maximum of 20 times each another. (In practice, if \textsc{ias} still returns estimates with elements at opposing boundaries of the lattice, this may indicate that we need to further increase the resolution\textemdash e.g. by setting $r=0.01$, which would allow representing relative importance weights up to 100 times one another). For each inverse example, the posterior distributions (using uniform priors) are generated as 1000 samples; with 300 initial ``burn-in'' samples discarded.

\textbf{Modeling Priors}.
We briefly mention here a point for (more applied) future work. In this paper we focus on developing a theoretical framework and demonstrating archetypical examples for modeling and understanding timely decision-making behavior. Therefore we do not concern ourselves with the (separate but related) problem of obtaining or modeling the priors $\mu_{0}$ themselves. Recall from Section \ref{sec:decision_problem} that we simply take it that $\mu_{0}$ for a given problem instance is available from an agent's experience, medical literature, etc. (Again, however, bear in mind the interpretive nature of \textsc{ias}: we are \textit{not} effectively assuming that decision-makers themselves possess such exact and common knowledge). In our numerical examples, we simulate episodes for $\mathcal{D}$ with $\mu_{0}$ uniformly randomly scattered throughout the simplex. In practical applications with real-world input data, we probably wish to model $\mu_{0}$ based on additional input (clearly, having a single constant prior may not provide nearly enough variation for meaningful estimation of preference weights). Any such model necessarily depends on the specific context; however, while we defer this topic to future work with a more applied focus, we note that in many cases existing domain-specific models (such as those in medicine) can be more or less adapted for this purpose. See \citet{petousis2018generating} for an example where such models are deployed for modeling initial beliefs also in an inverse setting (although with a very different approach, detailed in the next section). In the context of medical diagnosis, for instance, one can consider a rich literature of feature-based models \citep{freedman2005cancer}, including the widely used and validated Tammem\"agi and Gail risk models \citep{tammemagi2013selection, gail2011personalized, smedley2011platform} for lung cancer and breast cancer, which can consider a variety of baseline features such as age, race, body mass, smoking status, family history, and previous biopsies in generating accurate priors for use.

\vspace{-6pt}
\section{Related Work}\label{app:related_work}

\newcolumntype{L}{>{\centering\arraybackslash}m{1.4cm}}
\newcolumntype{M}{>{\centering\arraybackslash}m{1.5cm}}
\newcolumntype{N}{>{\centering\arraybackslash}m{1.6cm}}
\newcolumntype{O}{>{\centering\arraybackslash}m{1.5cm}}
\newcolumntype{P}{>{\centering\arraybackslash}m{1.6cm}}
\newcolumntype{Q}{>{\centering\arraybackslash}m{1.5cm}}
\newcolumntype{R}{>{\centering\arraybackslash}m{1.2cm}}
\newcolumntype{S}{>{\centering\arraybackslash}m{1.7cm}}
\newcolumntype{T}{>{\centering\arraybackslash}m{1.5cm}}
\newcolumntype{U}{>{\centering\arraybackslash}m{1.5cm}}

In this paper, we develop an expressive theoretical framework for evidence-based decision-making under time pressure, and illustrate how it enables modeling and understanding decision behavior via optimal and inverse active sensing. As such, it lends itself to contextualization within broader notions of both the forward and inverse problem settings. While relevant works have been noted throughout the manuscript, here we provide a more detailed overview.

\textbf{Active Sensing}. In the broadest sense, active sensing refers to the general process of directing one's attention towards extracting \textit{task-relevant} information through interaction with the world \citep{yang2018theoretical}. This broad notion of intentional information gathering has been applied in various settings such as multi-view learning \citep{yu2009active}, sensory processing \citep{schroeder2010dynamics}, personalized screening \citep{ahuja2017dpscreen}, time-series prediction \citep{yoon2018deep}, and black-box classification \citep{janisch2019classification}. While most applications focus on crafting function approximators to optimize performance on the downstream task, our focus is instead in developing an expressive framework for modeling and understanding the decision process itself.

\textit{Timely Decision-Making}. In particular, we study active sensing for the general problem of timely decision-making\textemdash that is, the goal-directed task of selecting which acquisitions to make, when to stop gathering information, and what decision to ultimately settle on. As such, it is related to the sequential identification problem in statistics \citep{naghshvar2013active}, neuroscience \citep{ahmad2013active}, and economics \citep{augenblick2018belief}\textemdash where a hypothesis is selected following observations of relevant evidence. Starting with the seminal work on binary hypothesis testing \citep{wald1948optimum}, a variety of studies have aimed to characterize a range of heuristic and/or optimal strategies, with such extensions as deadline pressure \citep{frazier2008sequential}, incorporating active choice \citep{castro2009active}, and comparisons of behavioral strategies \citep{ahmad2013active}. We emphasize the goal-directed nature of active sensing in general (and our timely decision-making setting): this is in contrast to pure exploration and surveillance problems, which do not involve a specific task (the decision problem).

{\textit{Generalized Setting}. Several key distinctions warrant special attention (see Table \ref{tab:related_representation}). We consider the most flexible decision-making setting: (1) acquisitions are \textit{active}\textemdash i.e. involving choices among multiple competing sensory options; (2) strategies are \textit{adaptive}\textemdash i.e. admitting context-dependent choices determined on the fly; and (3) samples are \textit{sequential}\textemdash i.e. requiring a variable number of observations per the endogenous choice of stopping and issuing a decision. These distinctions are critical\textemdash for instance, if sampling were passive (e.g. single stream of observations), then the task readily \mbox{reduces to the well-studied problem of op-} timal stopping \citep{frazier2008sequential, dayanik2013reward}. Further, as motivated throughout, we additionally account for (4) \textit{differential} costs of acquisition and the presence of (5) stochastic, \textit{endogenous}, and \textit{context-dependent} time pressure. Perhaps most importantly, we accommodate modeling and understanding (6) \textit{subjective} preferences in decision behavior, and uniquely focus on \textit{both} forward and inverse problems in our active sensing framework. Table \ref{tab:related_representation} sets out a comparison with related work in sequential analysis in general, and Table \ref{tab:related_forward} specifically as pertains timely decision-making. In this view, our work develops a most generalized framework to analyze both optimal and inverse problems.\parfillskip=0pt\par}

\begin{table*}[h]\small
\setlength\tabcolsep{3.0pt}
\renewcommand{\arraystretch}{1.05}
\vspace{-1.1em}
\caption{\textit{Summary comparison of \textsc{ias} and \textsc{irl}}. Although the two classes of \textsc{io} problems share superficial resemblance from the perspective of inverse learning from multiple observations, they have vastly different goals and multiple crucial distinctions. In particular, while learning medical diagnosis behavior can be alternatively cast in \textsc{irl} as a generic \textit{apprenticeship} problem, our proposed \textsc{ias} framework is much better suited for \textit{modeling} and \textit{understanding} the decision process itself in timely decision-making settings. $^{1}\pix$\citet{petousis2018generating}.}
\label{tab:related_irl}
\begin{center}
\begin{tabular}{LMNOPQRSTU}
\toprule
\textbf{Approach} & {Markov Process} & {Stopping Time} & {Behavior Parameters} & {Modeling Acquisitions} & {Modeling Decisions} & {Time Pressure} & {Parameters Interpretable} & {Downstream Goal} & {Accuracy of Decision} \\
\midrule
IRL (Petousis)$^{1}$
& States with Transitions
& Fixed
& Per-State Rewards
& Yes
& No
& No
& No
& Apprentice- ship
& Objective, Imposed
\\
IAS \textbf{(Ours)}
& Posterior \mbox{\& Survival}
& Stochastic, Endogenous
& Risk-based Preferences
& Yes
& Yes
& Yes
& Yes
& Understan- ding
& Subjective, Learned
\\
\bottomrule
\end{tabular}
\end{center}
\vspace{-1.8em}
\end{table*}

\textbf{Inverse Active Sensing}. For the inverse direction, we approach the problem from an \textit{inverse optimization} perspective. In general, {\small IO} turns optimization problems on their heads: Given (one or more) solutions to some problem, the goal is to infer (parameters of) the objective function \citep{ahuja2001inverse}. {\small IO} has been applied to a broad range of underlying problems, including inverse linear \citep{dempe2006inverse} and integer \citep{schaefer2009inverse} programming, inverse convex optimization \citep{keshavarz2011imputing}, inverse conic programming \citep{iyengar2005inverse}, and any manner of inverse combinatorial optimization problems \citep{heuberger2004inverse}. Table \ref{tab:related_inverse} shows inverse (optimal) active sensing alongside example formulations for some classic {\small IO} applications.

\setlength{\baselineskip}{11.83pt}
\textit{Multiple Observations}. In particular, inverse active sensing can be interpreted as a form of \textit{data-driven} {\small IO} with multiple observations (of solutions). Methods for data-driven {\small IO} are increasingly relevant with the exponentially growing availability of electronic patient data \cite{jarrett2020target}, and have been studied as pertains to imperfect information \citep{esfahani2018data} and noisy observations \citep{aswani2018inverse}, as well as using online learning \citep{barmann2017emulating, dong2018generalized}. Now, a popular application of this paradigm is inverse reinforcement learning (``\textsc{irl}''), which deals with inferring the reward function for a reinforcement learning agent \citep{abbeel2004apprenticeship, ziebart2008maximum}. Although \textsc{irl} may appear to bear resemblance to \textsc{ias}, they have vastly different goals and a number of crucial distinctions. These are best highlighted by direct comparison with \citet{petousis2018generating}, which applies \textsc{irl} for apprenticeship of expert cancer screening behavior (see Table \ref{tab:related_irl}). In the first instance, (1) the typical goal of \textsc{irl} lies in \textit{apprenticeship}; to that end, the central concern is in replicating some notion of (``true'') performance, using (potentially black-box) reward functions as an intermediary to parameterize behavior. In contrast, in \textsc{ias} the goal lies in \textit{modeling} and \textit{understanding} the decision process itself (in timely decision-making settings); to that end, the central concern is in recovering a (transparent) description of an agent's (subjective) preferences. This distinction becomes apparent in a number of aspects that render \textsc{irl} unsuitable for our purposes. An immediate difference lies in (2) the nature of the Markov process in question: Recall that our formulation tracks a posterior process (cf. Proposition \ref{thm:sufficient_statistic}) over the hypothesis space, with survival itself is informative (cf. Proposition \ref{thm:active_passive}). Applying the \textsc{irl} formulation instead as in \citet{petousis2018generating}, the ``state space'' is taken to be the space of hypotheses; the Markov process tracks where the agent him-/herself is located within the hypothesis space, and the ``transitions'' model the agent probabilistically moving between hypotheses over time. Now, (3) this abstraction is inherently opaque: What does it mean for the agent to ``be'' somewhere, and what how do the transition probabilities inform our understanding of what an agent prioritizes? This is fine simply as a mathematical intermediary to parameterize behavior, but is by no means interpretable as a vehicle for understanding behavior (see also point 5). In contrast, \textsc{ias} purely focuses on the specific task of estimating preferences for understanding. Moreover, (4) these transition parameters must be concomitantly learned, which adds an (unnecessary) layer of approximation. Equally importantly, (5) in the \textsc{irl} formulation (as is typical), the observed behavior is parameterized (and learned) in terms of per-state (and action) rewards, which\textemdash in timely decision-making\textemdash are \textit{not} amenable to interpretation: What does it mean to reward the agent for being ``in'' a given (intermediary) hypothesis at each point in time? Again, this is fine purely as mathematical means to parameterize data (e.g. in their apprenticeship setting), but makes less sense for our purposes of understanding. Instead, we directly parameterize behaviors as importance weights assigned to inherently interpretable elements of the loss function (Equation 1). On a more technical note\textemdash but perhaps even more significantly: (6) in our framework, not only is the stopping time itself is an endogenous variable, it is modeled as a conscious choice (cf. Proposition \ref{thm:continue_terminate}); this is critical, since the ultimate decision itself is in some sense the whole point. In contrast, the \textsc{irl} formulation (as is typical) employs fixed horizons, and does not accommodate modeling the conscious choice of stopping. In fact, to assess apprenticeship, the ``accuracy'' of their learned behavior is quantified via the post-hoc choice of equating some acquisitions to ``positive'' diagnoses (and others to ``negative'' diagnoses); accuracies (e.g. Type I and II errors) are therefore \textit{objective} and \textit{imposed} for evaluation. In contrast, we seek to model the entire decision process endogenously (not just acquisition behavior) via \textit{subjective} preferences over accuracies, deadlines, and costs\textemdash which are \textit{learned}. Last but not least is the technical distinction that (7) the contractive property of the operator $\mathbb{B}$ is not readily guaranteed in our setting (cf. Proposition \ref{thm:optimal_value}); this is in contrast with typical reinforcement learning (and \textsc{irl}) settings with fixed or infinite discounted horizons. Table \ref{tab:related_irl} summarizes main distinctions between the problem classes.
\setlength{\baselineskip}{12pt}

\textit{Bayesian Approach}. In terms of the objective, typical {\small IO} settings are chiefly concerned with notions of identifiability and optimality\textemdash that is, in recovering either some notion of a ``true'' parameter, or in prescribing behavior that performs ``as well as'' (or better than) observed solutions per the ``true'' parameter (this obviously includes inverse reinforcement learning). Instead, the focus of \textsc{ias} is on describing and understanding observed decision behavior; thus we embrace non-identifiability\textemdash after all, we seek the \textit{range} of strategies and preferences that can interpret or best explain behavior (there is no single right answer). In this sense, we are more aligned with Bayesian approaches to inverse problem settings \citep{ye2019optimization, bardsley2012mcmc, ramachandran2007bayesian}, which avoid confronting the convexity assumptions of duality-based approaches \citep{bertsimas2015data, keshavarz2011imputing}, nor the intractability of non-convex solutions \citep{aswani2018inverse, esfahani2018data}.

\textit{Preference Elicitation}. Finally, for completeness we note that preference elicitation is a well-studied problem in computational and social science: A range of works have approached the problem of (interactive) preference elicitation using gaussian processes \citep{guo2010gaussian}, Markov decision processes \citep{wray2016pomdp}, and differentiable networks \citep{vendrov2019gradient}. However, these lines of work are very differet in that what is being modeled (and optimized) is the process of \textit{explicitly} reaching out and querying user preferences efficiently\textemdash that is, the active preference elicitation task itself constitutes the forward problem. In contrast, our focus is on \textit{implicitly} understanding strategies and preferences from observed decision behavior.

\setlength{\baselineskip}{11.83pt}
\textbf{Relationship with POMDPs}. Throughout this work, we have taken a ``bottom-up'' approach in contextualizing our developments\textemdash that is, by taking the basic case of sequential identification and ``generalizing'' from there, which highlights structural results specific to the timely decision-making problem. As its complement, it is equally possible to take an opposite ``top-down'' approach\textemdash that is, by taking the generic POMDP formalism and ``specializing'' from there. In particular, the timely decision-making problem can be formulated as a POMDP with $|\Theta|$ decision states plus an additional ``terminal'' state, with transitions from each of the former into the latter, and self-loops for all states; stepwise decomposing Equation \ref{loss_function} yields a ``reward''. For instance, for the decision tree from Example \ref{eg:2}, the POMDP would consist of the state space $\mathcal{S}=\{\theta_{0},\theta_{1},\theta_{2},\theta_{3},\theta_{4}\}$ where $\theta_{0}$ is absorbing, action space $\mathcal{A}=\{\lambda_{0},\lambda_{12},\lambda_{34},\theta_{1},\theta_{2},\theta_{3},\theta_{4}\}$, emission kernels that correspond to generating distributions $\{q_{\theta,\lambda}\}_{\theta\in\Theta,\lambda\in\Lambda}$, and transition kernels to $\{p_{\theta,\lambda}\}_{\theta\in\Theta,\lambda\in\Lambda}$.

In light of this correspondence to POMDPs, note that Proposition \ref{thm:sufficient_statistic} follows by construction, providing an alternative proof. Note, however, that Propositions \ref{thm:active_passive}--\ref{thm:surprise_suspense} are structural results specific to active sensing for timely decision-making; in particular, we note\textemdash analogously to the passive case of \citet{dayanik2013reward}\textemdash that Proposition \ref{thm:active_passive} is not free due to the fact that this is neither a fixed-horizon nor discounted problem; likewise, concavity of $Q$ is similar to\textemdash but not the same as\textemdash the classic PWLC result. That said, the fact that the (forward) active sensing problem can be re-cast as a POMDP does mean that we can use generic algorithms to accomplish the inner-loop \texttt{ActiveSensing} sub-procedure in Algorithm {\color{mydarkblue}1} (bar minor technicalities in translation, such as the fact that applying off-the-shelf POMDP solvers requires the use of some nominal discount rate $\gamma$\pix$<$\pix$1$ to guarantee convergence). In our simulations, we verify using implementations from \href{http://pomdp.org/code/index.html}{http://pomdp.org/code/index.html} and \href{http://github.com/AdaCompNUS/sarsop}{http://github.com/AdaCompNUS/sarsop} for our examples that all results are virtually identical for any solver of choice, such as PBVI and SARSOP (with $\gamma$ nominally set to 0.99).

{In the inverse direction, as noted above {\small IAS} (with optimal $\kappa$) is likewise related to inverse optimal control; by casting the forward problem generically as a POMDP, solving the inverse optimal active sensing problem in our framework can be interpreted by analogy to a model-based, Bayesian solution to inverse reinforcement learning, but with partially-observable states instead, and a reward function parameterized by stepwise decomposing Equation \ref{loss_function}; though beyond the scope of this work, it is conceivable to derive ``max-margin'', ``max-likelihood'', etc. versions of {\small IAS} (with optimal $\kappa$) in addition to the {\small MAP} and {\small MCMC} versions presented here. Finally, note that non-Bayes-optimal strategies can alternatively be modeled by defining rewards as sums of hand-crafted features, or by using ``belief-dependent'' POMDPs. In the former case, however, this may require more prior knowledge than we have access to, and\textemdash more importantly\textemdash may not result in an interpretable functional form amenable to comparing preferences across decision agents (a key mission objective of ours); in the latter, note that approximating the forward solution to belief-dependent POMDPs in general requires that rewards be convex in $\mu$\textemdash which may be difficult to satisfy or verify in practice.\parfillskip=0pt\par}
\setlength{\baselineskip}{12pt}

\section{Proofs}\label{app:proofs}
\setcounter{theorem}{0}

\setlength{\abovedisplayskip}{6pt}
\setlength{\belowdisplayskip}{6pt}

\Paste{thm_sufficient_statistic}
For $\bar{M}$, we want that $\theta$ be assigned the probability:
\begin{align*}
&\mathbb{P}_{p,q}\{\theta|\lambda_{t-1},\mu_{t-1},\nu_{t-1}=1,\nu_{t}=0\} \numberthis \\
=\pix&\frac{\mathbb{P}_{p,q}\{\theta,\nu_{t}=0|\lambda_{t-1},\mu_{t-1},\nu_{t-1}=1\}}{\mathbb{P}_{p,q}\{\nu_{t}=0|\lambda_{t-1},\mu_{t-1},\nu_{t-1}=1\}} \numberthis \\
=\pix&\frac{\mathbb{P}_{p}\{\nu_{t}=0|\theta,\lambda_{t-1},\nu_{t-1}=1\}\mu_{t-1}(\theta)}{\sum_{\theta^{\prime}\in\Theta}\mathbb{P}_{p}\{\nu_{t}=0|\theta,\lambda_{t-1},\nu_{t-1}=1\}\mu_{t-1}(\theta)} \numberthis \\
=\pix&\frac{p_{\theta,\lambda_{t-1}}\mu_{t-1}(\theta)}{\sum_{\theta^{\prime}\in\Theta}p_{\theta^{\prime},\lambda_{t-1}}\mu_{t-1}(\theta^{\prime})} \numberthis
\end{align*}
For $M$, we want that $\theta$ be assigned the probability:
\begin{align*}
&\mathbb{P}_{p,q}\{\theta|\lambda_{t-1},\mu_{t-1},\nu_{t}=1,\omega_{t}\} \numberthis \\
=\pix&\frac{\mathbb{P}_{p,q}\{\theta,\nu_{t}=1,\omega_{t}|\lambda_{t-1},\mu_{t-1},\nu_{t-1}=1\}}{\mathbb{P}_{p,q}\{\nu_{t}=1,\omega_{t}|\lambda_{t-1},\mu_{t-1},\nu_{t-1}=1\}} \numberthis \\
=\pix&\mathbb{P}_{p}\{\theta,\nu_{t}=1|\lambda_{t-1},\mu_{t-1},\nu_{t-1}=1\}\cdot\mathbb{P}_{q}\{\omega_{t}|\theta, \\
&\lambda_{t-1},\nu_{t}=1\}/{\textstyle\sum}_{\theta^{\prime}\in\Theta}(\mathbb{P}_{p}\{\theta^{\prime},\nu_{t}=1|\lambda_{t-1}, \\
&\mu_{t-1},\nu_{t-1}=1\}\mathbb{P}_{q}\{\omega_{t}|\theta^{\prime},\lambda_{t-1},\nu_{t}=1\}) \numberthis \\
=\pix&\mathbb{P}_{p}\{\nu_{t}=1|\theta,\lambda_{t-1},\nu_{t-1}=1\}\cdot\mathbb{P}_{q}\{\omega_{t}|\theta, \\
&\lambda_{t-1},\nu_{t}=1\}\mu_{t-1}(\theta)/{\textstyle\sum}_{\theta^{\prime}\in\Theta}\mathbb{P}_{p}\{\nu_{t}=1|\theta, \\
&\lambda_{t-1},\nu_{t-1}=1\}\mathbb{P}_{q}\{\omega_{t}|\theta^{\prime},\lambda_{t-1},\nu_{t}=1\}\mu_{t-1}(\theta) \numberthis \\
=\pix&\frac{(1-p_{\theta,\lambda_{t-1}})q_{\theta,\lambda_{t-1}}(\omega_{t})\mu_{t-1}(\theta)}{\sum_{\theta^{\prime}\in\Theta}(1-p_{\theta^{\prime},\lambda_{t-1}})q_{\theta^{\prime},\lambda_{t-1}}(\omega_{t})\mu_{t-1}(\theta^{\prime})} \numberthis
\end{align*}
where we used $\mathbb{P}\{\theta|\lambda_{t-1},\mu_{t-1},\nu_{t-1}=1\}=\mu_{t-1}(\theta)$. To show this is a controlled Markov process, first note that:
\begin{align*}
&\mathbb{P}_{p,q}\{\mu_{t}|\lambda_{t-1},\mu_{t-1},\nu_{t-1},\nu_{t}\} \numberthis \\
=\pix&
(1-\nu_{t-1})
\mathbb{P}_{p}\{\mu_{t}|\lambda_{t-1},\mu_{t-1},\nu_{t-1}=0,\nu_{t}=0\} \\
&+
(
(1-\nu_{t})
\mathbb{P}_{p}\{\mu_{t}|\lambda_{t-1},\mu_{t-1},\nu_{t-1}=1,\nu_{t}=0\} \\
&+
\nu_{t}
\mathbb{P}_{p,q}\{\mu_{t}|\lambda_{t-1},\mu_{t-1},\nu_{t}=1\}
)\nu_{t-1} \numberthis \\
=\pix&
(1-\nu_{t-1})
\mathds{1}_{\{\mu_{t}=\mu_{t-1}\}} \\
&+
(
(1-\nu_{t})
\mathds{1}_{\{\mu_{t}=\bar{M}(\lambda_{t-1},\mu_{t-1})\}} \\
&+
\nu_{t}
\medmath{\frac{
\mathbb{P}_{p,q}\{\mu_{t},\nu_{t}=1|\lambda_{t-1},\mu_{t-1},\nu_{t-1}=1\}
}{
\mathbb{P}_{p}\{\nu_{t}=1|\lambda_{t-1},\mu_{t-1},\nu_{t-1}=1\}
}}
)\nu_{t-1} \numberthis \\
=\pix&
(1-\nu_{t-1})
\mathds{1}_{\{\mu_{t}=\mu_{t-1}\}} \\
&+
(
(1-\nu_{t})
\mathds{1}_{\{\mu_{t}=\bar{M}(\lambda_{t-1},\mu_{t-1})\}} \\
&+
\nu_{t}
{\textstyle\sum}_{\omega^{\prime}_{t}\in\Omega}
(
\mathds{1}_{\{\mu_{t}=M(\lambda_{t-1},\mu_{t-1},\omega^{\prime}_{t})\}} \\
&\cdot\medmath{\frac{
{\textstyle\sum}_{\theta^{\prime}\in\Theta}(1-p_{\theta^{\prime},\lambda_{t-1}})q_{\theta^{\prime},\lambda_{t-1}}(\omega_{t})\mu_{t-1}(\theta^{\prime})
}{
1-{\textstyle\sum}_{\theta^{\prime}\in\Theta}p_{\theta^{\prime},\lambda_{t-1}}\mu_{t-1}(\theta^{\prime})
}}
))\nu_{t-1} \numberthis
\end{align*}
Then the joint probability of the tuple is given by:
\begin{align*}
&\mathbb{P}_{p,q}\{\mu_{t},\nu_{t}|\lambda_{t-1},\mu_{t-1},\nu_{t-1}\} \numberthis \\
=\pix&
\mathbb{P}_{p,q}\{\mu_{t}|\lambda_{t-1},\mu_{t-1},\nu_{t-1},\nu_{t}\} \\
&\cdot\mathbb{P}_{p}\{\nu_{t}|\lambda_{t-1},\mu_{t-1},\nu_{t-1}\} \numberthis \\
=\pix&
(1-\nu_{t-1})
\mathds{1}_{\{\mu_{t}=\mu_{t-1}\}}
+
(
(1-\nu_{t})\\
&\cdot\mathds{1}_{\{\mu_{t}=\bar{M}(\lambda_{t-1},\mu_{t-1})\}}
{\textstyle\sum}_{\theta^{\prime}\in\Theta}p_{\theta^{\prime},\lambda_{t-1}}\mu_{t-1}(\theta^{\prime}) \\
&+
\nu_{t}
{\textstyle\sum}_{\omega^{\prime}_{t}\in\Omega}(
\mathds{1}_{\{\mu_{t}=M(\lambda_{t-1},\mu_{t-1},\omega^{\prime}_{t})\}}\pix\cdot \\
&{\textstyle\sum}_{\theta^{\prime}\in\Theta}(1-p_{\theta^{\prime},\lambda_{t-1}})q_{\theta^{\prime},\lambda_{t-1}}(\omega_{t})\mu_{t-1}(\theta^{\prime})
))\nu_{t-1} \numberthis
\end{align*}
and for any $f:\Delta(\Theta)\times\{0,1\}\rightarrow\mathbb{R}_{^{+}}$ we have:
\begin{align*}
&\mathbb{E}_{p,q}[f(\mu_{t},\nu_{t})|\lambda_{t-1},\mu_{t-1},\nu_{t-1}] \numberthis \\
=\pix&
\mathbb{E}_{p,q}[
(1-\nu_{t-1})f(\mu_{t-1},0) \\
&+(
(1-\nu_{t})f(\bar{M}(\lambda_{t-1},\mu_{t-1}),0)
+
\nu_{t}\pix\cdot \\
&f(M(\lambda_{t-1},\mu_{t-1},\omega_{t}),1)
)\nu_{t-1}
|\lambda_{t-1},\mu_{t-1},\nu_{t-1}] \numberthis \\
=\pix&
(1-\nu_{t-1})f(\mu_{t-1},0)
+
(
f(\bar{M}(\lambda_{t-1},\mu_{t-1}),0) \\
&\cdot{\textstyle\sum}_{\theta^{\prime}\in\Theta}p_{\theta^{\prime},\lambda_{t-1}}\mu_{t-1}(\theta^{\prime}) \\
&+
{\textstyle\sum}_{\omega^{\prime}_{t}\in\Omega}(
f(M(\lambda_{t-1},\mu_{t-1},\omega_{t}),1)\pix\cdot \\
&{\textstyle\sum}_{\theta^{\prime}\in\Theta}(1-p_{\theta^{\prime},\lambda_{t-1}})q_{\theta^{\prime},\lambda_{t-1}}(\omega_{t})\mu_{t-1}(\theta^{\prime})
))\nu_{t-1} \numberthis
\end{align*}
where we used the fact that 
$\mathbb{P}_{p}\{\nu_{t}=1|\lambda_{t-1},\mu_{t-1},\nu_{t-1}=1\}
=\pix
1-{\textstyle\sum}_{\theta^{\prime}\in\Theta}p_{\theta^{\prime},\lambda_{t-1}}\mu_{t-1}(\theta^{\prime})$
, that $\mathbb{P}_{p}\{\nu_{t}=0|\lambda_{t-1},$ $\mu_{t-1},\nu_{t-1}=1\}=\pix
{\textstyle\sum}_{\theta^{\prime}\in\Theta}p_{\theta^{\prime},\lambda_{t-1}}\mu_{t-1}(\theta^{\prime})$.
Likewise, it is also trivial to see that $
\mathbb{P}_{p}\{\nu_{t}=0|\lambda_{t-1},\mu_{t-1},\nu_{t-1}=0\}
=
1
$, as well as $
\mathbb{P}_{p}\{\nu_{t}=1|\lambda_{t-1},\mu_{t-1},\nu_{t-1}=0\}
=
0
$.

\Paste{thm_active_passive}
First, writing out the expectation:
\begin{align*}
&\mathbb{E}_{p,q}[\mu_{t}|\lambda_{t-1},\mu_{t-1},\nu_{t-1},\nu_{t}] \numberthis \\
=\pix&
\mathbb{E}_{p,q}[
(1-\nu_{t-1})\mu_{t-1}
+(
(1-\nu_{t})\bar{M}(\lambda_{t-1},\mu_{t-1}) \\
+&
\nu_{t}M(\lambda_{t-1},\mu_{t-1},\omega_{t})
)\nu_{t-1}
|\lambda_{t-1},\mu_{t-1},\nu_{t-1},\nu_{t}] \numberthis \\
=\pix&
(1-\nu_{t-1})
\mu_{t-1}
+
(
(1-\nu_{t})
\bar{M}(\lambda_{t-1},\mu_{t-1}) \\
&+
\nu_{t}
{\textstyle\sum}_{\omega^{\prime}_{t}\in\Omega}
(
M(\lambda_{t-1},\mu_{t-1},\omega^{\prime}_{t}) \\
\cdot&\frac{
{\textstyle\sum}_{\theta^{\prime}\in\Theta}(1-p_{\theta^{\prime},\lambda_{t-1}})q_{\theta^{\prime},\lambda_{t-1}}(\omega_{t})\mu_{t-1}(\theta^{\prime})
}{
1-{\textstyle\sum}_{\theta^{\prime}\in\Theta}p_{\theta^{\prime},\lambda_{t-1}}\mu_{t-1}(\theta^{\prime})
}
))\nu_{t-1} \numberthis
\end{align*}
Then for element $\theta$, this is equal to:
\begin{align*}
&(1-\nu_{t-1})
\mu_{t-1}
+
(
(1-\nu_{t})
\medmath{
\frac{p_{\theta,\lambda_{t-1}}\mu_{t-1}(\theta)}{\sum_{\theta^{\prime}\in\Theta}p_{\theta^{\prime},\lambda_{t-1}}\mu_{t-1}(\theta^{\prime})}} \\
&~~~+
\nu_{t}
{\textstyle\sum}_{\omega^{\prime}_{t}\in\Omega}
\medmath{
\frac{(1-p_{\theta,\lambda_{t-1}})q_{\theta,\lambda_{t-1}}(\omega_{t})\mu_{t-1}(\theta)}{1-{\textstyle\sum}_{\theta^{\prime}\in\Theta}p_{\theta^{\prime},\lambda_{t-1}}\mu_{t-1}(\theta^{\prime})}}
)\nu_{t-1} \numberthis \\
&=
\mu_{t-1}
+
(
(1-\nu_{t})
\frac{p_{\theta,\lambda_{t-1}}\mu_{t-1}(\theta)}{\sum_{\theta^{\prime}\in\Theta}p_{\theta^{\prime},\lambda_{t-1}}\mu_{t-1}(\theta^{\prime})} \\
&~~~+
\nu_{t}
\frac{(1-p_{\theta,\lambda_{t-1}})\mu_{t-1}(\theta)}{1-{\textstyle\sum}_{\theta^{\prime}\in\Theta}p_{\theta^{\prime},\lambda_{t-1}}\mu_{t-1}(\theta^{\prime})}
-\mu_{t-1})\nu_{t-1} \numberthis \\
&=
\mu_{t-1}
+
(
(1-\nu_{t}) \\
&~~~\cdot\frac{{p_{\theta,\lambda_{t-1}}-\textstyle\sum}_{\theta^{\prime}\in\Theta}p_{\theta^{\prime},\lambda_{t-1}}\mu_{t-1}(\theta^{\prime})}{\sum_{\theta^{\prime}\in\Theta}p_{\theta^{\prime},\lambda_{t-1}}\mu_{t-1}(\theta^{\prime})}
\mu_{t-1}(\theta)- \\
&~~~
\nu_{t}
\medmath{
\frac{p_{\theta,\lambda_{t-1}}-{\textstyle\sum}_{\theta^{\prime}\in\Theta}p_{\theta^{\prime},\lambda_{t-1}}\mu_{t-1}(\theta^{\prime})}{1-{\textstyle\sum}_{\theta^{\prime}\in\Theta}p_{\theta^{\prime},\lambda_{t-1}}\mu_{t-1}(\theta^{\prime})}}
\mu_{t-1}(\theta)
)\nu_{t-1} \numberthis
\end{align*}
Therefore it is straightforward to define the functions $\alpha_{t}=A(\mu_{t-1},\lambda_{t-1},\nu_{t-1},\nu_{t})$, 
$\beta_{t}=B(\mu_{t-1},\lambda_{t-1},\nu_{t-1},\nu_{t})$, as well as
$\tilde{\mu}_{t}=\mu_{t}-\alpha_{t}-\beta_{t}$,
where $\alpha_{0}=\beta_{0}=0$ and:
\begin{align*}
\alpha_{t}&(\theta)
=
\alpha_{t-1}(\theta)
-
\mu_{t-1}(\theta) \\
&\cdot\medmath{\frac{p_{\theta,\lambda_{t-1}}-{\textstyle\sum}_{\theta^{\prime}\in\Theta}p_{\theta^{\prime},\lambda_{t-1}}\mu_{t-1}(\theta^{\prime})}{1-{\textstyle\sum}_{\theta^{\prime}\in\Theta}p_{\theta^{\prime},\lambda_{t-1}}\mu_{t-1}(\theta^{\prime})}}
\nu_{t-1}\nu_{t} \numberthis \\
\beta_{t}&(\theta)
=
\beta_{t-1}(\theta)
+
\mu_{t-1}(\theta) \\
&\cdot\medmath{\frac{p_{\theta,\lambda_{t-1}}-{\textstyle\sum}_{\theta^{\prime}\in\Theta}p_{\theta^{\prime},\lambda_{t-1}}\mu_{t-1}(\theta^{\prime})}{{\textstyle\sum}_{\theta^{\prime}\in\Theta}p_{\theta^{\prime},\lambda_{t-1}}\mu_{t-1}(\theta^{\prime})}}
(1-\nu_{t})\nu_{t-1} \numberthis
\end{align*}
Finally, for $\tilde{\mu}_{t}$ observe that:
\begin{align*}
\alpha_{t}&+\beta_{t}= \\
&
{\textstyle\sum}_{t^{\prime}=1}^{t}(\mathbb{E}_{p,q}[\mu_{t^{\prime}}-\mu_{t^{\prime}-1}|\lambda_{t^{\prime}-1},\mu_{t^{\prime}-1},\nu_{t^{\prime}-1},\nu_{t^{\prime}}]) \numberthis
\end{align*}
therefore the difference between two steps is:
\begin{align*}
\tilde{\mu}_{t}-\tilde{\mu}_{t-1}&=\mu_{t}-\mu_{t-1} \\
&-\mathbb{E}_{p,q}[\mu_{t}-\mu_{t-1}|\lambda_{t-1},\mu_{t-1},\nu_{t-1},\nu_{t}] \numberthis
\end{align*}
hence\textemdash taking expectations\textemdash we can write:
\begin{align*}
\mathbb{E}_{p,q}&[\tilde{\mu}_{t}-\tilde{\mu}_{t-1}|\lambda_{t-1},\mu_{t-1},\nu_{t-1}]=0 \numberthis \\
\Rightarrow~&
\mathbb{E}_{p,q}[\tilde{\mu}_{t}|\lambda_{t-1},\mu_{t-1},\nu_{t-1}]
=
\tilde{\mu}_{t-1} \numberthis
\end{align*}

\Paste{thm_optimal_value}
Each of the $Q$-factors for decisions is given by:
\begin{align*}
&\bar{Q}_{\hat{\theta}}(\mu_{t},\nu_{t};\eta) \numberthis \\
\doteq\pix&
\mathbb{E}_{p,q}[
\ell(\lambda_{0:\tau-1},\tau,\hat{\theta};\eta)
|\lambda_{0:t-1},\tau=t,\hat{\theta},\mu_{t},\nu_{t}] \\
&-
\textstyle\sum_{t^{\prime}=0}^{t-1}
\eta_{\text{\pix c},\lambda_{t^{\prime}}}
c_{\lambda_{t^{\prime}}} \numberthis \\
=\pix&
\mathbb{E}_{p,q}\big[
{\textstyle\sum}_{\theta^{\prime}\in\Theta}
\eta_{\text{\pix a},\theta^{\prime}}
\mathds{1}_{\{\theta=\theta^{\prime},\theta\neq\hat{\theta},\tau<\delta\}} \\
&+
{\textstyle\sum}_{\theta^{\prime}\in\Theta}
\eta_{\text{\pix b},\theta^{\prime}}
\mathds{1}_{\{\theta=\theta^{\prime},\tau=\delta\}} \\
&+
\textstyle\sum_{t^{\prime}=0}^{\tau-1}
\eta_{\text{\pix c},\lambda_{t^{\prime}}}
c_{\lambda_{t^{\prime}}}
|\lambda_{0:t-1},\tau=t,\hat{\theta},\mu_{t},\nu_{t}\big] \\
&-
\textstyle\sum_{t^{\prime}=0}^{t-1}
\eta_{\text{\pix c},\lambda_{t^{\prime}}}
c_{\lambda_{t^{\prime}}} \numberthis \\
=\pix&
\mathbb{E}_{p,q}\big[{\textstyle\sum}_{\theta^{\prime}\in\Theta}
\eta_{\text{\pix a},\theta^{\prime}}
\mathds{1}_{\{\theta=\theta^{\prime},\theta\neq\hat{\theta},t<\delta\}} \\
&+
{\textstyle\sum}_{\theta^{\prime}\in\Theta}
\eta_{\text{\pix b},\theta^{\prime}}
\mathds{1}_{\{\theta=\theta^{\prime},t=\delta\}}
|\hat{\theta},\mu_{t},\nu_{t}\big] \numberthis \\
=\pix&
\nu_{t}
{\textstyle\sum}_{\theta^{\prime}\in\Theta,\theta\neq\hat{\theta}}
\eta_{\text{\pix a},\theta^{\prime}}
\mu_{t}(\theta^{\prime}) \\
&+
(1-\nu_{t}){\textstyle\sum}_{\theta^{\prime}\in\Theta}
\eta_{\text{\pix b},\theta^{\prime}}
\mu_{t}(\theta^{\prime}) \numberthis
\end{align*}
For acquisitions, first observe that:
\begin{align*}
&Q_{\lambda_{t}}(\mu_{t},\nu_{t};\eta) \numberthis \\
\doteq\pix&
\mathbb{E}_{p,q}[
V(\mu_{t+1},\nu_{t+1};\eta)
|\lambda_{t},\mu_{t},\nu_{t}]
+
\eta_{\text{\pix c},\lambda_{t}}
c_{\lambda_{t}} \numberthis \\
=\pix&
(1-\nu_{t})
\mathbb{E}_{p,q}[
V(\mu_{t+1},\nu_{t+1};\eta)
|\lambda_{t},\mu_{t},\nu_{t}=0] \\
&+
\mathbb{E}_{p,q}[
V(\mu_{t+1},\nu_{t+1};\eta)
|\lambda_{t},\mu_{t},\nu_{t}=1]
\nu_{t} \\
&+
\eta_{\text{\pix c},\lambda_{t}}
c_{\lambda_{t}} \numberthis \\
=\pix&
(1-\nu_{t})V(\mu_{t},0;\eta)
+
\eta_{\text{\pix c},\lambda_{t}}
c_{\lambda_{t}} \\
&+
(
\mathbb{E}_{p,q}[V(
(1-\nu_{t+1})\bar{M}(\lambda_{t},\mu_{t}) \\
&+
\nu_{t+1}M(\lambda_{t},\mu_{t},\omega_{t+1})
,
1
;\eta
)
|\lambda_{t},\mu_{t},\nu_{t}=1]
)\nu_{t} \numberthis
\end{align*}
For the expectation term:
\begin{align*}
&\mathbb{E}_{p,q}[V(
(1-\nu_{t+1})\bar{M}(\lambda_{t},\mu_{t}) \\
&+
\nu_{t+1}M(\lambda_{t},\mu_{t},\omega_{t+1})
,
1
;\eta
)
|\lambda_{t},\mu_{t},\nu_{t}=1] \numberthis \\
=\pix&
{\textstyle\sum}_{\omega^{\prime}_{t+1}\in\Omega}(
\mathbb{P}_{p,q}\{\nu_{t+1}=1,\omega_{t+1}|\lambda_{t},\mu_{t},\nu_{t}=1\} \\
&\cdot V(M(\lambda_{t},\mu_{t},\omega_{t+1}),1;\eta))+ \\
&
\mathbb{P}_{p}\{\nu_{t+1}=0|\lambda_{t},\mu_{t},\nu_{t}=1\}
V(\bar{M}(\lambda_{t},\mu_{t}),0;\eta) \numberthis \\
=\pix&
{\textstyle\sum}_{\omega^{\prime}_{t+1}\in\Omega}(
V(M(\lambda_{t},\mu_{t},\omega_{t+1}),1;\eta) \\
&\cdot{\textstyle\sum}_{\theta^{\prime}\in\Theta}(
\mathbb{P}_{p}\{\theta^{\prime},\nu_{t+1}=1|\lambda_{t},\mu_{t},\nu_{t}=1\} \\
&\cdot\mathbb{P}_{q}\{\omega_{t+1}|\theta^{\prime},\lambda_{t},\nu_{t+1}=1\}))+ \\
&
\mathbb{P}_{p}\{\nu_{t+1}=0|\lambda_{t},\mu_{t},\nu_{t}=1\}
V(\bar{M}(\lambda_{t},\mu_{t}),0;\eta) \numberthis \\
=\pix&
{\textstyle\sum}_{\omega^{\prime}_{t+1}\in\Omega}(
V(M(\lambda_{t},\mu_{t},\omega_{t+1}),1;\eta)\\
&\cdot{\textstyle\sum}_{\theta^{\prime}\in\Theta}(
\mathbb{P}_{p}\{\nu_{t+1}=1|\theta,\lambda_{t},\nu_{t}=1\} \\
&\cdot\mathbb{P}_{q}\{\omega_{t+1}|\lambda_{t},\theta^{\prime},\nu_{t+1}=1\}\mu_{t}(\theta))) \\
&+
V(\bar{M}(\lambda_{t},\mu_{t}),0;\eta)
{\textstyle\sum}_{\theta^{\prime}\in\Theta}p_{\theta^{\prime},\lambda_{t}}\mu_{t}(\theta^{\prime}) \numberthis \\
=\pix&
{\textstyle\sum}_{\omega^{\prime}_{t+1}\in\Omega}(
V(M(\lambda_{t},\mu_{t},\omega_{t+1}),1;\eta) \\
&\cdot{\textstyle\sum}_{\theta^{\prime}\in\Theta}(1-p_{\theta^{\prime},\lambda_{t}})q_{\theta^{\prime},\lambda_{t}}(\omega_{t+1})\mu_{t}(\theta^{\prime})) \\
&+
V(\bar{M}(\lambda_{t},\mu_{t}),0;\eta)
{\textstyle\sum}_{\theta^{\prime}\in\Theta}p_{\theta^{\prime},\lambda_{t}}\mu_{t}(\theta^{\prime}) \numberthis
\end{align*}
Therefore each $Q$-factor for acquisition is given by:
\begin{align*}
Q_{\lambda_{t}}(&\mu_{t},\nu_{t};\eta)
=
(1-\nu_{t})V(\mu_{t},0;\eta)
+
\eta_{\text{\pix c},\lambda_{t}}
c_{\lambda_{t}} \\
&+
\big(V(\bar{M}(\lambda_{t},\mu_{t}),0;\eta)
{\textstyle\sum}_{\theta^{\prime}\in\Theta}p_{\theta^{\prime},\lambda_{t}}\mu_{t}(\theta^{\prime}) \\
&+
{\textstyle\sum}_{\omega^{\prime}_{t+1}\in\Omega}(V(M(\lambda_{t},\mu_{t},\omega^{\prime}_{t+1}),1;\eta) \\
&~~\cdot{\textstyle\sum}_{\theta^{\prime}\in\Theta}(1-p_{\theta^{\prime},\lambda_{t}})q_{\theta^{\prime},\lambda_{t}}(\omega^{\prime}_{t+1})\mu_{t}(\theta^{\prime}))
\big)\nu_{t} \numberthis
\end{align*}
For the contractive property, we want that $\|\mathbb{B}V^{i}-\mathbb{B}V^{j}\|\leq\gamma\|V^{i}-V^{j}\|$ for some $\gamma<1$, but where we do \textit{not} have the benefit of an explicit discount factor $\gamma$ for this purpose. For notational brevity, in the following we omit the functional dependence of value functions and $Q$-factors on $\eta$:
\begin{align*}
&|(\mathbb{B}V^{i})(\mu_{t},\nu_{t})-(\mathbb{B}V^{j})(\mu_{t},\nu_{t})| \numberthis \\
=\pix&|
\min\{
{\textstyle\inf}_{\hat{\theta}^{\prime}\in\Theta}
\bar{Q}^{i}_{\hat{\theta}^{\prime}}(\mu_{t},\nu_{t})
,
{\textstyle\inf}_{\lambda^{\prime}_{t}\in\Lambda}
Q^{i}_{\lambda^{\prime}_{t}}(\mu_{t},\nu_{t})
\} \\
&\scalebox{1}{$-$}
\min\{
{\textstyle\inf}_{\hat{\theta}^{\prime}\in\Theta}
\bar{Q}^{j}_{\hat{\theta}^{\prime}}(\mu_{t},\nu_{t})
,
{\textstyle\inf}_{\lambda^{\prime}_{t}\in\Lambda}
Q^{j}_{\lambda^{\prime}_{t}}(\mu_{t},\nu_{t})
\}
| \numberthis \\
=\pix&|
\min\{
{\textstyle\inf}_{\hat{\theta}\in\Theta}(
\nu_{t}
{\textstyle\sum}_{\theta^{\prime}\in\Theta,\theta\neq\hat{\theta}}
\eta_{\text{\pix a},\theta^{\prime}}
\mu_{t}(\theta^{\prime})
+
(1-\nu_{t}) \\
&\cdot{\textstyle\sum}_{\theta^{\prime}\in\Theta}
\eta_{\text{\pix b},\theta^{\prime}}
\mu_{t}(\theta^{\prime})
)
,
{\textstyle\inf}_{\lambda^{\prime}_{t}\in\Lambda}(
(1-\nu_{t})V^{i}(\mu_{t},0) \\
&+
\big(
{\textstyle\sum}_{\omega^{\prime}_{t+1}\in\Omega}(
V^{i}(M(\lambda^{\prime}_{t},\mu_{t},\omega_{t+1}),1) \\
&\cdot{\textstyle\sum}_{\theta^{\prime}\in\Theta}(1-p_{\theta^{\prime},\lambda^{\prime}_{t}})q_{\theta^{\prime},\lambda^{\prime}_{t}}(\omega_{t+1})\mu_{t}(\theta^{\prime})) \\
&+
V^{i}(\bar{M}(\lambda^{\prime}_{t},\mu_{t}),0)
{\textstyle\sum}_{\theta^{\prime}\in\Theta}p_{\theta^{\prime},\lambda^{\prime}_{t}}\mu_{t}(\theta^{\prime})
\big)\nu_{t} \\
&+
\eta_{\text{\pix c},\lambda^{\prime}_{t}}
c_{\lambda^{\prime}_{t}}
)
\} \\
&-\min\{
{\textstyle\inf}_{\hat{\theta}\in\Theta}(
\nu_{t}
{\textstyle\sum}_{\theta^{\prime}\in\Theta,\theta\neq\hat{\theta}}
\eta_{\text{\pix a},\theta^{\prime}}
\mu_{t}(\theta^{\prime})
+
(1-\nu_{t}) \\
&\cdot{\textstyle\sum}_{\theta^{\prime}\in\Theta}
\eta_{\text{\pix b},\theta^{\prime}}
\mu_{t}(\theta^{\prime})
)
,
{\textstyle\inf}_{\lambda^{\prime}_{t}\in\Lambda}(
(1-\nu_{t})V^{j}(\mu_{t},0) \\
&+
\big(
{\textstyle\sum}_{\omega^{\prime}_{t+1}\in\Omega}(
V^{j}(M(\lambda^{\prime}_{t},\mu_{t},\omega_{t+1}),1) \\
&\cdot{\textstyle\sum}_{\theta^{\prime}\in\Theta}(1-p_{\theta^{\prime},\lambda^{\prime}_{t}})q_{\theta^{\prime},\lambda^{\prime}_{t}}(\omega_{t+1})\mu_{t}(\theta^{\prime})) \\
&+
V^{j}(\bar{M}(\lambda^{\prime}_{t},\mu_{t}),0)
{\textstyle\sum}_{\theta^{\prime}\in\Theta}p_{\theta^{\prime},\lambda^{\prime}_{t}}\mu_{t}(\theta^{\prime})
\big)\nu_{t} \\
&+
\eta_{\text{\pix c},\lambda^{\prime}_{t}}
c_{\lambda^{\prime}_{t}}
)
\}| \numberthis \\
\leq\pix&
|
{\textstyle\inf}_{\lambda^{\prime}_{t}\in\Lambda}(
(1-\nu_{t})V^{i}(\mu_{t},0)
+
\eta_{\text{\pix c},\lambda^{\prime}_{t}}
c_{\lambda^{\prime}_{t}} \\
&+
\big(
{\textstyle\sum}_{\omega^{\prime}_{t+1}\in\Omega}(
V^{i}(M(\lambda^{\prime}_{t},\mu_{t},\omega_{t+1}),1) \\
&\cdot{\textstyle\sum}_{\theta^{\prime}\in\Theta}(1-p_{\theta^{\prime},\lambda^{\prime}_{t}})q_{\theta^{\prime},\lambda^{\prime}_{t}}(\omega_{t+1})\mu_{t}(\theta^{\prime})) \\
&+
V^{i}(\bar{M}(\lambda^{\prime}_{t},\mu_{t}),0)
{\textstyle\sum}_{\theta^{\prime}\in\Theta}p_{\theta^{\prime},\lambda^{\prime}_{t}}\mu_{t}(\theta^{\prime})
\big)\nu_{t}
) \\
&-
{\textstyle\inf}_{\lambda^{\prime}_{t}\in\Lambda}(
(1-\nu_{t})V^{j}(\mu_{t},0)
+
\eta_{\text{\pix c},\lambda^{\prime}_{t}}
c_{\lambda^{\prime}_{t}} \\
&+
\big(
{\textstyle\sum}_{\omega^{\prime}_{t+1}\in\Omega}(
V^{j}(M(\lambda^{\prime}_{t},\mu_{t},\omega_{t+1}),1) \\
&\cdot{\textstyle\sum}_{\theta^{\prime}\in\Theta}(1-p_{\theta^{\prime},\lambda^{\prime}_{t}})q_{\theta^{\prime},\lambda^{\prime}_{t}}(\omega_{t+1})\mu_{t}(\theta^{\prime})) \\
&+
V^{j}(\bar{M}(\lambda^{\prime}_{t},\mu_{t}),0)
{\textstyle\sum}_{\theta^{\prime}\in\Theta}p_{\theta^{\prime},\lambda^{\prime}_{t}}\mu_{t}(\theta^{\prime})
\big)\nu_{t}
)
| \numberthis \\
=\pix&
|
(1-\nu_{t})V^{i}(\mu_{t},0)
+
\eta_{\text{\pix c},\lambda^{*}_{t}}
c_{\lambda^{*}_{t}} \\
&+
\big(
{\textstyle\sum}_{\omega^{\prime}_{t+1}\in\Omega}(
V^{i}(M(\lambda^{*}_{t},\mu_{t},\omega_{t+1}),1) \\
&\cdot{\textstyle\sum}_{\theta^{\prime}\in\Theta}(1-p_{\theta^{\prime},\lambda^{*}_{t}})q_{\theta^{\prime},\lambda^{*}_{t}}(\omega_{t+1})\mu_{t}(\theta^{\prime})) \\
&+
V^{i}(\bar{M}(\lambda^{*}_{t},\mu_{t}),0)
{\textstyle\sum}_{\theta^{\prime}\in\Theta}p_{\theta^{\prime},\lambda^{*}_{t}}\mu_{t}(\theta^{\prime})
\big)\nu_{t} \\
&-
{\textstyle\inf}_{\lambda^{\prime}_{t}\in\Lambda}(
(1-\nu_{t})V^{j}(\mu_{t},0)
+
\eta_{\text{\pix c},\lambda^{\prime}_{t}}
c_{\lambda^{\prime}_{t}} \\
&+
\big(
{\textstyle\sum}_{\omega^{\prime}_{t+1}\in\Omega}(
V^{j}(M(\lambda^{\prime}_{t},\mu_{t},\omega_{t+1}),1) \\
&\cdot{\textstyle\sum}_{\theta^{\prime}\in\Theta}(1-p_{\theta^{\prime},\lambda^{\prime}_{t}})q_{\theta^{\prime},\lambda^{\prime}_{t}}(\omega_{t+1})\mu_{t}(\theta^{\prime})) \\
&+
V^{j}(\bar{M}(\lambda^{\prime}_{t},\mu_{t}),0)
{\textstyle\sum}_{\theta^{\prime}\in\Theta}p_{\theta^{\prime},\lambda^{\prime}_{t}}\mu_{t}(\theta^{\prime})
\big)\nu_{t}
)
| \numberthis \\
\leq\pix&
|
(1-\nu_{t})V^{i}(\mu_{t},0)
+
\eta_{\text{\pix c},\lambda^{*}_{t}}
c_{\lambda^{*}_{t}} \\
&+
\big(
{\textstyle\sum}_{\omega^{\prime}_{t+1}\in\Omega}(V^{i}(M(\lambda^{*}_{t},\mu_{t},\omega_{t+1}),1) \\
&\cdot{\textstyle\sum}_{\theta^{\prime}\in\Theta}(1-p_{\theta^{\prime},\lambda^{*}_{t}})q_{\theta^{\prime},\lambda^{*}_{t}}(\omega_{t+1})\mu_{t}(\theta^{\prime})) \\
&+
V^{i}(\bar{M}(\lambda^{*}_{t},\mu_{t}),0)
{\textstyle\sum}_{\theta^{\prime}\in\Theta}p_{\theta^{\prime},\lambda^{*}_{t}}\mu_{t}(\theta^{\prime})
\big)\nu_{t} \\
&-
(1-\nu_{t})V^{j}(\mu_{t},0)
-
\eta_{\text{\pix c},\lambda^{*}_{t}}
c_{\lambda^{*}_{t}} \\
&-
\big(
{\textstyle\sum}_{\omega^{\prime}_{t+1}\in\Omega}(
V^{j}(M(\lambda^{*}_{t},\mu_{t},\omega_{t+1}),1) \\
&\cdot{\textstyle\sum}_{\theta^{\prime}\in\Theta}(1-p_{\theta^{\prime},\lambda^{*}_{t}})q_{\theta^{\prime},\lambda^{*}_{t}}(\omega_{t+1})\mu_{t}(\theta^{\prime})) \\
&-
V^{j}(\bar{M}(\lambda^{*}_{t},\mu_{t}),0)
{\textstyle\sum}_{\theta^{\prime}\in\Theta}p_{\theta^{\prime},\lambda^{*}_{t}}\mu_{t}(\theta^{\prime})
\big)\nu_{t}
| \numberthis \\
=\pix&
|
{\textstyle\sum}_{\omega^{\prime}_{t+1}\in\Omega}(
(
V^{i}(M(\lambda^{*}_{t},\mu_{t},\omega_{t+1}),1) \\
&-
V^{j}(M(\lambda^{*}_{t},\mu_{t},\omega_{t+1}),1)
) \\
&\cdot{\textstyle\sum}_{\theta^{\prime}\in\Theta}(1-p_{\theta^{\prime},\lambda^{*}_{t}})q_{\theta^{\prime},\lambda^{*}_{t}}(\omega_{t+1})\mu_{t}(\theta^{\prime})
\nu_{t})
| \numberthis \\
\leq\pix&
|
(1-\textstyle{\inf}_{\theta^{\prime}\in\Theta,\lambda^{\prime}\in\Lambda}p_{\theta^{\prime},\lambda^{\prime}}) \\
&\cdot{\textstyle\sum}_{\omega^{\prime}_{t+1}\in\Omega}(
(
V^{i}(M(\lambda^{*}_{t},\mu_{t},\omega_{t+1}),1) \\
&-
V^{j}(M(\lambda^{*}_{t},\mu_{t},\omega_{t+1}),1)
) \\
&\cdot{\textstyle\sum}_{\theta^{\prime}\in\Theta}q_{\theta^{\prime},\lambda^{*}_{t}}(\omega_{t+1})\mu_{t}(\theta^{\prime})
\nu_{t})
| \numberthis \\
\leq\pix&
(1-\textstyle{\inf}_{\theta^{\prime}\in\Theta,\lambda^{\prime}\in\Lambda}p_{\theta^{\prime},\lambda^{\prime}}) \\
&\cdot\textstyle{\sup}_{\mu^{\prime}_{t+1}\in\Delta(\Theta)}
|
V^{i}(\mu^{\prime}_{t+1},1)
-
V^{j}(\mu^{\prime}_{t+1},1)
| \numberthis \\
\leq\pix&
\gamma
\|
V^{i}
-
V^{j}
\| \numberthis
\end{align*}
where in the fourth equality \smash{$\lambda^{*}_{t}\doteq{\textstyle\arg\inf}_{\lambda^{\prime}_{t}\in\Lambda}Q^{k}_{\lambda^{\prime}_{t}}(\mu_{t},\nu_{t})$} in which \smash{$k\doteq{\textstyle\arg\inf}_{k^{\prime}\in\{i,j\}}{\textstyle\inf}_{\lambda^{\prime}_{t}\in\Lambda}Q^{k^{\prime}}_{\lambda^{\prime}_{t}}(\mu_{t},\nu_{t})$}, and in the last step $\gamma\doteq1-\textstyle{\inf}_{\theta^{\prime}\in\Theta,\lambda^{\prime}\in\Lambda}p_{\theta^{\prime},\lambda^{\prime}}<1$, and we also used the fact that $V(\mu_{t},0)=\bar{Q}_{\hat{\theta}}(\mu_{t},0)={\textstyle\sum}_{\theta^{\prime}\in\Theta}
\eta_{\text{\pix b},\theta^{\prime}}
\mu_{t}(\theta^{\prime})$.

For the uniqueness property, consider two such fixed points $V^{*}$ and $V^{\prime*}$. But $\|V^{*}-V^{\prime*}\|=\|\mathbb{B}V^{*}-\mathbb{B}V^{\prime*}\|\leq\gamma\|V^{*}-V^{\prime*}\|$, therefore it must the case that $\|V^{*}-V^{\prime*}\|=0$.

\Paste{thm_continue_terminate}
We first show that $V^{*}$ is concave. Since $V^{*}$ is the limit of successive approximations by application of $\mathbb{B}$, we simply want to show if $V$ is concave that $\mathbb{B}V$ is then concave. Suppose $V$ is concave. Since $\mathbb{B}V$ is the minimum between ${\textstyle\inf}_{\hat{\theta}^{\prime}\in\Theta}
\bar{Q}_{\hat{\theta}^{\prime}}(\mu_{t},\nu_{t};\eta)$ and ${\textstyle\inf}_{\lambda^{\prime}_{t}\in\Lambda}
Q_{\lambda^{\prime}_{t}}(\mu_{t},\nu_{t};\eta)$ and the former clearly concave, it remains to show that each $Q_{\lambda_{t}}$ in the latter is concave. This is obvious for $\nu_{t}=0$ since $V(\mu_{t},0)=\bar{Q}_{\hat{\theta}}(\mu_{t},0)$ is concave. For $\nu_{t}=1$, we want that \smash{$
{\textstyle\sum}_{\omega^{\prime}_{t+1}\in\Omega}(
V(M(\lambda_{t},\mu_{t},\omega_{t+1}),1;\eta)
{\textstyle\sum}_{\theta^{\prime}\in\Theta}(1-$}
\smash{$p_{\theta^{\prime},\lambda_{t}})q_{\theta^{\prime},\lambda_{t}}(\omega_{t+1})\mu_{t}(\theta^{\prime}))
$}
be concave. Let $\upsilon\in(0,1)$. We similarly omit functional dependence on $\eta$ for brevity:
\begin{align*}
&\upsilon{\textstyle\sum}_{\omega^{\prime}_{t+1}\in\Omega}(
V(M(\lambda_{t},\mu_{t},\omega_{t+1}),1) \\
&\cdot{\textstyle\sum}_{\theta^{\prime}\in\Theta}(1-p_{\theta^{\prime},\lambda_{t}})q_{\theta^{\prime},\lambda_{t}}(\omega_{t+1})\mu_{t}(\theta^{\prime})) \\
&+
(1-\upsilon){\textstyle\sum}_{\omega^{\prime}_{t+1}\in\Omega}(
V(M(\lambda_{t},\mu^{\prime}_{t},\omega_{t+1}),1) \\
&\cdot{\textstyle\sum}_{\theta^{\prime}\in\Theta}(1-p_{\theta^{\prime},\lambda_{t}})q_{\theta^{\prime},\lambda_{t}}(\omega_{t+1})\mu^{\prime}_{t}(\theta^{\prime})) \numberthis \\
=\pix&
{\textstyle\sum}_{\omega^{\prime}_{t+1}\in\Omega}((
\upsilon V(M(\lambda_{t},\mu_{t},\omega_{t+1}),1) \\
&\cdot{\textstyle\sum}_{\theta^{\prime}\in\Theta}(1-p_{\theta^{\prime},\lambda_{t}})q_{\theta^{\prime},\lambda_{t}}(\omega_{t+1})\mu_{t}(\theta^{\prime}) \\
&/
(\upsilon {\textstyle\sum}_{\theta^{\prime}\in\Theta}(1-p_{\theta^{\prime},\lambda_{t}})q_{\theta^{\prime},\lambda_{t}}(\omega_{t+1})\mu_{t}(\theta^{\prime}) \\
&+
(1-\upsilon){\textstyle\sum}_{\theta^{\prime}\in\Theta}(1-p_{\theta^{\prime},\lambda_{t}})q_{\theta^{\prime},\lambda_{t}}(\omega_{t+1})\mu^{\prime}_{t}(\theta^{\prime})) \\
&+
(1-\upsilon)V(M(\lambda_{t},\mu^{\prime}_{t},\omega_{t+1}),1) \\
&\cdot{\textstyle\sum}_{\theta^{\prime}\in\Theta}(1-p_{\theta^{\prime},\lambda_{t}})q_{\theta^{\prime},\lambda_{t}}(\omega_{t+1})\mu^{\prime}_{t}(\theta^{\prime}) \\
&/
(\upsilon {\textstyle\sum}_{\theta^{\prime}\in\Theta}(1-p_{\theta^{\prime},\lambda_{t}})q_{\theta^{\prime},\lambda_{t}}(\omega_{t+1})\mu_{t}(\theta^{\prime}) \\
&+
(1-\upsilon){\textstyle\sum}_{\theta^{\prime}\in\Theta}(1-p_{\theta^{\prime},\lambda_{t}})q_{\theta^{\prime},\lambda_{t}}(\omega_{t+1})\mu^{\prime}_{t}(\theta^{\prime}))) \\
&\cdot
(\upsilon {\textstyle\sum}_{\theta^{\prime}\in\Theta}(1-p_{\theta^{\prime},\lambda_{t}})q_{\theta^{\prime},\lambda_{t}}(\omega_{t+1})\mu_{t}(\theta^{\prime}) \\
&+
(1\scalebox{1}{$-$}\upsilon){\textstyle\sum}_{\theta^{\prime}\in\Theta}(1-p_{\theta^{\prime},\lambda_{t}})q_{\theta^{\prime},\lambda_{t}}(\omega_{t+1})\mu^{\prime}_{t}(\theta^{\prime}))) \numberthis \\
\leq\pix&
{\textstyle\sum}_{\omega^{\prime}_{t+1}\in\Omega}(
V(
(
\upsilon M(\lambda_{t},\mu_{t},\omega_{t+1}) \\
&\cdot{\textstyle\sum}_{\theta^{\prime}\in\Theta}(1-p_{\theta^{\prime},\lambda_{t}})q_{\theta^{\prime},\lambda_{t}}(\omega_{t+1})\mu_{t}(\theta^{\prime}) \\
&+
(1-\upsilon) M(\lambda_{t},\mu^{\prime}_{t},\omega_{t+1}) \\
&\cdot {\textstyle\sum}_{\theta^{\prime}\in\Theta}(1-p_{\theta^{\prime},\lambda_{t}})q_{\theta^{\prime},\lambda_{t}}(\omega_{t+1})\mu^{\prime}_{t}(\theta^{\prime})
) \\
&/
(\upsilon {\textstyle\sum}_{\theta^{\prime}\in\Theta}(1-p_{\theta^{\prime},\lambda_{t}})q_{\theta^{\prime},\lambda_{t}}(\omega_{t+1})\mu_{t}(\theta^{\prime}) \\
&+
(1-\upsilon){\textstyle\sum}_{\theta^{\prime}\in\Theta}(1-p_{\theta^{\prime},\lambda_{t}})q_{\theta^{\prime},\lambda_{t}}(\omega_{t+1})\mu^{\prime}_{t}(\theta^{\prime}))
,1) \\
&\cdot
(\upsilon {\textstyle\sum}_{\theta^{\prime}\in\Theta}(1-p_{\theta^{\prime},\lambda_{t}})q_{\theta^{\prime},\lambda_{t}}(\omega_{t+1})\mu_{t}(\theta^{\prime}) \\
&+
(1\scalebox{1}{$-$}\upsilon){\textstyle\sum}_{\theta^{\prime}\in\Theta}(1-p_{\theta^{\prime},\lambda_{t}})q_{\theta^{\prime},\lambda_{t}}(\omega_{t+1})\mu^{\prime}_{t}(\theta^{\prime}))
) \numberthis \\
=\pix&
{\textstyle\sum}_{\omega^{\prime}_{t+1}\in\Omega}(
V(M(\upsilon\mu_{t}+(1-\upsilon)\mu^{\prime}_{t})) \\
&\cdot
{\textstyle\sum}_{\theta^{\prime}\in\Theta}(
(1-p_{\theta^{\prime},\lambda_{t}})q_{\theta^{\prime},\lambda_{t}}(\omega_{t+1}) \\
&\cdot
(\upsilon\mu_{t}(\theta^{\prime})
+
(1-\upsilon)\mu^{\prime}_{t}(\theta^{\prime})
)
)
) \numberthis
\end{align*}
Now, $V^{*}$ is simply the limit of successive approximations by application of $\mathbb{B}$, so by induction $V^{*}$ is concave. Finally, each $Q^{*}_{\lambda_{t}}$ and therefore $Q^{*}$ is concave since $V^{*}$ is concave.

For the inequality, note if $\nu_{t}=1$ then \smash{$\bar{Q}_{\hat{\theta}}$} at each vertex is simply zero for any choice of $\hat{\theta}$. But clearly $Q^{*}_{\lambda_{t}}$ is at least $\eta_{\text{\pix c},\lambda_{t}}c_{\lambda_{t}}$ for any choice of $\lambda_{t}$, so it must be true that $Q^{*}>\bar{Q}$. Finally, consider the intersection (if any) of $Q^{*}$ and $\bar{Q}_{\hat{\theta}}$ when $\nu_{t}=1$, for any $\hat{\theta}$. Let $\mu_{t},\mu_{t}^{\prime}\in\Delta(\Theta)$ be two points for which \smash{${\hat{\theta}=\textstyle\arg\inf}_{\hat{\theta}^{\prime}\in\Theta}
\bar{Q}_{\hat{\theta}^{\prime}}(\cdot,\nu_{t};\eta)$}. Since the former is concave and the latter is affine, we can write:
\begin{align*}
&\bar{Q}_{\hat{\theta}}(\upsilon\mu_{t}+(1-\upsilon)\mu_{t},1;\eta) \numberthis \\
=\pix&
\upsilon\bar{Q}_{\hat{\theta}}(\mu_{t},1;\eta)
+
(1-\upsilon)\bar{Q}_{\hat{\theta}}(\mu_{t},1;\eta) \numberthis \\
=\pix&
\upsilon V^{*}(\mu_{t},1;\eta)
+
(1-\upsilon) V^{*}(\mu_{t},1;\eta) \numberthis \\
\leq\pix&
V^{*}(\upsilon\mu_{t}+(1-\upsilon)\mu_{t},1;\eta) \numberthis \\
\leq\pix&
\bar{Q}(\upsilon\mu_{t}+(1-\upsilon)\mu_{t},1;\eta) \numberthis \\
\leq\pix&
\bar{Q}_{\hat{\theta}}(\upsilon\mu_{t}+(1-\upsilon)\mu_{t},1;\eta) \numberthis
\end{align*}
for $\upsilon\in(0,1)$, hence the set $\bar{Q}_{\hat{\theta}}<Q^{*}$ is convex. Finally, the overall termination set $\mathcal{T}(\eta)$ is the union of $|\Theta|$ such regions. For completeness, consider the other (trivial) case where $\nu_{t}=0$; clearly $Q^{*}_{\lambda_{t}}=\bar{Q}+\eta_{\text{\pix c},\lambda_{t}}c_{\lambda_{t}}$, so convexity is automatic and there is no intersection (i.e. $\mathcal{T}(\eta)$ is empty).

\Paste{thm_surprise_suspense}
Each optimal $Q$-factor for acquisitions is given by:
\begin{align*}
&Q^{*}_{\lambda_{t}}(\mu_{t},\nu_{t};\eta) \numberthis \\
=\pix&
(1-\nu_{t})V^{*}(\mu_{t},0;\eta)
+
\eta_{\text{\pix c},\lambda_{t}}
c_{\lambda_{t}} \\
&+
(
\mathbb{E}_{p,q}[V^{*}(
(1-\nu_{t+1})\bar{M}(\lambda_{t},\mu_{t}) \\
&+
\nu_{t+1}M(\lambda_{t},\mu_{t},\omega_{t+1})
,
1
;\eta
)
|\lambda_{t},\mu_{t},\nu_{t}=1]
)\nu_{t} \numberthis \\
=\pix&
(1-\nu_{t})V^{*}(\mu_{t},0;\eta)
+
\eta_{\text{\pix c},\lambda_{t}}
c_{\lambda_{t}} \\
&+
(V^{*}(\bar{M}(\lambda_{t},\mu_{t}),0;\eta)
{\textstyle\sum}_{\theta^{\prime}\in\Theta}p_{\theta^{\prime},\lambda_{t}}\mu_{t}(\theta^{\prime}) \\
&+
{\textstyle\sum}_{\omega^{\prime}_{t+1}\in\Omega}(V^{*}(M(\lambda_{t},\mu_{t},\omega^{\prime}_{t+1}),1;\eta) \\
&\cdot{\textstyle\sum}_{\theta^{\prime}\in\Theta}(1-p_{\theta^{\prime},\lambda_{t}})q_{\theta^{\prime},\lambda_{t}}(\omega^{\prime}_{t+1})\mu_{t}(\theta^{\prime}))
)\nu_{t} \numberthis
\end{align*}
Note that the expectation term can also be expressed:
\begin{align*}
&
\mathbb{E}_{p,q}[V^{*}(
(1-\nu_{t+1})\bar{M}(\lambda_{t},\mu_{t}) \\
&+
\nu_{t+1}M(\lambda_{t},\mu_{t},\omega_{t+1})
,
1
;\eta
)
|\lambda_{t},\mu_{t},\nu_{t}=1] \numberthis \\
=\pix&
\mathbb{P}_{p}\{\nu_{t+1}=1|\lambda_{t},\mu_{t},\nu_{t}=1\} \\
&\cdot\mathbb{E}_{p,q}[V^{*}(M(\lambda_{t},\mu_{t},\omega_{t+1}),1;\eta)|\lambda_{t},\mu_{t}, \\
&~~~\nu_{t+1}=1]
+
\mathbb{P}_{p}\{\nu_{t+1}=0|\lambda_{t},\mu_{t},\nu_{t}=1\} \\
&\cdot V^{*}(\bar{M}(\lambda_{t},\mu_{t}),0;\eta) \numberthis
\end{align*}
So we can rewrite:
\begin{align*}
&
{\textstyle\sum}_{\omega^{\prime}_{t+1}\in\Omega}
\mathbb{P}_{p,q}\{\nu_{t+1}=1,\omega_{t+1}|\lambda_{t},\mu_{t},\nu_{t}=1\} \\
&\cdot V^{*}(M(\lambda_{t},\mu_{t},\omega_{t+1}),1;\eta) \numberthis \\
=\pix&
{\textstyle\sum}_{\omega^{\prime}_{t+1}\in\Omega}(
V^{*}(M(\lambda_{t},\mu_{t},\omega_{t+1}),1;\eta) \\
&\cdot{\textstyle\sum}_{\theta^{\prime}\in\Theta}(1-p_{\theta^{\prime},\lambda_{t}})q_{\theta^{\prime},\lambda_{t}}(\omega_{t+1})\mu_{t}(\theta^{\prime})) \numberthis \\
=\pix&
\mathbb{P}_{p}\{\nu_{t+1}=1|\lambda_{t},\mu_{t},\nu_{t}=1\} \\
&\cdot\mathbb{E}_{p,q}[V^{*}(M(\lambda_{t},\mu_{t},\omega_{t+1}),1;\eta)|\lambda_{t},\mu_{t}, \\
&~~~\nu_{t+1}=1] \numberthis
\end{align*}
Hence each $Q$-factor for acquisitions can be expressed:
\begin{align*}
&Q^{*}_{\lambda_{t}}(\mu_{t},\nu_{t};\eta) \numberthis \\
=\pix&
(1-\nu_{t})V^{*}(\mu_{t},0;\eta)
+
\eta_{\text{\pix c},\lambda_{t}}
c_{\lambda_{t}} \\
&+
(V^{*}(\bar{M}(\lambda_{t},\mu_{t}),0;\eta)
{\textstyle\sum}_{\theta^{\prime}\in\Theta}p_{\theta^{\prime},\lambda_{t}}\mu_{t}(\theta^{\prime}) \\
&+
{\textstyle\sum}_{\omega^{\prime}_{t+1}\in\Omega}(V^{*}(M(\lambda_{t},\mu_{t},\omega^{\prime}_{t+1}),1;\eta) \\
&\cdot{\textstyle\sum}_{\theta^{\prime}\in\Theta}(1-p_{\theta^{\prime},\lambda_{t}})q_{\theta^{\prime},\lambda_{t}}(\omega^{\prime}_{t+1})\mu_{t}(\theta^{\prime}))
)\nu_{t} \numberthis \\
=\pix&
(1-\nu_{t})V^{*}(\mu_{t},0;\eta)
+
\eta_{\text{\pix c},\lambda_{t}}
c_{\lambda_{t}} \\
&+
(V^{*}(\bar{M}(\lambda_{t},\mu_{t}),0;\eta)
{\textstyle\sum}_{\theta^{\prime}\in\Theta}p_{\theta^{\prime},\lambda_{t}}\mu_{t}(\theta^{\prime}) \\
&+
\mathbb{P}_{p}\{\nu_{t+1}=1|\lambda_{t},\mu_{t},\nu_{t}=1\} \\
&\cdot\mathbb{E}_{p,q}[V^{*}(M(\lambda_{t},\mu_{t},\omega_{t+1}),1;\eta)|\lambda_{t},\mu_{t}, \\
&~~~\nu_{t+1}=1]
)\nu_{t} \numberthis \\
=\pix&
(1-\nu_{t})V^{*}(\mu_{t},0;\eta)
+
\eta_{\text{\pix c},\lambda_{t}}
c_{\lambda_{t}} \\
&+
(\frac{{\textstyle\sum}_{\theta^{\prime}\in\Theta}\eta_{\text{\pix b},\theta^{\prime}}p_{\theta^{\prime},\lambda_{t}}\mu_{t}(\theta^{\prime})}{{\textstyle\sum}_{\theta^{\prime}\in\Theta}p_{\theta^{\prime},\lambda_{t-1}}\mu_{t-1}(\theta^{\prime})}
{\textstyle\sum}_{\theta^{\prime}\in\Theta}p_{\theta^{\prime},\lambda_{t}}\mu_{t}(\theta^{\prime}) \\
&-
(
V^{*}(\mu_{t},1;\eta)
-
\mathbb{P}_{p}\{\nu_{t+1}=1|\lambda_{t},\mu_{t},\nu_{t}=1\} \\
&\cdot\mathbb{E}_{p,q}[V^{*}(M(\lambda_{t},\mu_{t},\omega_{t+1}),1;\eta)|\lambda_{t},\mu_{t}, \\
&~~~\nu_{t+1}=1]
)
+
V^{*}(\mu_{t},1;\eta)
)\nu_{t} \numberthis \\
=\pix&
(1-\nu_{t})V^{*}(\mu_{t},0;\eta)
-
I_{t}(\lambda_{t})
+
\eta_{\text{\pix c},\lambda_{t}}
c_{\lambda_{t}} \\
&+
(\frac{{\textstyle\sum}_{\theta^{\prime}\in\Theta}\eta_{\text{\pix b},\theta^{\prime}}p_{\theta^{\prime},\lambda_{t}}\mu_{t}(\theta^{\prime})}{{\textstyle\sum}_{\theta^{\prime}\in\Theta}\eta_{\text{\pix b},\theta^{\prime}}\mu_{t}(\theta^{\prime})}
{\textstyle\sum}_{\theta^{\prime}\in\Theta}\eta_{\text{\pix b},\theta^{\prime}}\mu_{t}(\theta^{\prime}) \\
&+
V^{*}(\mu_{t},1;\eta)
)\nu_{t} \numberthis \\
=\pix&
(1-\nu_{t})V^{*}(\mu_{t},0;\eta)
-
I_{t}(\lambda_{t})
+
\eta_{\text{\pix c},\lambda_{t}}
c_{\lambda_{t}} \\
&+
((1-S_{t}(\lambda_{t}))
{\textstyle\sum}_{\theta^{\prime}\in\Theta}\eta_{\text{\pix b},\theta^{\prime}}\mu_{t}(\theta^{\prime}) \\
&+
V^{*}(\mu_{t},1;\eta)
)\nu_{t} \numberthis
\end{align*}
Consider $\nu_{t}=1$, and suppose $\mu_{t}\in\mathcal{T}(\eta)$. Then:
\begin{align*}
Q^{*}_{\lambda_{t}}
&=
V^{*}(\mu_{t},1;\eta)
-
I_{t}(\lambda_{t}) \\
&-
(S_{t}(\lambda_{t})-1)
{\textstyle\sum}_{\theta^{\prime}\in\Theta}\eta_{\text{\pix b},\theta^{\prime}}\mu_{t}(\theta^{\prime})
+
\eta_{\text{\pix c},\lambda_{t}}
c_{\lambda_{t}} \numberthis \\
&=
-h(I_{t}(\lambda_{t}),S_{t}(\lambda_{t}))
+
\eta_{\text{\pix c},\lambda_{t}}
c_{\lambda_{t}} \numberthis
\end{align*}
for some $h$ increasing in $I_{t}(\lambda_{t})$ and $S_{t}(\lambda_{t})$, since other terms do not depend on the choice of $\lambda_{t}$. Hence minimizing $Q^{*}_{\lambda_{t}}$ is equivalent to maximizing $h(I_{t}(\lambda_{t}),S_{t}(\lambda_{t}))
-
\eta_{\text{\pix c},\lambda_{t}}
c_{\lambda_{t}}$. For completeness, consider also $\nu_{t}=0$. But clearly $\mathcal{T}(\eta)$ is empty since $Q^{*}_{\lambda_{t}}=\bar{Q}+\eta_{\text{\pix c},\lambda_{t}}c_{\lambda_{t}}$, therefore $\mu_{t}\not\in\mathcal{T}(\eta)$ and there is no acquisition hence no tradeoff.

\Paste{thm_bayesian_inference}
First, the likelihood term is given by:
\begin{align*}
&\mathbb{P}_{p,q}\{\dataset|\pi_{\rho}^{\kappa}\text{\small(}...;\eta\text{\small)}\} \numberthis \\
=\pix&
\mathbb{P}_{p,q}\{(\tilde{\lambda}_{n,0:\tau-1},\tilde{\omega}_{n,1:\tau})_{n=1}^{N}|\kappa,\eta,\rho\} \numberthis \\
=\pix&
{\textstyle\int}_{\Delta(\Theta)}
{\textstyle\prod}_{n=1}^{N}
{\textstyle\prod}_{t=0}^{\tau-1}
(
\mathbb{P}\{\tilde{\lambda}_{n,t}|\mu_{n,t},\nu_{n,t},\kappa,\eta,\rho\} \\
&\cdot
\mathbb{P}_{p,q}\{\nu_{n,t+1},\tilde{\omega}_{n,t+1}|\tilde{\lambda}_{n,t},\mu_{n,t},\nu_{n,t}\}) \\
&
~~~d\mathbb{P}\{\mu_{n,t+1}|\tilde{\lambda}_{n,t},\mu_{n,t},\tilde{\omega}_{n,t+1}\} \numberthis \\
=\pix&
{\textstyle\prod}_{n=1}^{N}
{\textstyle\prod}_{t=0}^{\tau-1}
(
\mathbb{P}\{\tilde{\lambda}_{n,t}|\mu_{n,t}= \\
&~~M(\lambda_{n,t-1},\mu_{n,t-1},\tilde{\omega}_{n,t}),\nu_{n,t},\kappa,\eta,\rho\} \\
&\cdot\mathbb{P}_{p,q}\{\nu_{n,t+1},\tilde{\omega}_{n,t+1}|\tilde{\lambda}_{n,t},\mu_{n,t},\nu_{n,t}\}) \numberthis \\
=\pix&
{\textstyle\prod}_{n=1}^{N}
{\textstyle\prod}_{t=0}^{\tau-1}
\pi_{\rho}^{\kappa}(\tilde{\lambda}_{n,t}|M(\lambda_{n,t-1},\mu_{n,t-1},\tilde{\omega}_{n,t}), \\
&\nu_{n,t};\eta)
\mathbb{P}_{p,q}\{\nu_{n,t+1},\tilde{\omega}_{n,t+1}|\tilde{\lambda}_{n,t},\mu_{n,t},\nu_{n,t}\} \numberthis
\end{align*}
where for third equality recall the Bayesian recognition model (which involves no uncertainty), and the fourth equality is just our definition of a strategy. So the posterior is:
\begin{align*}
&\mathbb{P}_{p,q}\{\pi_{\rho}^{\kappa}\text{\small(}...;\eta\text{\small)}|\dataset\} \numberthis \\
=\pix&
\frac{1}{Z}
\mathbb{P}\{\kappa\}
\mathbb{P}\{\eta|\kappa\}
\mathbb{P}\{\rho\}
{\textstyle\prod}_{n=1}^{N}
{\textstyle\prod}_{t=0}^{\tau-1}( \\
&\pi_{\rho}^{\kappa}(\tilde{\lambda}_{n,t}|
M(\lambda_{n,t-1},\mu_{n,t-1},\tilde{\omega}_{n,t}),\nu_{n,t};\eta) \\
&\cdot
\mathbb{P}_{p,q}\{\nu_{n,t+1},\tilde{\omega}_{n,t+1}|\tilde{\lambda}_{n,t},\mu_{n,t},\nu_{n,t}\}) \numberthis
\end{align*}
where the normalizing constant is given by:
\begin{align*}
Z=&{\textstyle\int}_{\mathcal{K}}
{\textstyle\int}_{\mathcal{H}}
{\textstyle\int}_{\mathbb{R}}
{\textstyle\prod}_{n=1}^{N}
{\textstyle\prod}_{t=0}^{\tau-1}
( \\
&\pi_{\rho}^{\kappa}(\tilde{\lambda}_{n,t}|M(\lambda_{n,t-1},\mu_{n,t-1},\tilde{\omega}_{n,t}),\nu_{n,t};\eta) \\
&\cdot\mathbb{P}_{p,q}\{\nu_{n,t+1},\tilde{\omega}_{n,t+1}|\tilde{\lambda}_{n,t},\mu_{n,t},\nu_{n,t}\}
) \\
&~~~d\mathbb{P}\{\rho\}
d\mathbb{P}\{\eta|\kappa\}
d\mathbb{P}\{\kappa\} \numberthis
\end{align*}
Note that the dynamics term does not depend on $\kappa$, $\eta$, or $\rho$ and cancels out from the numerator and denominator, so:
\begin{align*}
&\mathbb{P}\{\pi_{\rho}^{\kappa}\text{\small(}...;\eta\text{\small)}|\dataset\} \numberthis \\
=\pix&
\frac{1}{Z^{\prime}}
\mathbb{P}\{\kappa\}
\mathbb{P}\{\eta|\kappa\}
\mathbb{P}\{\rho\}
{\textstyle\prod}_{n=1}^{N}
{\textstyle\prod}_{t=0}^{\tau_{n}-1}( \\
&\pi_{\rho}^{\kappa}(\tilde{\lambda}_{n,t}|M(\lambda_{n,t-1},\mu_{n,t-1},\tilde{\omega}_{n,t}),\nu_{n,t};\eta)) \numberthis
\end{align*}

\Paste{thm_diff_posterior}
First, we show each $\tilde{Q}\raisebox{1.5pt}{$^{*}_{\tilde{\lambda_{n,t}}}$}(\mu_{n,t},\nu_{n,t};\eta)$ is concave in $\eta$, for which it is sufficient to show each $V^{*}(\mu_{n,t},\nu_{n,t};\eta)$ is concave. Let $\pi$ be the Bayes-optimal strategy corresponding to the point $\upsilon\eta+(1-\upsilon)\eta^{\prime}$ for $\upsilon\in(0,1)$. Then:
\begin{align*}
&V^{*}(\mu_{n,t},\nu_{n,t};\upsilon\eta+(1-\upsilon)\eta^{\prime}) \numberthis \\
=\pix&
V^{\pi}(\mu_{n,t},\nu_{n,t};\upsilon\eta+(1-\upsilon)\eta^{\prime}) \numberthis \\
=\pix&
\upsilon V^{\pi}(\mu_{n,t},\nu_{n,t};\eta)
+
(1-\upsilon)V^{\pi}(\mu_{n,t},\nu_{n,t};\eta^{\prime}) \numberthis \\
\geq\pix&
\upsilon V^{*}(\mu_{n,t},\nu_{n,t};\eta)
+
(1-\upsilon)V^{*}(\mu_{n,t},\nu_{n,t};\eta^{\prime}) \numberthis
\end{align*}
{where the second equality follows from linearity of expectations, and the inequality from the fact that any the optimal strategy for $\eta$ and $\eta^{\prime}$ respectively is by definition at least as good as any other strategy $\pi$ (which in this case is only known to be optimal for some other point $\upsilon\eta+(1-\upsilon)\eta^{\prime}$). But for any function $f:\mathbb{R}^{d}\rightarrow\mathbb{R}$ for some finite $d$ that is concave, the set of points of non-differentiability is at most countable. Therefore $\tilde{Q}\raisebox{1.5pt}{$^{*}_{\tilde{\lambda_{n,t}}}$}(\mu_{n,t},\nu_{n,t};\eta)$ is differentiable (almost everywhere). Now, the likelihood is a differentiable in $\rho$ and in each $\tilde{Q}\raisebox{1.5pt}{$^{*}_{\tilde{\lambda_{n,t}}}$}(\mu_{n,t},\nu_{n,t};\eta)$, so the posterior is differentiable (almost everywhere) in $\eta$ and $\rho$ as long as the priors $\mathbb{P}\{\eta\pix|*\}$ and $\mathbb{P}\{\rho\}$ themselves are differentiable.\parfillskip=0pt\par}

\end{document}